%% file: main.tex
\documentclass[sigconf]{acmart}

\AtBeginDocument{%
  \providecommand\BibTeX{{%
    \normalfont B\kern-0.5em{\scshape i\kern-0.25em b}\kern-0.8em\TeX}}}

\usepackage{nicefrac}
\usepackage{multirow}
\usepackage{verbatim}
\usepackage[english]{babel}
\usepackage{graphicx}
\usepackage{amsmath}
\usepackage{url}
\usepackage{hyperref}        
\usepackage{booktabs}        
\usepackage{tabularx}        
\usepackage{amsthm,amsfonts,mathtools,bbm}
\usepackage{enumerate}
\usepackage{subfigure}
\usepackage{microtype}
\usepackage{minitoc}
\usepackage{comment,bm}
\usepackage[toc,page,header]{appendix}
\usepackage{textcomp}
\usepackage{cancel}
\usepackage{tikz}
\usepackage{algorithm}
\usepackage{algpseudocode}
\usepackage[labelfont=bf,format=plain,justification=raggedright,singlelinecheck=false]{caption}
\usepackage{enumitem}
\setlist[itemize]{leftmargin=*}
\setlist[enumerate]{leftmargin=*}
\usepackage{float}
\usepackage{ragged2e}

\setcopyright{acmlicensed}
\copyrightyear{2018}
\acmYear{2018}
\acmDOI{XXXXXXX.XXXXXXX}

\acmConference[Conference acronym 'XX]{Make sure to enter the correct
  conference title from your rights confirmation emai}{June 03--05,
  2018}{Woodstock, NY}
\acmBooktitle{Woodstock '18: ACM Symposium on Neural Gaze Detection,
 June 03--05, 2018, Woodstock, NY} 
\acmISBN{978-1-4503-XXXX-X/18/06}

\input{utils}

\begin{document}

\title{\textsc{SABLE}: Secure And Byzantine robust LEarning}


\author{Antoine Choffrut}
\affiliation{
  \institution{CEA LIST}
  \city{Paris}
  \country{France}}
\email{antoine.choffrut@cea.fr}

\author{Rachid Guerraoui}
\affiliation{
  \institution{Ecole Polytechnique Federale de Lausanne (EPFL)}
  \city{Lausanne}
  \country{Switzerland}}
\email{rachid.guerraoui@epfl.ch}

\author{Rafaël Pinot}
\affiliation{
  \institution{Ecole Polytechnique Federale de Lausanne (EPFL)}
  \city{Lausanne}
  \country{Switzerland}}
\email{rafael.pinot@epfl.ch}

\author{Renaud Sirdey}
\affiliation{
  \institution{CEA LIST}
  \city{Paris}
  \country{France}}
\email{renaud.sirdey@cea.fr}

\author{John Stephan}
\affiliation{
  \institution{Ecole Polytechnique Federale de Lausanne (EPFL)}
  \city{Lausanne}
  \country{Switzerland}}
\email{john.stephan@epfl.ch}

\author{Martin Zuber}
\affiliation{
  \institution{CEA LIST}
  \city{Paris}
  \country{France}}
\email{martin.zuber@cea.fr}

\renewcommand{\shortauthors}{Trovato and Tobin, et al.}

\input{0_abstract}



\maketitle

\input{1_intro}

\input{2_background}

\input{3_algorithm}

\input{4_hts}

\input{5_experiments}

\input{6_related_work}


\bibliographystyle{ACM-Reference-Format}
\bibliography{sample-base}

\appendix

\input{notation}

\input{app_byzantine_friendliness}

\input{app_exp_setup}
\input{app_exp_results}

\end{document}

%% file: utils.tex
\newcommand{\algoname}{\textsc{SABLE}}
\newcommand{\aggregator}{\textsc{HTS}}
\newcommand{\aggregatortwo}{\textsc{HMED}}

\newcommand{\defn}[1]{\textit{#1}}
\newcommand{\1}{\mathbf{1}}

\newcommand{\slot}{\texttt{s}}
\newcommand{\vslots}{\texttt{v}}
\newcommand{\digit}{\texttt{u}}
\newcommand{\pt}{\texttt{pt}}
\newcommand{\ct}{\texttt{ct}}

\newcommand{\pk}{\texttt{pk}}
\newcommand{\sk}{\texttt{sk}}
\newcommand{\encrypt}{\texttt{Encr}}
\newcommand{\decrypt}{\texttt{Decr}}

\newcommand{\batch}{\texttt{Batch}}

\newcommand{\negative}{\texttt{Neg}}
\newcommand{\zero}{\texttt{Zero}}

\newcommand{\lessthan}{\texttt{LT}}

\newcommand{\btwn}{\texttt{Btw}}
\newcommand{\trimmedsum}{\aggregator{}}
\newcommand{\median}{\aggregatortwo} 
\newcommand{\extract}{\texttt{Ext}}

\newcommand{\rank}{\texttt{Rank}}

\newcommand{\ordp}{N} 
\newcommand{\nslots}{\ell}

\newtheorem{remark}{Remark}

\algdef{SE}[SUBALG]{Indent}{EndIndent}{}{\algorithmicend\ }%
\algtext*{Indent}
\algtext*{EndIndent}

%% file: 0_abstract.tex
\begin{abstract}
Due to the widespread availability of data, machine learning (ML) algorithms are increasingly being implemented in distributed topologies, wherein various nodes collaborate to train ML models via the coordination of a central server.
However, distributed learning approaches face significant vulnerabilities, primarily stemming from two potential threats.
Firstly, the presence of \emph{Byzantine} nodes poses a risk of corrupting the learning process by transmitting inaccurate information to the server.
Secondly, a \emph{curious} server may compromise the privacy of individual nodes, sometimes reconstructing the entirety of the nodes' data.
Homomorphic encryption (HE) has emerged as a leading security measure to preserve privacy in distributed learning under non-Byzantine scenarios. However, the extensive computational demands of HE, particularly for high-dimensional ML models, have deterred attempts to design purely homomorphic operators for non-linear robust aggregators.
This paper introduces SABLE, the first homomorphic and Byzantine robust distributed learning algorithm.
\textsc{SABLE} leverages \textsc{HTS}, a novel and efficient homomorphic operator implementing the prominent coordinate-wise trimmed mean robust aggregator.
Designing \textsc{HTS} enables us to implement \textsc{HMED}, a novel homomorphic median aggregator. 
Extensive experiments on standard ML tasks demonstrate that \textsc{SABLE} achieves practical execution times while maintaining an ML accuracy comparable to its non-private counterpart.
\end{abstract}

\keywords{Homomorphic Encryption, Byzantine Machine Learning}

%% file: 1_intro.tex
\section{Introduction}~\label{sec_intro}
The quest for more accurate and versatile AI technologies has led machine learning (ML) researchers to considerably increase the complexity of the tasks they consider. As such, the size of learning models and datasets is increasing daily. To scale the computational resources and adapt to these new tasks, it is now common to rely on distributed methodologies in which the learning procedure is fragmented into several sub-tasks partitioned over different machines, or \emph{nodes}. The nodes execute their respective tasks in parallel, and coordinate their actions usually with the help of a central \emph{server}. Many distributed ML algorithms require the distribution of gradient computations across nodes, as in the renown distributed stochastic gradient descent (DSGD) method~\cite{bertsekas2015parallel}.
In DSGD, the server maintains an estimate of the model which is updated iteratively by \emph{averaging} the gradients computed by the nodes. However, as effective as it can be, distributed ML is also vulnerable to {\em misbehaving} nodes.
Indeed, a handful of nodes sending incorrect information 
can highly disrupt the learning. Such a behavior can be caused by software and hardware bugs, poisoned data, or even by malicious nodes controlling part of the system. Following the terminology of distributed computing, these nodes are called {\em Byzantine}~\cite{lamport82}.
In the context of DSGD, Byzantine nodes may thus send erroneous gradients to the server, such as the negative of the true gradient. 
The presence of such nodes, which is arguably inevitable in real-world settings, affects the trust in the efficacy of distributed ML solutions. Standard distributed learning algorithms, such as DSGD, completely fail to output accurate ML models in such an adversarial setting~\cite{krum}.

To mitigate the impact of Byzantine nodes in distributed learning, one can use a {\em robust} variant of DSGD, called Robust-DSGD.\footnote{We use Robust-DSGD to denote the robust framework of distributed learning algorithms introduced in~\cite{RESAM}. See Section~\ref{sec_robust_DSGD} for more details.}
The main modification compared to DSGD is to replace the averaging operation at the server by a \emph{non-linear} robust aggregator.
In fact, several such aggregation schemes have been proposed, such as coordinate-wise median (CWMED)~\cite{yin2018byzantine}, geometric median (GM)~\cite{chen2017distributed}, coordinate-wise trimmed mean (CWTM)~\cite{yin2018byzantine}, Krum~\cite{krum}, among others.
When choosing the robust aggregator in a suitable way, Robust-DSGD can be \emph{proven} to be robust to Byzantine nodes, as described and analyzed in~\cite{RESAM,allouah2023fixing,karimireddy2022byzantinerobust,Karimireddy2021History}. Furthermore, this solution performs well in practice against Byzantine attacks.

Yet, Robust-DSGD assumes that the central server is trusted. Third party servers are often operated by service providers, and one might reasonably assume that they seldom fail. However, the provider itself can be {\em curious}. One of the motivations of distributed learning is that nodes do not share their respective data, providing some level of sovereignty over their training datasets.
However, exchanging gradients in the clear enables the server to infer sensitive information, and sometimes even recover the totality of the nodes' data~\cite{DLG, IDLG, InvertingGradients}. Therefore, it is paramount to design solutions that, in addition to being robust to Byzantine nodes, allow the honest nodes to protect their data against a \emph{curious} server.

To overcome this problem, homomorphic encryption (HE) has been proposed as a security solution to preserve privacy in distributed settings where clients attempt to offload part of their computations to a third party server, particularly for ML applications~\cite{sok23, speed, privilege, fl-vc,juvekar18_gazelle,lou20_glyph,gilad-bachrach16_CryptoNet}.
In short, HE schemes allow certain mathematical operations (such as additions and multiplications) to be performed directly on encrypted data and without prior decryption. When decrypted, the output is the same as what would have been produced from the cleartext data.
This property unlocks the use of HE schemes in distributed learning tasks for high-stake applications such as medicine and finance.
Indeed, this enables several data providers to collaboratively train ML models with the help of a third party server, without leaking any information to the server. For example, in DSGD, applying an \emph{additive HE} scheme would allow an untrusted server to average encrypted gradients without compromising the privacy of the nodes.
Previous works~\cite{zhang2020, sébert2022protecting} have demonstrated that homomorphic averaging could be performed on models of 500k parameters with sub-1\% computation overhead throughout the training.
However, doing the same kind of efficient homomorphic adaptation to Robust-DSGD remains a very challenging task. The following question thus naturally arises.
\begin{center}
    \textit{\textbf{Can we design an HE implementation of Robust-DSGD?}}
\end{center}
Answering this question boils down to overcoming the performance limitations of HE.
While its costs are considered to be relatively low when implementing linear operations (such as averaging), HE is notoriously much more computationally expensive when dealing with non-linear operations on high-dimensional data, which are at the core of Robust-DSGD.
Indeed, HE schemes come with a limited set of (native) homomorphic operators, i.e., additions and multiplications, notably excluding comparisons as well as divisions.
Operations that one would consider elementary often offer a challenge in the encrypted domain to HE researchers. The crux is that HE schemes typically feature few data structures and limited flexibility to customize their size and shape to one's needs.
In order to perform more elaborate operations beyond those native to the HE cryptosystem, one must design an algorithm implementing the desired operation together with an encoding of data into plaintexts that is amenable to this algorithm. Comparison operators, which are the essence of (non-linear) Byzantine robust aggregators, constitute a perfect example of this challenge and have been the subject of a number of recent papers~\cite{TanEtAl, IZ21, ZS21, CZ22}.

Unfortunately, some of these solutions~\cite{TanEtAl, IZ21} only allow a very limited set of homomorphic operations on ciphertexts \emph{after the operation of comparison}.
These methods are therefore inadequate for the design of robust aggregators, which typically require much more complex operators such as \textit{min}/\textit{max} and \textit{nearest neighbors} computations.
Other works~\cite{ZS21, CZ22} propose solutions that are only applicable to input vectors of small dimension and therefore are not suited in our case.
This certainly makes their efficiency questionable when applied to modern-day and practically-relevant ML models where the number of parameters, i.e., the size of the gradients, can easily exceed the millions (or even billions).
This inherent difficulty in making HE amenable to non-linear operations in high dimensions thus introduces prohibitive computational costs when homomorphically implementing Byzantine robust aggregators.

\subsection{Our Contributions}
We consider a distributed learning system comprised of $n$ nodes and a server. We assume $f < \nicefrac{n}{2}$ nodes are Byzantine (i.e., they may send any vector to the server), while the remaining nodes are honest (i.e., correct). Furthermore, the server is considered to be \textit{honest-but-curious}, i.e., the server correctly aggregates the nodes' vectors, but it also tries to use the received gradients to infer information about the nodes' data.

In this context, we propose a novel algorithm for Secure And Byzantine Robust LEarning, namely \algoname{}.
To the best of our knowledge, this algorithm constitutes the first proposal of a homomorphic extension of DSGD that is robust to Byzantine nodes.
\algoname{} incorporates the novel robust aggregator \textit{Homomorphic Trimmed Sum} (\aggregator{}), which we detail below, that simultaneously mitigates the impact of Byzantine nodes and provides strong cryptographic security ($\lambda\geq128$).

Contrarily to previous approaches (see Section~\ref{sec_conclusion}), \algoname{} does not leak any information on the nodes' data to the server, thereby preserving the privacy of the nodes. 
From a high level, in each iteration of \algoname{}, every node computes the gradient on a batch of its data, homomorphically encrypts the gradient, and then sends it to the server.
The server then employs \aggregator{} to robustly aggregate the encrypted gradients, and sends the resulting ciphertext back to the nodes.
Finally, the nodes decrypt the encrypted aggregate, and then locally update their estimate of the model parameters.
Furthermore, we feature in \algoname{} the option of \textit{node subsampling}, whereby the server randomly selects a subset of the nodes and computes \aggregator{} only on their corresponding ciphertexts. This technique decreases the computational cost of the aggregation in the encrypted domain. We show in Section~\ref{sec_exp_he_perf} that node subsampling significantly improves the execution time of \aggregator{} (with no impact on the HE security). Besides, by carefully tuning the number of nodes to be sampled, \algoname{} with node subsampling yields similar empirical performances to the original algorithm in terms of ML accuracy.
Beneath the foundation of \algoname{} lie three distinct contributions, expounded upon in the ensuing paragraphs.

\medskip
\textbf{HE-compliance of existing aggregators.}
First, we determine the most appropriate aggregator to implement homomorphically in terms of computational cost.
We consider an exhaustive list of prominent robust aggregators from the literature, and qualitatively identify their major computational bottlenecks in the encrypted domain.
Our analysis promotes the coordinate-wise aggregators CWTM~\cite{yin2018byzantine} and CWMED~\cite{yin2018byzantine} for their inherent compatibility with batching-friendly HE cryptosystems, a feature we argue is crucial for high-dimensional ML models. Indeed, batching allows many cleartext operations to be parallelized into a single homomorphic computation, significantly amortizing their computational cost.
We choose CWTM as the main robust aggregator to be used in \algoname{} due to its theoretical and empirical superiority in ML performance~\cite{RESAM, allouah2023fixing}.
(Due to space limitations,  details of the analysis are deferred  to Appendix~\ref{app_fhe_frendliness}.)

\medskip
\textbf{Novel homomorphic operators.}
Second, we introduce \aggregator{}, the first solution for computing CWTM homomorphically.
The design of \aggregator{} poses a number of challenges. While batching can help to implement CWTM homomorphically, it is not sufficient on its own as CWTM requires a sorting operation which is costly to perform in the encrypted domain. Indeed, some batching-friendly homomorphic sorting operators have been proposed in the literature \cite{TanEtAl, IZ21, NarumanchiGEG17}; but they cannot be directly used for CWTM because the latter also requires the addition of a selection of elements \emph{after} they have been sorted.
In fact, the encoding of integer values into plaintexts in these solutions is ill-adapted to our problem. To circumvent this limitation, we devise a plaintext encoding method that exploits an extra degree of freedom in HE schemes (specifically, we decouple the base in the integer decomposition from the plaintext modulus).
This serves multiple purposes simultaneously: (1) it allows to customize data structures that are better adapted to the dimension of vectors and the range of their values; (2) it reduces the multiplicative depth of the homomorphic circuits and the computational time of their execution; and (3) it offers better security margins. 
The upshot is that with this extra degree of freedom, we are able to implement CWTM homomorphically, while it was either impractical~\cite{ZS21,CZ22} or out of reach~\cite{TanEtAl,IZ21} with existing solutions. 
As an interesting special case of \aggregator{}, we also obtain a homomorphic implementation of CWMED, which we call \textit{Homomorphic Median} (\aggregatortwo{}), without incurring additional costs.



\medskip
\textbf{Empirical evaluation.}
Finally, we thoroughly benchmark the performance of our algorithm both in terms of computational cost and learning accuracy. Indeed, we implement \algoname{} on standard image classifications tasks with large models, and provide unit execution times of our implementation.
Our results show that \aggregator{} has practical execution times, especially when distributing the computations at the server across several cores. Furthermore, node subsampling significantly accelerates the computations up to 9$\times$.
We evaluate the robustness of \algoname{} in terms of ML accuracy in two distributed systems and under different Byzantine regimes.
Our experiments show that \algoname{} is consistently robust to state-of-the-art Byzantine attacks (such as \emph{fall of empires}~\cite{FOE}, \emph{a little is enough}~\cite{ALIE}, and \emph{label flipping}~\cite{allen2020byzantine}) and matches the performance of its non-private counterpart Robust-DSGD on MNIST~\cite{mnist} and CIFAR-10~\cite{cifar}, even when using small bit precisions (see Section~\ref{sec_exp}).
Moreover, we show in our experiments that, in executions where all nodes behave correctly (i.e., there is no Byzantine behavior), \algoname{} yields very close accuracies to its non-robust counterpart (i.e., private DSGD using additive HE).
This confirms that the robustness cost of our solution is small (in terms of loss in accuracy), and demonstrates the empirical relevance of \algoname{} in practice.

\subsection{Paper Outline}\label{sec:paper.outline}
The remainder of this paper is organized as follows. 
Section~\ref{sec_background} contains background information on Byzantine ML and HE needed to understand the paper. Section~\ref{sec_algo} presents \algoname{}, while Section~\ref{sec_fhe_aggregators} introduces our novel operator \aggregator{} and explains our encoding scheme. Section~\ref{sec_exp} evaluates the performance of \algoname{} on benchmark ML tasks. Finally, Section~\ref{sec_conclusion} discusses the different frameworks that we believe to be related to our work and presents concluding remarks. For ease of reading, we include a list of notations in Appendix~\ref{sec:notation}.

%% file: 2_background.tex
\section{Background}~\label{sec_background}
In Section~\ref{sec_robust_DSGD}, we explain the Robust-DSGD protocol and discuss its robustness guarantees against Byzantine nodes. Then, we introduce the concept of HE in Section~\ref{sec_he_for_ml}, and present the coordinate-wise robust aggregators CWTM and CWMED in Section~\ref{sec_fhe_friendliness}.

\subsection{Distributed Byzantine Learning}~\label{sec_robust_DSGD}
We consider a distributed architecture composed of $n$ nodes and an honest-but-curious server. Every node $i$ has a local dataset $D_i$.
Furthermore, we assume the presence of $f < \nicefrac{n}{2}$ Byzantine nodes in the system.\footnote{Otherwise, no meaningful learning guarantees can be obtained~\cite{liu2021approximate}.} 
Following the terminology of Byzantine ML~\cite{RESAM, allouah2023fixing, Karimireddy2021History}, Byzantine nodes \textit{only} aim to disrupt the algorithm in terms of ML performance by sending erroneous vectors to the server, i.e., they do not deviate from the rest of the protocol.
The remaining $n - f$ nodes are considered to be correct (a.k.a., honest), and follow the prescribed protocol exactly. The objective of the honest nodes is to train a joint ML model over the collection of their individual datasets with the help of the server, despite the presence of Byzantine nodes.

As shown in~\cite{krum}, the averaging operation at the server in DSGD is arbitrarily manipulable by a single Byzantine gradient.
This renders the underlying algorithm vulnerable to Byzantine nodes, making it impossible for honest nodes to build an accurate ML model.
Consequently, recent works proposed a robust variant of DSGD, that we call Robust-DSGD~\cite{RESAM,allouah2023fixing, Karimireddy2021History, karimireddy2022byzantinerobust}.
In summary, Robust-DSGD enhances the robustness of DSGD by incorporating two additional features.
First, instead of sending the gradients directly to the server, the honest nodes first compute the Polyak's momentum~\cite{polyak1964some} on their gradients.
Second, the averaging operation at the server is replaced by a more complex Byzantine robust aggregator $F$. $F$ is designed to aggregate the nodes' momentums in a way that would mitigate the attacks launched by Byzantine nodes (in the form of incorrect gradients).

However, the aggregation method $F$ must be chosen carefully. Indeed, only a specific subset of methods has been proven to confer theoretical robustness guarantees to Robust-DSGD~\cite{RESAM, allouah2023fixing}.
This subset comprises the following aggregations: \textit{Krum} (and \textit{Multi-Krum})~\cite{krum}, \emph{geometric median} (GM)~\cite{chen2017distributed}, \emph{minimum diameter averaging} (MDA)~\cite{brute_bulyan}, \emph{coordinate-wise trimmed mean} (CWTM)~\cite{yin2018byzantine}, \emph{coordinate-wise median} (CWMED)~\cite{yin2018byzantine}, and \emph{mean-around-median} (MeaMed)~\cite{meamed}.\footnote{Despite the existence of other robust methods in the literature (e.g., CGE~\cite{gupta2020fault} and CC~\cite{Karimireddy2021History}), we only consider resilient aggregators that possess theoretical guarantees ensuring the convergence of the algorithm in the presence of $f < \nicefrac{n}{2}$ Byzantine nodes.}
Besides offering theoretical guarantees, these robust methods are shown to empirically mitigate state-of-the-art Byzantine attacks, when combined with momentum~\cite{RESAM, allouah2023fixing, distributed-momentum}. 

Our objective in this work is to design an HE-based version of Robust-DSGD, where the aggregator $F$ is implemented completely in the homomorphic domain. Note that in current ML tasks, the number of model parameters grows due to the large-scale availability of data. This implies that nodes in distributed learning would compute and exchange gradients of significant size. Given that HE is notoriously expensive when performing non-linear operations (a crucial pillar of Byzantine robustness~\cite{krum}) in high dimensions, we would like to use a cryptosystem that possesses \textit{batching} capabilities (see~\eqref{diag:batching}) to benefit from the parallelization of computations when encrypting gradients.
This would significantly amortize the execution times of homomorphic operators. Furthermore, since HE operators generally induce large computation and communication costs, we consider a cross-silo ML setting with a relatively small number of nodes (not exceeding 15) while tolerating large models.




\input{HE_for_ML}


\input{byzantine_friendliness}

%% file: HE_for_ML.tex
\subsection{Homomorphic Encryption for ML}\label{sec_he_for_ml}
HE generally refers to several cryptosystem families.
The first one is \textit{partial homomorphic encryption} (PHE), i.e.,  schemes that only support either additions or multiplications on encrypted data.
These schemes are thus intrinsically not able to implement non-linear functions, a critical constraint required to ensure robustness~\cite{krum}.

The second second family consists of \textit{fully homomorphic encryption} (FHE) schemes with a bootstrapping procedure, which allows an indefinite number of homomorphic operations, even after the instance parameters have been set. So far, TFHE~\cite{tfhe} is the only practical solution in this family.
While able to efficiently compute distances on small sets of low-dimensional vectors and having a programmable bootstrapping well suited for non-linear operations, TFHE does not offer large plaintext domains (e.g., it is typically limited to $\mathbb{Z}_{16}$) and does not support batching. The absence of these two critical properties, given the large number of dimensions of current ML models, makes TFHE a priori less suitable for this work.

Another family of HE cryptosystems is the class of \emph{somewhat homomorphic encryption} (SHE), for which an instance's parameters must be set according to the number and types of homomorphic operations that will be performed.  This family includes BGV~\cite{BGV} and BFV~\cite{B, FV} which work on integers, as well as CKKS~\cite{CKKS} which works on real and complex values.
These cryptosystems have a larger plaintext domain thanks to batching (see \eqref{diag:batching} in Section~\ref{subsec:bgv}), meaning that they can pack a number of values into a single ciphertext.
As a result, the homomorphic operators apply independently on all these values in a Single Instruction Multiple Data (SIMD) parallel fashion.
This means that cryptosystems in this family generally lead to acceptable \emph{amortized} costs for applications that apply a given algorithm to many inputs in parallel.
The parallelization and batch-friendliness properties of SHE schemes makes BGV, BFV, and CKKS desirable candidates for the cryptosystem to be used.
BGV and BFV both naturally manipulate integer values but can also handle fixed-precision values by scaling them and encoding them as integers, as long as the algorithm tolerates these precision errors (an issue which is more often than not overlooked by the HE community).
Since real-valued gradient coefficients can be adequately represented by rescaling, both BGV and BFV are adequate for our purpose, which is implementing a robust aggregator on real-valued vectors.
Furthermore, as shall be developed in this paper, both BGV and BFV provide us with the mechanics for performing homomorphic comparisons and sorting which are needed for the non linear functionality of the robust aggregators.

In this work, we focus on BGV and BFV, and do not investigate solutions over the CKKS scheme. Indeed, 
CKKS stands out as a distinctive variant of SHE cryptosystems, 
having the desirable property of operating more natively over floating-point data.
However, it presently lacks the machinery necessary for non-arithmetic operations, although some solutions may be emerging in the near future (e.g.,~\cite{DBLP:conf/cscml/AharoniDKMS23}). Regarding empirical evaluation, we concentrate on BGV within the mainstream implementations available for both BGV and BFV~\cite{seal,openFHE,helib}. Accordingly, we use the HElib~\cite{helib} library (version 2.2.0) to derive experimental results. This choice is made without sacrificing generality, as our results and methodologies are equally applicable to BFV. The selection of HElib is also motivated by its provision of the low-level flexibility needed to fine-tune the cryptosystem parameters to meet our specific requirements.



%% file: byzantine_friendliness.tex
\subsection{HE-compliance of Byzantine Aggregators}\label{sec_fhe_friendliness}
We qualitatively review the six prominent Byzantine aggregators mentioned in Section~\ref{sec_robust_DSGD} with respect to their compliance with the mainstream HE cryptosystems in terms of evaluation cost. Due to space limitations, we defer our complete analysis 
to Appendix~\ref{app_fhe_frendliness}.

We classify these operators into two categories.
On the one hand, geometric approaches~\cite{krum, brute_bulyan, chen2017distributed} select one or more vectors based on a global criterion, but are subject to a selection bottleneck in high dimensions. This makes these approaches difficult to implement efficiently in the homomorphic domain when dealing with high-dimensional ML models.
On the other hand, coordinate-wise aggregators~\cite{yin2018byzantine, meamed} are intrinsically batching-friendly since all vector coordinates can be processed independently.
These approaches are therefore a priori promising candidates to be implemented over BGV or any other batching-friendly SHE scheme.
In fact, their homomorphic evaluation scales well to large models, thanks to the SIMD parallelism available in each ciphertext and the fact that many such ciphertexts can be straightforwardly processed in parallel.
Amongst the considered coordinate-wise approaches, our analysis promotes the use of CWMED and CWTM, presented below.


\begin{itemize}
    \item \textbf{Coordinate-wise median (CWMED)~\cite{yin2018byzantine}}. Median calculations have recently been studied over batching-friendly cryptosystems~\cite{IZ21}. However, computing a median even over a small number of values requires large computational times. For example, \cite{IZ21} reports that a median over $16$ vectors of size $9352$ populated with $8$-bit values takes around $1$ hour. The \emph{amortized} time per component appears more reasonable ($\approx 0.38s$). These results are promising and hint that more practical computational times may be obtained using further optimizations or by decreasing the precision of the vectors and/or the number of inputs given to the operator.
    \item \textbf{Coordinate-wise trimmed mean (CWTM)~\cite{yin2018byzantine}.}
This method performs a coordinate-wise sorting of $n$ vectors and then averages the ($n-2f$) coordinates between positions $f$ and $n-f-1$.
Specifically, given $n$ vectors $x_0, ..., x_{n-1}$ of dimension $d$, CWTM sorts the vectors per coordinate (indexed with a superscript in~\eqref{eq:coordinate-wise.sorted}) to obtain $\overline{x}_0, \dots, \overline{x}_{n-1}$ satisfying
\begin{equation}
    \overline{x}_{0}^{(j)} \leq ... \leq \overline{x}_{n-1}^{(j)} \qquad 0\le j\le d-1\,,
    \label{eq:coordinate-wise.sorted}
\end{equation}
and returns the vector $y$ such that
\begin{equation}
    y = \frac{1}{n-2f} \sum_{i=f}^{n-f-1} \overline{x}_{i}
    \label{eq:cwtm}
\end{equation}
This operator should have a similar cost to CWMED, since the selection is replaced by an averaging over statically selected coordinates, and summation is generally a low-cost homomorphic operation.
Note that since divisions are quite impractical in the homomorphic domain, the averaging operation in CWTM can be replaced by a homomorphic summation followed by a post-decryption division on the nodes' side.
There is no added value in terms of security to perform the division prior to the decryption as the number of nodes is a public information.
\end{itemize}


Be it for the trimmed mean or the median, the algorithm in the encrypted domain requires to sort the values and compute their ranks. Since data-dependent branching is not possible in a homomorphic setting, all pairs of values must be compared, yielding quadratic complexity.
Also, in both algorithms, the rank of all elements must be homomorphically determined in order to decide which ones enter the final computation. 
Both algorithms thus exhibit the same computational complexity and multiplicative depth.

From an ML viewpoint, these two popular methods have been proven to provide state-of-the-art Byzantine robustness~\cite{RESAM, Karimireddy2021History, karimireddy2022byzantinerobust}. Furthermore, CWTM has been shown to be empirically and theoretically superior to CWMED~\cite{allouah2023fixing}.
As such, we choose in this work to achieve Byzantine resilience by homomorphically implementing CWTM over the BGV cryptosystem.

%% file: 3_algorithm.tex
\section{\algoname{}: Secure And Byzantine robust LEarning}\label{sec_algo}
In this section, we introduce and explain \algoname{}, our algorithm for secure and Byzantine robust learning.
We also present and discuss the acceleration that node subsampling provides.

\subsection{Algorithm Description}~\label{sec_algo_description}
\algoname{} is composed of a \textit{setup} phase followed by three steps that are iteratively and sequentially executed, namely \textit{momentum computation and encryption}, \textit{homomorphic robust aggregation}, and \textit{decryption and model update}.
\medskip

\textbf{Setup.}
The server chooses the initial model parameters $\theta_0$ and sends them to all nodes.
Every node $i \in \{0, \dots, n-1\}$ thus sets its initial set of parameters to $\theta_0$.
In every iteration $t \in \{1, \dots, T\}$, \algoname{} sequentially executes three different steps, as shown below.
\medskip

\textbf{1) Momentum computation and encryption.}
First, every node $i$ randomly samples a batch of datapoints from its local dataset $D_i$.
On this batch, a stochastic estimate of the gradient is computed using the model parameters from the previous iteration $\theta_{t-1}$, and is referred to as $g_t^{(i)}$.
Then, node $i$ computes the momentum $m_t^{(i)}$ of the gradient: $m_{t}^{(i)}= \beta m_{t-1}^{(i)} + (1-\beta) g_t^{(i)},$ where $m_0^{(i)} = 0$ and $\beta \in (0, 1)$ is referred to as the momentum coefficient~\cite{polyak1964some}.
Prior to encryption, every node must first quantize the computed momentum using the clamp parameter $C > 0$ and the bit precision $\delta > 1$ to obtain $\tilde{m}_{t}^{(i)}$ = QUA$(m_{t}^{(i)}, \delta, C)$.
To do so, node $i$ first \textit{clamps} $m_{t}^{(i)}$ using $C$, thus bringing all the coordinates of the clamped momentum into the range $[-C, C]$.
More formally, every coordinate $x$ of $m_{t}^{(i)}$ is replaced by $\min \left (\max(x, -C), C \right )$.
The clamped momentum is then multiplied by the quantization parameter $Q = \nicefrac{2^{\delta-1} - 1}{C}$, and rounded to the nearest integer per coordinate.
In other words, quantization guarantees that every coordinate of the vector $\tilde{m}_{t}^{(i)}$ is represented by $\delta$ bits.

Finishing this first step requires a few extra operations.
Each node $i$ then executes a \textit{split-encode-batch-encrypt} procedure that produces a vector $\vec{c}_t^{(i)}$ composed of multiple ciphertexts, and sends the resulting vector to the server.
Specifically, the aforementioned procedure consists in the following.
First, the dimension $d$ of the vector $\tilde{m}_t^{(i)}$ is typically much larger than the number $\nslots$ of quantities that can be stored into a single ciphertext.
Therefore, $\tilde{m}_t^{(i)}$ is \emph{split} into $K:=\lceil d/\nslots \rceil$ smaller vectors of dimension $\nslots$.
The integer values in each such smaller vector are then \emph{encoded} into the appropriate data structure in the BGV cryptosystem, namely polynomials whose length is denoted $\ordp$ in Section~\ref{sec_fhe_aggregators}.
Next, each such vector of $\nslots$ polynomials is stored into a plaintext via \emph{batching} as described in~\eqref{diag:batching}, Section~\ref{subsec:bgv}. Finally, each such plaintext is \emph{encrypted} using the BGV cryptosystem. 
In conclusion, every node thus sends a vector $\vec{c}_t^{(i)}$ composed of $K$ ciphertexts encrypting $\nslots$ values each.
\medskip

\textbf{2) Homomorphic robust aggregation.}
Upon receiving all vectors of ciphertexts from the nodes, the server robustly aggregates the submitted vectors using \aggregator{}, our homomorphic implementation of CWTM~\cite{yin2018byzantine} which we meticulously present in Section~\ref{sec_fhe_aggregators}.
The result is a vector of encrypted \textit{trimmed sums} $\vec{C}_t$ which is then broadcast to all nodes.
In this step, it is also possible for the server to aggregate the ciphertexts using \aggregatortwo{}, our homomorphic implementation of CWMED~\cite{yin2018byzantine} (see Section~\ref{sec_fhe_aggregators}).
This underscores the modularity of \algoname{} by the inclusion of a flexible option for robust aggregation, achieved through the use of \aggregator{} or \aggregatortwo{}.
\medskip

\textbf{3) Decryption and model update.}
Upon receiving $\vec{C}_t$ from the server, each node performs a \textit{decrypt-unbatch-decode-regroup} operation, which is essentially the reverse operation of \textit{split-encode-batch-encrypt} explained in step 1.
This results in a (single) plaintext vector $P_t$ of cleartext integers, which is identical for all nodes.
As explained in Section~\ref{sec_fhe_friendliness},  
each node then computes the \textit{trimmed mean} 
$TM_t = \frac{1}{n - 2f} P_t$
(see~\eqref{eq:coordinate-wise.sorted}-\eqref{eq:cwtm}).
Every node then locally updates its current estimate of the model parameters
$\theta_{t} = \theta_{t-1} - \gamma TM_{t},$ where $\gamma$ is the learning rate.
Note that when using \aggregatortwo{} instead of \aggregator{}, 
the plaintext $P_t$ at every node must not be divided by $n - 2f$ since \aggregatortwo{} directly computes the coordinate-wise median.
The update rule of the model thus becomes: $\theta_{t} = \theta_{t-1} - \gamma P_t$.
\medskip


We present the totality of the described protocol (including the option of node subsampling, which we explain in Section~\ref{sec_server_sub}) in Algorithm~\ref{algo}.
Note that \aggregator{} is used with some abuse of notation in Algorithm~\ref{algo}, which there refers to the coordinate-wise application of \aggregator{} on the vectors of ciphertexts received from all nodes.
In summary, \algoname{} defends against Byzantine nodes by aggregating the nodes' vectors using the robust homomorphic operators \aggregator{} or \aggregatortwo{}, which provide state-of-the-art robustness guarantees~\cite{allouah2023fixing, RESAM}.
Finally, \algoname{} protects the privacy of the nodes from the curious server via the encryption of the nodes' momentums.

\begin{remark}
As mentioned in Section~\ref{sec_robust_DSGD}, Byzantine nodes within the context of \algoname{} are constrained to arbitrary deviations solely affecting the learning process.
Byzantine nodes cannot collude with the server.
However, they may transmit erroneous or maliciously crafted momentums to the server (i.e., momentum computation in step 1), thereby contaminating the learning process.
Otherwise, they follow correctly the rest of the protocol (e.g., encryption in step 1).
Their influence is thus limited to the learning aspect and does not extend to compromising the overall security and communication protocol, such as tampering during encryption.
\end{remark}


\subsection{Acceleration by Node Subsampling}\label{sec_server_sub}
As mentioned in Section~\ref{sec_fhe_friendliness}, 
the running time of \aggregator{} (and \aggregatortwo{}) is quadratic in the number of aggregated ciphertexts.
Therefore, having more nodes in the system leads to a considerable increase in the execution time of the homomorphic aggregators.
For example, increasing $n$ from 5 to 15 leads to a significant increase of the \textit{amortized} execution time of \aggregator{} from 3.17 ms to 30.75 ms ($10\times$), when using a precision of 4 bits.
This motivates us to investigate the use of \emph{node subsampling} at the server prior to the aggregation, explained below.
In fact, when $2f + 1 < n$ and node subsampling is enabled, the server samples uniformly at random and without replacement $2f + 1$ nodes among $n$ in every step $t$, and executes \aggregator{} only on their corresponding ciphertexts.
This enables the server to aggregate fewer vectors, while still guaranteeing the presence of an honest majority among the subsampled nodes (since at most $f$ nodes among the $2f + 1$ sampled can be Byzantine).
This technique significantly reduces the computational time of the aggregators, and thus allows us to tolerate distributed systems of larger size.
Furthermore, we note that executing \aggregator{} with node subsampling is exactly equivalent to computing \aggregatortwo{} on the subsampled ciphertexts.
Indeed, \eqref{eq:cwtm} returns exactly the coordinate-wise median of the input vectors when $n = 2f + 1$.

\begin{algorithm}[ht!]
\caption{\algoname{}}
\label{algo}
\begin{flushleft}
\Comment{\justifying Initial model $\theta_0$, initial momentum $m_0^{(i)} = 0$ for all nodes $i$, clamp parameter $C > 0$, bit precision $\delta > 1$, learning rate $\gamma$, momentum coefficient $\beta \in (0, 1)$, number of nodes $n > 0$, number of Byzantine nodes $f \geq 0$, Boolean \textit{subsampling}, and number of steps $T$.
}
\end{flushleft}

\begin{algorithmic}[1]

\For{$t=1, \dots, T$}
\State{\textbf{\textup{Every}} \textcolor{teal}{\bf node} $i$ \textbf{\textup{does in parallel}}}
\Comment{as per Section~\ref{sec_algo_description}, \textbf{1)}}
\Indent
\State Compute gradient $g_t^{(i)}$ on $\theta_{t-1}$
\State Compute momentum $m_{t}^{(i)}= \beta m_{t-1}^{(i)} + (1-\beta) g_t^{(i)}$
\State Quantize momentum $\tilde{m}_{t}^{(i)}$ = QUA$(m_{t}^{(i)}, \delta, C)$
\State Compute $\vec{c}_{t}^{(i)}$ via \textit{split-encode-batch-encrypt} on $\tilde{m}_{t}^{(i)}$
\State Send $\vec{c}_{t}^{(i)}$ to \textcolor{violet}{\bf Server}
\EndIndent
\State{\textbf{end block}}

\State $S_t =\{0, \dots, n-1\}$
\If{\textit{subsampling}}
    \State $S_t = \textrm{Unif}^{(2f+1)}\left ( \{0, \dots, n-1\}\right)$\footnotemark
\EndIf

\State \textcolor{violet}{\bf Server} aggregates $\vec{C}_t = \aggregator{} \left ( \{\vec{c}_{t}^{(i)} \: | \: i \in S_t\} \right)$

\State \textcolor{violet}{\bf Server} sends $\vec{C}_t$ to all nodes
\Comment{as per Section~\ref{sec_algo_description}, \textbf{2)}}

\State{\textbf{\textup{Every}} \textcolor{teal}{\bf node} $i$ \textbf{\textup{does in parallel}}}
\Comment{as per Section~\ref{sec_algo_description}, \textbf{3)}}
\Indent
\State Compute $P_t$ via \textit{decrypt-unbatch-decode-regroup} on $\vec{C}_t$
\State Compute trimmed mean $TM_t = \frac{1}{|S_t| - 2f} P_t$
\State Locally update model $\theta_{t} = \theta_{t-1} - \gamma TM_t$

\EndIndent
\State{\textbf{end block}}

\EndFor
\State \textbf{Every} \textcolor{teal}{\bf node} $i$ returns $\theta_{T}$
\end{algorithmic}
\end{algorithm}
\footnotetext{$\textrm{Unif}^{(a)}\left (A \right )$ denotes the sampling of $a$ elements from the set $A$ uniformly at random and without replacement.}

%% file: 4_hts.tex
\section{Novel Homomorphic Operators}\label{sec_fhe_aggregators}
In this section, we introduce \aggregator{}, our homomorphic implementation of CWTM 
given in~\eqref{eq:coordinate-wise.sorted} and~\eqref{eq:cwtm} over BGV.
As previously mentioned, for efficiency purposes, we offload the division by the publicly known factor $(n-2f)$ at the end of CWTM 
to the nodes' side, and implement a slightly modified version of CWTM.
We also obtain the median operator \aggregatortwo{} as a special case.

\subsection{The BGV Cryptosystem}\label{subsec:bgv}
We recall the key elements from BGV used in our implementation
and refer the reader to \cite{BGV} for notation and a more thorough description of the algebraic structures involved in BGV.
The \defn{plaintext space} is the modular polynomial ring
$\mathbb{Z}_p[X]/\langle\Phi_m(X)\rangle$, where $p$ is the
\defn{plaintext modulus} and $\Phi_m(X)$ is the
\defn{cyclotomic~polynomial} of index $m$ with degree $\varphi(m)$ given by Euler's totient function.
An \defn{encryption} operator $\encrypt_\pk$, requiring a public key \pk ,
maps the plaintext space into the \defn{ciphertext space}
$\mathbb Z_q[X]/\langle\Phi_m(X)\rangle$, where $q$ is the \defn{ciphertext modulus}.
A \defn{decryption} operator $\decrypt_\sk$, requiring a private key \sk, maps the ciphertext back to the plaintext space.

BGV comes with a natural plaintext encoding
scheme, performed via the \emph{Number Theoretic Transform}
(NTT)~\cite{Pollard1971TheFF} and referred to as
\defn{batching}.
Batching takes place prior to encryption and encodes
a vector $\vslots\,=\,\left[\slot^{(0)}, \dots, \slot^{(\nslots-1)}\right]$\, of $\nslots$ polynomials $\slot^{(i)}$ of equal length $\ordp$, called \defn{slots}, into a plaintext polynomial \pt:
\begin{align}
  \batch[\vslots] = \pt
  \label{diag:batching}
\end{align}
Specifically, a slot \slot\ is an element of the modular
polynomial ring $\mathbb Z_p[X]\langle G(X)\rangle$,
where the polynomial $G(X)$ is a prime factor of $\Phi_m(X)$ over
$\mathbb{Z}_p$.  The degree $\ordp$ of $G(X)$ coincides with the multiplicative order of $p$ modulo $m$, namely the smallest $N>0$
such that $p^\ordp\equiv 1 \mod m$. Also, there holds  $\varphi(m)\,=\,\nslots\ordp$.
Batching requires that $p$ be a prime number not dividing $m$
while $m$ can be otherwise arbitrary.  Yet, for simplicity we choose
$m$ to be a prime number, as in this case the degree of $\Phi_m(X)$ is simply given by $\varphi(m)=m-1$.

The upshot is that the addition and multiplication of vectors are performed slot-wise via suitable operations on the associated plaintexts thanks to the Chinese Remainder Theorem.
Polynomial operations are thus parallelizable in an SIMD fashion.
\medskip

An operator \texttt{Op} on vectors $\vslots_i$ for $i=1,\,\dots,\,k$
is said to be \defn{homomorphic} 
if there exists an operator $\texttt{\textbf{Op}}_\pk$,
depending on the public key \pk\, and acting on ciphertexts $\ct_i$ for $i=1,\,\dots,\,k$ such that,
if 
\begin{equation}
  \ct_i = \encrypt_\pk[\batch[\vslots_i]],\qquad i=1,\,\dots,\,k
\end{equation}
then
\begin{align}
  \batch^{-1}[\decrypt_\sk[\texttt{\textbf{Op}}_\pk[\ct_1,\,\dots,\,\ct_k]]]
  = \texttt{Op}\left[\vslots_1,\,\dots,\,\vslots_k\right]
  \label{defn:eval.homomorphic}
\end{align}
and this homomorphic property is closed under composition.

Core homomorphic operators in BGV include slot-wise addition and
multiplication on vectors of slots,
denoted $\vslots+\vslots'$ and $\vslots\cdot\vslots'$ respectively,
where addition and multiplication on slots are the operations on
$\mathbb{Z}_p[X]/\langle G(X)\rangle$.
The same holds for slot-wise addition and multiplication with vectors of
cleartext polynomials in $\mathbb Z_p[X]/\langle G(X)\rangle$.
In turn, slot-wise polynomial operators are also homomorphic.
We shall also use the \defn{digit extraction} operators
$\extract_i$ indexed by the position $i=0,\,\dots,\,N-1$,
and defined by
\begin{equation}
  \extract_i[\vslots]
  := \left[
  \slot_i^{(0)},\,\dots,\,\slot_i^{(\nslots-1)}
  \right]
  \label{def:digit.extraction}
\end{equation}
where $\slot_i^{(j)}$ is the coefficient of degree $i$ of the polynomial representing slot $\slot^{(j)}$, respectively for $j=0,\,\dots,\,\nslots-1$.

\subsection{Plaintext Encoding Schemes}\label{subsec:plaintext.encoding.schemes}
A reasonable choice for the encoding of an integer $a$ into a slot
\slot\ is to expand it with respect to a \defn{base} $B$ and map it
to a slot in $\mathbb{Z}_p[X]/\langle G(X)\rangle$ according to
\begin{equation}
  a\,=\,\sum_{i=0}^{\ordp-1}c_iB^i
  \quad \mapsto \quad \slot \,=\,\sum_{i=0}^{\ordp-1}c_iX^i
  \label{eq:integer.digit.decomposition}
\end{equation}
where $0\,\le\,c_i\,<\,B$ are the unique digits of $a$ in base $B$.
The only requirements are that  $B\le p$ and $\ordp$
coincides with the number of coefficients in each slot (see the
discussion around \eqref{diag:batching} for notation).
Furthermore, $\ordp$ and $B$ must guarantee that $\{0, \dots, B^\ordp-1\}$ covers a given range of integer values.

A common inclination is to take $B\,=\,p$ in order to maximize the
storage capacity of a slot.
This is adequate in certain applications, for example a single
comparison in the bivariate approach \cite{TanEtAl, IZ21}.
Unfortunately, this precludes even a single addition of ciphertexts as
the resulting coefficients in a slot will typically overflow the range
$\{0,\,\dots,\,p-1\}$.
This problem arises precisely in the univariate approach of the comparison operators, where the \emph{difference} on the digits of the slots is tested for negativity.
To remedy this, \cite{ShaulFR20, NarumanchiGEG17, IZ21} take $B\,=\,(p-1)/2$. 
These implementations are adequate for plain comparison, min/max,
sorting, or one addition of two ciphertexts.
However, they cannot handle CWTM as it involves at least two additions after comparison as evidenced in~\eqref{eq:trimmed_sum}.

Another crucial and completely independent observation is that the
\emph{ciphertext} modulus $q$, not the \emph{plaintext} modulus $p$, directly impacts performance on core operations, in particular multiplications of ciphertexts.
In fact, $p$ has very limited impact there and should therefore be exploited towards improving the performance on compound operators, such as the comparison operator on ciphertexts \lessthan\ introduced in~\eqref{eq:homomorphic.LT} later in Section~\ref{subsec:homomorphic.trimmed.sum}.
We achieve this by decoupling the base $B$ in the integer expansion~\eqref{eq:integer.digit.decomposition} from the modulus $p$ of the plaintext space.
Indeed, with $m$ fixed, we choose the length $\ordp$ and base $B$ optimizing the \emph{amortized} performance: the smaller the $\ordp$, the more the slots to pack in a ciphertext and the fewer the digits to process (e.g., see~\eqref{eq:homomorphic.LT}). 
However, the smaller the $B$, the lower the complexity and multiplicative depth of the operators applied to these digits, such as \negative\ or \zero\ used in~\eqref{eq:homomorphic.LT}.
Unlike the existing solutions mentioned earlier, we are able to achieve any value of $\ordp$ of our choice precisely because we have decoupled $B$ from $p$.

\subsection{Building Blocks}
\label{subsec:homomorphic.building.blocks}
This section introduces low-level operators upon which 
our operators \aggregator{} and \aggregatortwo{} are built.
\medskip

\textbf{Terminology.}
All operators on vectors of slots that will be used throughout are
homomorphic and operate slot-wise.
For emphasis, this means that all slots are processed in parallel, independently of one another.
We recall that a slot in a generic vector \vslots\ is a
polynomial whose coefficients record the digits of an integer
represented in its base-$B$ expansion as in
\eqref{eq:integer.digit.decomposition}.
The digits can be individually recorded into new vectors of
slots thanks to the digit extraction operators $\extract_i$ for $i=0,\,\dots,\,N-1$,
see \eqref{def:digit.extraction}.
This results in vectors, all of whose slots are represented by constant polynomials, and we therefore speak of \defn{constant-polynomial vectors} and of the \defn{values} of their slots.

Finally, we define the function $\1$ on Boolean values by 
\begin{equation}
  \1\left(\texttt{true}\right)=0,\qquad
  \1\left(\texttt{false}\right)=1
\end{equation}
\hfill

\medskip
\textbf{Three auxiliary operators: \zero, \negative, and \btwn.}
These operators are obtained via Lagrange polynomial interpolation, and are 
thus homomorphic and operate slot-wise on their input.
\smallskip

Given integers $M_1\le M_2$, the \defn{zero-test} operator
$\zero[\cdot;\,M_1,\,M_2]$ is implemented as a 
polynomial operator of degree $M_2-M_1$.
If its input is a constant-polynomial vector \digit\ with values in $\{M_1,\,\dots,\,M_2\}$,
$\zero[\digit;\,M_1,\,M_2]$ is a constant-polynomial vector with values equal to $1$ where the corresponding values in \digit\ are zero, and $0$ otherwise.

The \defn{negativity-test} operator
$\negative[\cdot;\,M_1,\,M_2]$ is implemented as a 
polynomial operator of degree $M_2-M_1$.
If its input is a constant-polynomial vector \digit\ with values in $\{M_1,\,\dots,\,M_2\}$,
$\negative[\digit;\,M_1,\,M_2]$ is a constant-polynomial vector with values equal to $1$ where the corresponding values in \digit\ are negative, and $0$ otherwise.

Given furthermore that $\underline r\le \overline r$,
$\btwn[\cdot,\,\underline r,\,\overline r;\,M_1,\,M_2]$ is implemented as a polynomial operator of degree $M_2-M_1$.
If its input is a constant-polynomial vector $\texttt{r}$ with values in the range $\{M_1,\,\dots,\,M_2\}$, 
    $\btwn[\texttt{r},\,\underline r,\,\overline r;\,M_1,\,M_2]$
is a constant-polynomial vector with values equal to $1$ where the corresponding  values in \texttt{r} are between $\underline r$ and $\overline r$ (inclusive),
and $0$ otherwise.
\medskip

\textbf{The less-than operator \texttt{LT}.}
Two integers $a\,=\,\sum_{i=0}^{\ordp-1}c_iB^i$ and
$a'\,=\,\sum_{i=0}^{\ordp-1}c_i'B^i$ expanded according to
\eqref{eq:integer.digit.decomposition} can be compared via the
lexicographic ordering on their digits with the formula
\begin{equation}
  \1\left(a<a'\right) \,=\, 
  \sum_{i=0}^{\ordp-1} \Big(\1\left(c_i-c_i'<0\right)
  \times\prod_{j=i+1}^{\ordp-1}
  \1\left(c_j-c_j'==0\right) \Big)
  \label{eq:LT.lexicographic}
\end{equation}
Taking our cue from this identity, we introduce our \defn{less-than}
operator on vectors of slots as
\begin{align}
  \texttt{LT}[\vslots,\,\vslots';\,B] 
  := \,&
  \sum_{i=0}^{\ordp-1} \Big(
  \texttt{Neg}[\digit_i-\digit'_i;\,-B,\,B]
  \label{eq:homomorphic.LT}
  \\
  &\quad
  \times\prod_{j=i+1}^{\ordp-1}
  \texttt{Zero}[\digit_j-\digit'_j;\,-B,\,B]
  \Big)
  \notag
\end{align}
where for $i=0,\,\dots,\,N-1$ we use the short-hand notation
\begin{equation}
  \digit_i := \extract_i[\vslots]\qquad\textrm{and}\qquad
  \digit'_i:=\extract_i[\vslots']
\end{equation}
If the slots stored in \vslots\ and $\vslots'$ represent integers in
their base-$B$ expansion as in \eqref{eq:integer.digit.decomposition}, $\texttt{LT}[\vslots,\,\vslots';\,B]$
is a constant-polynomial vector with values equal to $1$ where the corresponding slot in \vslots\ represents an integer less than its counterpart from $\vslots'$, and $0$ otherwise.

By composition, the operator \lessthan\ is homomorphic and operates
slot-wise since this is true for \extract, \negative\, and \zero.
\medskip

\textbf{The operator $\rank$.}
Given an ordered list 
$a_0, \dots, a_{n-1}$ of integers,
we wish to compute their respective ranks
$\textrm{rk}_0,\,\dots,\,\textrm{rk}_{n-1}$ in the list,
namely a reordering of the range $\{0,\,\dots,\,n-1\}$ satisfying
\begin{equation}
  \begin{aligned}
    \textrm{rk}_i<\textrm{rk}_j   \quad & \textrm{if}\quad a_i<a_j\,,\quad \textrm{or if}\quad a_i\,==\,a_j\quad \textrm{and}\quad i\ne j\,.
  \end{aligned}
  \label{eq:rank.properties}
\end{equation}
This echoes \eqref{eq:coordinate-wise.sorted}.
For each $i$, the sum $\sum_{j=0}^{n-1}\1(a_i<a_j)$ returns
$\textrm{rk}_i$ only if the value of $a_i$ is not repeated in the list.
Following an idea from \cite{CetinEtAl}, this issue is overcome
by taking instead
\begin{equation}
  \textrm{rk}_i := \sum_{j\le i}\1(a_i<a_j) + \sum_{j>i}\1(a_i\le a_j)
\end{equation}

In order to implement this idea when constructing the \rank\ operator on
vectors of slots, we introduce the auxiliary operator
\begin{equation}
  \texttt{Comp}\left(\vslots,\,\vslots',\,\texttt{b};\,B\right)
  \,=\,
  \left\{
  \begin{array}{rl}
    \lessthan[\vslots,\,\vslots';\,B]&\textrm{if}\quad \texttt{b}\\
    1-\lessthan[\vslots',\,\vslots;\,B]&\textrm{if}\quad \texttt{not b}
  \end{array}
  \right.
  \label{eq:comp}
\end{equation}
where $\texttt{b}\in\{\texttt{true},\,\texttt{false}\}$.
Then, given an ordered list of vectors of slots
$\vslots_0,\,\dots,\,\vslots_{n-1}$,
our \rank\ operator is defined as
\begin{equation}
    \label{eq:rk}
    \rank[
      \{\vslots_j\}_{j=0}^{n-1},\,i;\,B
    ] 
    \,:=\, \sum_{j=0}^{n-1} 
    \texttt{Comp}[\vslots_i,\,\vslots_j,\,\1(i<j);\,B]
\end{equation}
which by composition is homomorphic and operates slot-wise.
If the slots in the vectors $\{\vslots_j\}_{j=0}^{n-1}$
represent integers in their base-$B$ expansion as in \eqref{eq:integer.digit.decomposition},
$\rank[\{\vslots_j\}_{j=0}^{n-1},\,i;\,B]$ is a constant-polynomial
vector.
The value of slot $k$ in this vector equals the rank (in the sense of \eqref{eq:rank.properties}) of the value of slot $k$ from $\vslots_i$ among the values of the slots of index $k$ from all the $\vslots_j$'s.
\medskip

\subsection{Homomorphic Trimmed Sum and Median}
\label{subsec:homomorphic.trimmed.sum}
\textbf{Our \aggregator{} operator.}
Echoing \eqref{eq:cwtm}, we introduce our homomorphic trimmed
sum operator for a base $B$ and a number $f$ of Byzantine nodes:
\begin{align}
  \label{eq:trimmed_sum}
  &\trimmedsum[\{\vslots_j\}_{j=0}^{n-1};\,f,\,B]\\
    &\quad\quad :=\sum_{i=0}^{n-1}
    \btwn[\texttt{rk}_i,\,f,\,n-f-1;\,0,\,n-1]\cdot\vslots_i    
\end{align}
using auxiliary operator \btwn\ and the short-hand notation
\begin{equation}
  \texttt{rk}_i := \rank[\{\vslots_j\}_{j=0}^{n-1},\,i;\,B]
\end{equation}
By composition,
\aggregator{} is  homomorphic and operates slot-wise.
If the slots in the vectors $\{\vslots_j\}_{j=0}^{n-1}$
represent integers in their base-$B$ expansion,
$\trimmedsum[\{\vslots_j\}_{j=0}^{n-1},\,f;\,B]$
is a vector containing in its slot $k$
the sum of the slots of rank between $f$ and $n-f-1$ among the slots
of index $k$ from all $\vslots_j$'s. 

\medskip
\textbf{Our \aggregatortwo{} operator.}
The homomorphic median operator is a special case of \aggregator{}
where only the value with rank $\lfloor\nicefrac{n}{2}\rfloor$ is
selected:
\begin{align}
  \label{eq:median}
  \median[\{\vslots_j\}_{j=0}^{n-1}\}]
  \,=\,
  \trimmedsum[\{\vslots_j\}_{j=0}^{n-1};\,\lfloor n/2\rfloor,\,B)]
\end{align}

%% file: 5_experiments.tex
\section{Empirical Evaluation}\label{sec_exp}
In this section, we empirically evaluate the performance of \algoname{} on standard image classification tasks, in a simulated distributed cross-silo system, and in several adversarial scenarios.
In Section~\ref{sec_exp_setup}, we discuss the experimental setup including the datasets and models used.
In Section~\ref{sec_exp_results}, we showcase the empirical performance of \algoname{} in terms of learning accuracy, showing that \algoname{} induces a negligible accuracy loss compared to its non-private counterpart.
In Section~\ref{sec_exp_he_perf}, we present the computational cost of \aggregator{} (and \aggregatortwo{}), and discuss how node subsampling can be leveraged to significantly decrease the cost of our solution in practice.
Finally, we discuss in Section~\ref{subsec:accelerating.homomorphic.sorting} the gain in performance that \aggregator{} yields compared to the state-of-the-art algorithm for homomorphic sorting.

\subsection{Experimental Setup}\label{sec_exp_setup}

\textbf{Datasets and data heterogeneity.}
We consider three image classification datasets, namely MNIST~\cite{mnist}, Fashion-MNIST~\cite{fashion-mnist}, and CIFAR-10~\cite{cifar}.
It is commonly accepted that these datasets are not very challenging learning problems.
However, in the context of Byzantine ML, most relevant previous works do not go beyond these simple tasks~\cite{RESAM, allouah2023fixing, karimireddy2022byzantinerobust, Karimireddy2021History, allouah2023trilemna, farhadkhani23a}, due to the inherent difficulty of the problem at hand.
While learning on MNIST and CIFAR-10 is straightforward in the typical honest setting, defending against a fraction of Byzantine nodes while also providing privacy safeguards makes the problem significantly more difficult.
In order to simulate a realistic cross-silo environment where nodes could have different data distributions, we split the datasets across nodes in a heterogeneous fashion.
We make the different nodes sample from the original datasets using a Dirichlet distribution of parameter $\alpha$, as proposed in~\cite{dirichlet}.
The smaller the $\alpha$, the more heterogeneous the setting.
For MNIST, we set $\alpha = 1$, while for the more challenging Fashion-MNIST dataset, we fix $\alpha = 5$.
A pictorial representation of the induced heterogeneity depending on $\alpha$
can be found in Appendix~\ref{app_exp_setup_dist}.
For CIFAR-10, we consider a homogeneous distribution due to the intrinsic difficulty of the task, and split the dataset across nodes uniformly at random.
\medskip

\textbf{Distributed system and Byzantine regimes.}
We consider a cross-silo distributed system composed of an honest-but-curious server and $n$ nodes.
We execute \algoname{} in two settings, namely $n = 15$ for MNIST and Fashion-MNIST and $n = 9$ for CIFAR-10.
We evaluate our algorithm in Byzantine regimes of increasing strength by varying the number of Byzantine nodes present in the system.
Specifically, we choose $f \in \{1, \dots, 7\}$ on MNIST and Fashion-MNIST, and $f \in \{1, \dots, 4\}$ on CIFAR-10.
\medskip

\textbf{Models and hyperparameters.}
On MNIST, we train a feed-forward neural network of $d = 79510$ parameters composed of two fully connected linear layers (model 1); while on Fashion-MNIST a convolutional neural network (CNN) of $d = 431080$ parameters (model 2) is used.
On both datasets, we run \algoname{} for $T = 1000$ steps, using a batch-size $b = 25$, a fixed learning rate $\gamma = 0.5$ for model 1 and 0.1 for model 2, momentum parameter $\beta = 0.99$, and clamp parameter $C = 0.001$.
Moreover, the negative log likelihood loss is used as well as l2-regularization of $10^{-4}$.
Furthermore, we consider a bit precision of $\delta = 2$ bits and 3 bits on MNIST and Fashion-MNIST, respectively.
On the most challenging task, i.e., CIFAR-10, we train an even larger CNN of $d = 712854$ parameters (model 3) for $T = 2000$ steps, with $b = 50$, $\gamma = 0.5$, $\beta = 0.99$, and $C = 0.004$.
We consider a bit precision of $\delta = 4$ bits, and the cross entropy loss is used to train model 3.
The values of these hyperparameters ($b$, $\beta$, $\gamma$, regularization) are chosen to be coherent with the Byzantine ML literature~\cite{RESAM, allouah2023fixing, karimireddy2022byzantinerobust, Karimireddy2021History}.
The detailed architectures of the models are presented in~\ref{app_sec_model_arch}.

\medskip
\textbf{Benchmarking and performance evaluation.} We evaluate the learning performance of \algoname{} compared to its non-private counterpart Robust-DSGD,\footnote{We execute Robust-DSGD using CWTM as robust aggregator (see Section~\ref{sec_robust_DSGD})} and its non-robust counterpart HE-DSGD.\footnote{Equivalent to running DSGD in the encrypted domain using homomorphic averaging.}
Our objective is to assess the \textit{privacy} and \textit{robustness} costs of our algorithm in terms of learning compared to Robust-DSGD and HE-DSGD, respectively.
In other words, we are interested in how much the accuracy of the model drops when upgrading Robust-DSGD with privacy safeguards (by quantizing and encrypting momentums), and when granting robustness to HE-DSGD.
To do so, we execute \algoname{} in the optimistic setting where there are no Byzantine nodes in the system (i.e., the algorithm still expects $f$ nodes to be Byzantine, but all nodes behave correctly in the execution). We denote this execution by \algoname{}-0 Byzantine (\textit{\algoname{}-0B}), and compare its performance to HE-DSGD. As a benchmark, we also implement the standard DSGD algorithm.
Finally, we run \algoname{} when node subsampling is enabled, and refer to it as \textit{\algoname{}-SUB}.
\medskip

\begin{figure*}[!ht]
    \centering
    \includegraphics[width=0.4\textwidth]{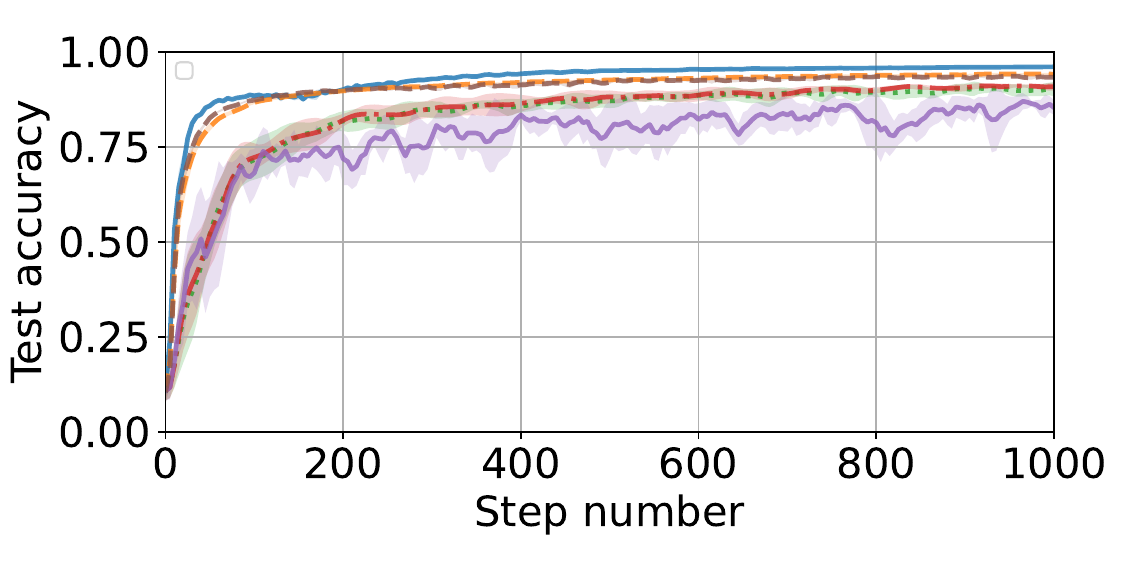}
    \includegraphics[width=0.4\textwidth]{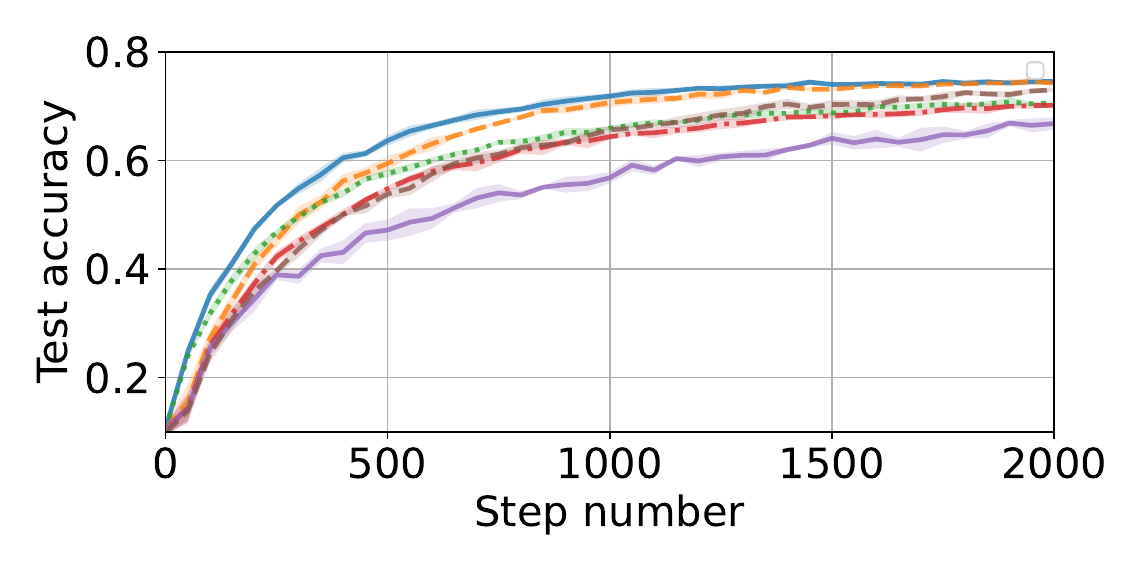}\\
    \vspace{-2mm}
    \includegraphics[width=0.4\textwidth]{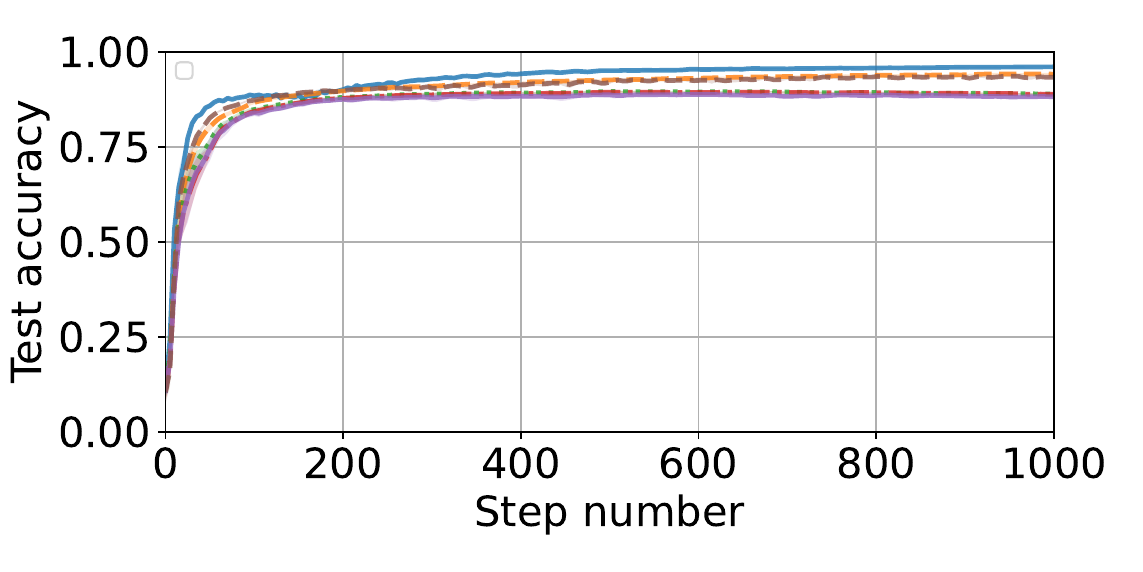}
    \includegraphics[width=0.4\textwidth]{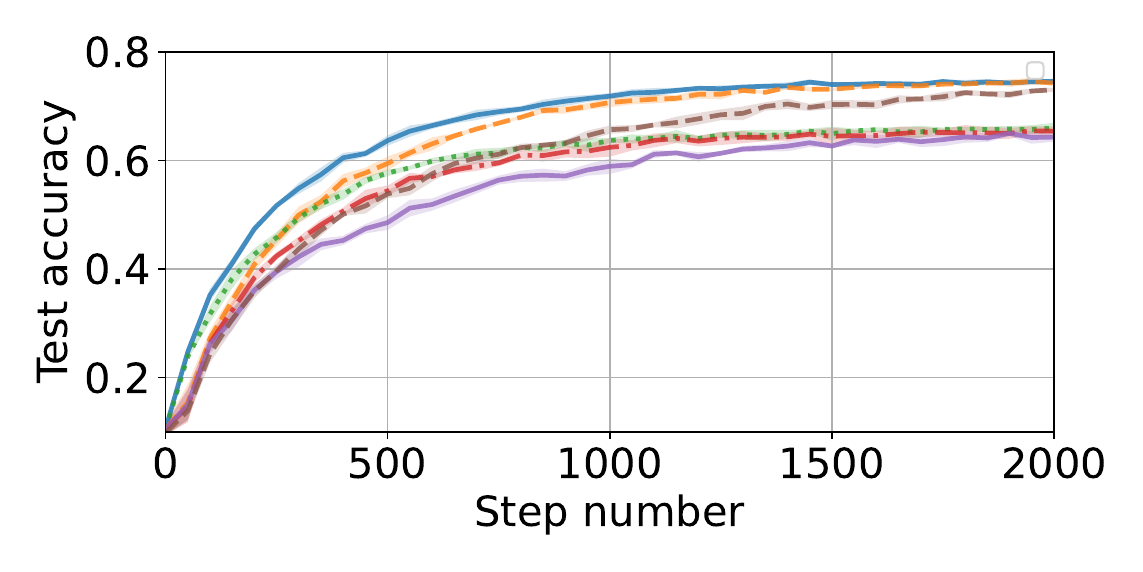}\\
    \vspace{-2mm}
    \includegraphics[width=0.4\textwidth]{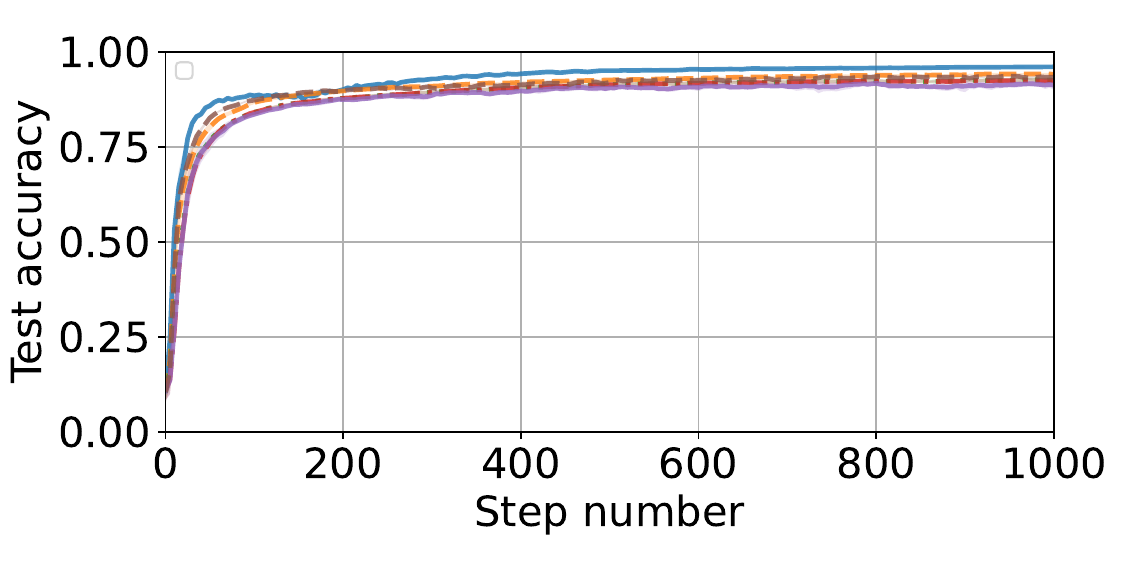}
    \includegraphics[width=0.4\textwidth]{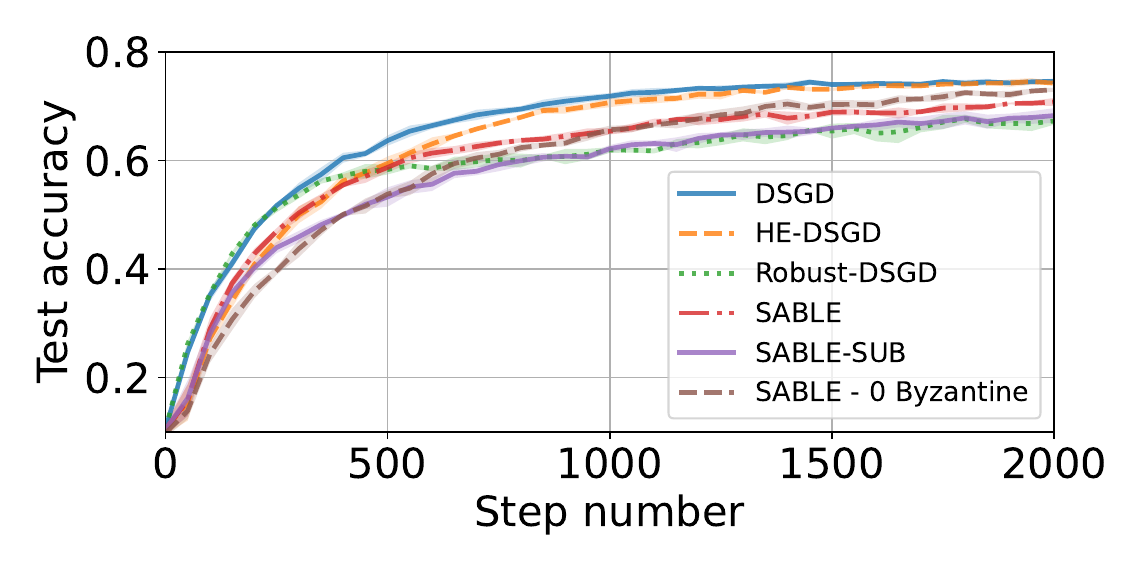}%
    \vspace{-2mm}
    \caption{\justifying \textit{Column 1}: MNIST with $f=5$ and $n = 15$. \textit{Column 2}: CIFAR-10 with $f=2$ and $n = 9$. The Byzantine nodes execute the FOE (\textit{row 1}), ALIE (\textit{row 2}), and LF (\textit{row 3}) attacks. The plot of the mimic attack on MNIST can be found in Appendix~\ref{app_exp_results}. \algoname{} consistently matches the performance of its non-private counterpart Robust-DSGD, and subsampling works well empirically.}
\label{fig:plots_main}
\end{figure*}

\textbf{Byzantine attacks.}
In our experiments, the Byzantine nodes execute four state-of-the-art attacks, namely label flipping (LF)~\cite{allen2020byzantine}, mimic~\cite{karimireddy2022byzantinerobust}, a little is enough (ALIE)~\cite{ALIE}, and fall of empires (FOE)~\cite{FOE}, detailed below.
Let $\overline{v}_t$ be the average of the vectors sent by the honest nodes in step $t$.

\begin{enumerate}
    \item \textbf{LF:} Byzantine nodes perform a label flip/rotation by replacing every label $l$ in their datasets by $9 - l$, since the labels of the considered datasets are in $\{0, \dots, 9\}$. Their gradients are thus computed on flipped labels.
    \item \textbf{Mimic:} In heterogeneous settings, Byzantine nodes mimic honest nodes by sending their vectors.
    We implement the heuristic in~\cite{karimireddy2022byzantinerobust} to determine the optimal node to mimic in every step $t$.
    \item \textbf{FOE:} Byzantine nodes send $(1 - \tau) \overline{v}_t$ in step $t$, where $\tau \in \mathbb{R}$ represents the attack factor.
    Note that executing FOE with $\tau = 2$ is equivalent to the well-known sign flipping attack~\cite{allen2020byzantine}.
    \item \textbf{ALIE:} Byzantine nodes send $\overline{v}_t + \tau . \sigma_t$ in step $t$, where $\sigma_t$ is the coordinate-wise standard deviation of the honest vectors and $\tau \in \mathbb{R}$ represents the attack factor.
\end{enumerate}
In our experiments, we implement optimized versions of ALIE and FOE.
The attack factor $\tau$ is determined using a greedy algorithm that performs a linear search over a predefined range of values.
In every step $t$, the optimal $\tau$ is determined by choosing the value maximising the $L2$-distance between $\overline{v}_t$ and the output of the aggregation. 
\medskip

\textbf{Code details.}
We use the standard Python library for deep learning PyTorch~\cite{pytorch} to run all our ML experiments.
We execute DSGD and Robust-DSGD using PyTorch's default 32-bit precision.
HE-DSGD is run with the same bit precision as \algoname{}.
The ML experiments are executed using seeds 1 to 5.
Experiments evaluating the computational time of \aggregator{} are executed on the same machine with a 12th Gen Intel© Core™ i7-12700H × 14 processor and $64$GB of RAM.
We simulate the distributed system using one machine, and neglect network latency in our experiments since it does not impact neither accuracy nor compute time.
Our code can be found on the private Bitbucket repository {\url{https://bitbucket.org/he-sar-dsgd/workspace/overview/}}. To log in, use the username "he.sar.dsgd@gmail.com" with the password "sable-code-asiaccs-2024".
We also plan to make the code public.

\subsection{ML Performance of \algoname{}}\label{sec_exp_results}
In this section, we evaluate the performance of \algoname{} in terms of learning accuracy.
We consider three datasets of increasing difficulty, namely MNIST, Fashion-MNIST, and CIFAR-10, on which we train three models of increasing complexity.
The empirical results on MNIST with $f = 5$ and CIFAR-10 with $f = 2$ are displayed in Figure~\ref{fig:plots_main}.
We present the remaining plots in Appendix~\ref{app_exp_results}.
Due to space limitations, we are unable to include all experiments for the remaining values of $f$ we have tested.
Therefore, we show in Appendix~\ref{app_exp_results} the results for $f \in \{3, \dots, 7\}$ on MNIST, $f \in \{3, 5, 7\}$ on Fashion-MNIST, and $f \in \{3, 4\}$ on CIFAR-10.
The missing plots convey the same observations made hereafter.

\medskip
\textbf{MNIST.}\label{sec_exp_results_mnist}
First, our algorithm manifests strong robustness to Byzantine nodes, consistently against all four Byzantine attacks (see also Figure~\ref{fig:plots_mnist_f=5_hetero} in Appendix~\ref{app_exp_results}).
Second, despite using a much smaller bit precision (2 vs 32 bits), \algoname{} matches the learning performance of its non-private (and full-precision) counterpart Robust-DSGD, while preserving the privacy of the nodes.
This highlights the practical relevance of our solution, and suggests that its privacy cost compared to Robust-DSGD is low in most settings.

We also observe from Figure~\ref{fig:plots_main} that the robustness cost of our algorithm is also negligible.
Indeed, despite the incorporation of \aggregator{} instead of averaging, \algoname{}-0B is still able to match the performance of HE-DSGD (and even DSGD) in executions where all nodes are correct, while simultaneously protecting against five Byzantine nodes.
The slight difference between the performances of DSGD and HE-DSGD is only due to the much smaller bit precision $\delta = 2$ used by the latter algorithm. 
Finally, we note that subsampling is a solution that works well empirically.
Indeed, despite taking into account fewer vectors, \algoname{}-SUB still showcases similar accuracies to the original algorithm while aggregating less ciphertexts.

\medskip
\textbf{CIFAR-10.}~\label{sec_exp_results_cifar}
We first notice a larger discrepancy between the performances of \algoname{}-0B and HE-DSGD on CIFAR-10 compared to MNIST.
Despite its slower convergence rate compared to HE-DSGD, \algoname{}-0B is still able to reach a final accuracy close to that of HE-DSGD, indicating that the robustness cost of our algorithm is not large.
The robustness cost of \algoname{} is higher in this case (compared to MNIST), due to the intrinsic difficulty of the task at hand.
However, this cost is inevitable in order to be able to successfully defend against Byzantine nodes in the system.

Furthermore, although \algoname{} uses a small bit precision of only $\delta = 4$ bits, our algorithm still matches the performance of its full-precision counterpart, namely Robust-DSGD, consistently against all three Byzantine attacks.
This highlights the low privacy cost of our solution resulting from the encryption of gradients, and confirms the relevance of \algoname{} in practical systems.

An interesting observation is that \algoname{} even outperforms Robust-DSGD under the LF attack.
We conjecture that this is due to the quantization of momentums prior to encryption in \algoname{}.
In fact, quantization could help in reducing the intrinsic variance present in the momentums of honest nodes; thus bringing honest vectors closer together, and potentially making it easier for the robust aggregation to filter out Byzantine values.
Quantization also decreases the attack space of Byzantine nodes which would have a reduced set of malicious vectors to choose from.
Finally, node subsampling is effective under attack as well, as Figure~\ref{fig:plots_main} shows that \algoname{}-SUB almost matches the original algorithm.

\subsection{Execution Time of \aggregator{}}\label{sec_exp_he_perf}
In this section, we evaluate the computational performance of \aggregator{} when executing \algoname{} with (Table~\ref{table:he-times-with-subsampling}) and without (Table~\ref{table:he-times-no-subsampling}) node subsampling.
We consider four different hardware models for the server, and quantify its computational power using the number of cores available to be used concurrently.
We consider in our experiments machines having 1, 16, 32, and 64 cores.
As previously mentioned, since \aggregator{} is a coordinate-wise operator, the different cores can be used in parallel to execute \aggregator{} on separate and independent blocks of coordinates, accelerating the computations. In all experiments, we guarantee an HE security level $\lambda \geq 128$.


\input{table_no_subsampling}

\medskip
\textbf{More cores, less time.}
Table~\ref{table:he-times-no-subsampling} shows that using more powerful machines directly translates into lower execution times.
Indeed, the time required by 1 core to execute \aggregator{} on $n = 15$ ciphertexts on MNIST is 17.34 minutes.
This time is considerably reduced $14\times$ to reach 1.24 minutes, when at least 16 cores are used instead.
Note that these practical execution times are also imputable to the lower precisions considered ($\delta = 2, 3, 4$ bits, depending on the dataset). As discussed in Section~\ref{sec_exp_results}, despite this decrease in precision, \algoname{} still matches the ML accuracy of its full-precision counterpart Robust-DSGD.
Moreover, we can observe that more complex learning tasks typically need more powerful machines in order to compute \aggregator{}.
Indeed, Table~\ref{table:he-times-no-subsampling} shows that a 16-core machine is required in order to efficiently perform \aggregator{} on MNIST in 1.24 minutes.
On Fashion-MNIST, a server machine with at least 32 cores is needed in order to obtain a feasible time (5.81 minutes); given that $n = 15$ and the model trained has $d=431080$ parameters.
Finally, on CIFAR-10, despite the smaller number of nodes considered, the challenging nature of the task (i.e., the large model size) requires 32 cores to guarantee a reasonable time below 5 minutes.
A server with 64 cores halves that time to obtain 2.02 minutes.

\medskip
\textbf{Parallelization limits.}
As indicated by MNIST in Table~\ref{table:he-times-no-subsampling}, there is a point beyond which adding more cores no longer decreases the aggregation time.
The reason is that there is a base homomorphic computation that is done over data encrypted into one ciphertext.
If the model size is large enough, more than one ciphertext is needed, and we can then parallelize trivially over those additional ciphertexts.
Therefore, our gains stop when the number of cores reaches the number of ciphertexts used.
\medskip


\input{table_subsampling_new}

\textbf{Subsampling considerably decreases the computational cost.}
Node subsampling significantly accelerates \aggregator{} on all three learning tasks.
Looking at Table~\ref{table:he-times-no-subsampling}, we can see that using a 64-core machine is the only practically relevant alternative to train a model of 712854 parameters on CIFAR-10.
The corresponding execution time is 2.02 minutes.
Downgrading the server
directly leads to a linear increase in the computational time by 2$\times$, 4$\times$, and 64$\times$ when using 32, 16, and one cores, respectively.
Interestingly, node subsampling enables \aggregator{} to reach almost the same computational performance using fourth and half of the resources when $f = 2$ and 3, respectively.
Indeed, 16 cores are able to reach 2.38 minutes per aggregation when $f = 2$, and 32 cores are enough to obtain 2.70 minutes when $f = 3$.
Contrastingly, 64 cores are required to break the 2 minute mark without node subsampling (Table~\ref{table:he-times-no-subsampling}).
Furthermore, applying node subsampling on 64 cores leads to a time acceleration of 3.11$\times$ and 1.50$\times$ when $f = 2$ and 3, respectively.
The same analysis is applicable for the two remaining datasets.
Moreover, an interesting observation on MNIST is that the time for one homomorphic aggregation without subsampling and using one core is 17.34 minutes, which is really impractical.
However, subsampling accelerates \aggregator{} by $6.64\times$ when $f = 3$, resulting in the reasonable 2.61 minutes using only one core.
This highlights the practicality of subsampling by allowing us to opt for cheaper machines when performing homomorphic aggregations.

\medskip
\textbf{\aggregatortwo{}.}
As seen from \eqref{eq:median}, \aggregatortwo{} is computed as a specific sub-case of \aggregator{}. When evaluating the performance of HMED, we find no discernible difference with the timings presented in Tables~\ref{table:he-times-no-subsampling} and~\ref{table:he-times-with-subsampling}.
Since \aggregatortwo{} boils down to using a different polynomial evaluation at the last step of \aggregator{}, no additional cost is incurred.

\subsection{Accelerating Homomorphic Sorting}
\label{subsec:accelerating.homomorphic.sorting}
Both our operators \aggregator{} and \aggregatortwo{} are implemented through the means of our novel encoding method.
To demonstrate the efficacy of this method in general, we present here results for homomorphic sorting (as there is no other homomorphic trimmed mean operator in the literature).
Indeed, we illustrate the gain in performance of our technique for the task of homomorphic sorting compared to the current state-of-the-art algorithm~\cite{IZ21}, see Table~\ref{table:comparison-with-IZ}.
The key insight of \cite{IZ21} is that by taking $B=(p-1)/2$, roughly half of the coefficients in a certain polynomial operator vanish, and they are able to improve on prior work~\cite{TanEtAl}.
However, these cancellations occur only for this specific choice of parameters, and force a larger value for $B$ than is necessary because there is no freedom to choose $B$ and $p$ independently.
Results shown in Table~\ref{table:comparison-with-IZ} demonstrate that, despite this noteworthy insight from \cite{IZ21} that may prove advantageous in certain specific situations, our technique outperforms \cite{IZ21} in the general case.

Refer to Section~\ref{subsec:bgv} and Appendix~\ref{sec:notation} for notation.
For the purpose of illustration, we have chosen to present evaluations
of coordinate-wise sorting over $4$ vectors of dimension $\nslots=9352$ and $5764$,
and where each coordinate contains an $8$-bit value.
This choice of values is justified below.
In every setting, $q$ is optimized to be as small as possible while
still ensuring correctness.
The security level $\lambda$ is evaluated in bits and obtained using the latest commit of the lattice estimator~\cite{albrecht}.\footnote{\url{https://lattice-estimator.readthedocs.io/en/latest/readme_link.html}}

\begin{table}[ht]
\centering
\caption{ \justifying Speed-up of our method compared to \cite{IZ21} when evaluating a coordinate-wise sorting over $\boldsymbol{4}$ vectors with $\boldsymbol{8}$-bit coordinates. The dimension is $\boldsymbol{\nslots=9352}$ and 5764 for $\boldsymbol{m=28057}$ and $\boldsymbol{17293}$ respectively. For each method, we report the optimum $\boldsymbol{q}$ (in bits). The amortized (Am.) time (in ms) and size (in Bytes) per coordinate are also shown. We display the amortized time as it enables a comparison that is independent of the dimension of the vectors. The time and size improvements ($\boldsymbol{\times}$) with respect to the first row are also reported. The red value \textcolor{red}{106} highlights a weak bit security ($\boldsymbol{\lambda < 128}$).}
\label{table:comparison-with-IZ}
\begin{tabular}{| c | c | c | c | c | c | c |}
\hline
    Algo & $m$ & $p$ & $q$  & $\lambda$ & Am. Time  & Am. Size\\
    \hline
    \hline
    \cite{IZ21} & $28057$ & $167$ & $410$ & $166$ & $5.6$ & $308$\\
    \hline
    \hline
    Ours & $28057$ & $167$ & $300$ & $238$ & $4.8$ ($\boldsymbol{1.2{\times}})$ & $225$ ($\boldsymbol{1.4\times}$) \\
    \hline
    \cite{IZ21} & $17293$ & $131$ & $360$ & \color{red} $106$ & $4.6$ ($\boldsymbol{1.2\times}$) & $270$ ($\boldsymbol{1.1\times}$)\\
    \hline
    Ours & $17293$ & $131$ & $280$ & $152$ & $3.8$ ($\boldsymbol{1.5\times}$) & $210$ ($\boldsymbol{1.5\times}$) \\
    \hline
  \end{tabular}
\end{table}

The first row of the table shows the results of our implementation of \cite{IZ21}'s sorting algorithm using the parameters given in their paper ($m=28057$ and $p=167$, for which $\nslots=9352$ and $\ordp=3$, see~\cite[Table~3]{IZ21}), which serves as a baseline.
Using the same parameters $m$ and $p$ (and thus the same $\nslots$ and $\ordp$) but a significantly smaller $B = 7$, the second row shows how our method significantly decreases $q$ thanks to the reduced multiplicative depth of our algorithm. This results in a higher security level $\lambda = 236$ bits for our solution, and a 1.2$\times$ gain in computational performance over the other method.

Since our method provides a substantial increase in the security level, in the fourth row we use a lower value for $m$ in order to make the security level closer to the standard $128$ bits. We take $m\,=\,17293$ and $p\,=\,131$, for which $\nslots=5764$ and $\ordp=3$, 
thus achieving a security level similar to the baseline (row 1), and with $B\,=\,7$ we gain a $1.5\times$ speed-up in computational time. As an added bonus, the smaller value for $q$ that the flexibility of our method incurs also leads to smaller ciphertexts. Indeed, our ciphertexts are $1.5\times$ smaller in size than the ciphertexts in previous works.

The third row shows the results of using the same parameters $m$ and $p$ as in the fourth row but with \cite{IZ21}'s method.
Their resulting security level $\lambda\,=\,106$ dips below the $128$-bit security standard.


%% file: table_no_subsampling.tex
\begin{table}[ht]
\centering
\caption{ \justifying Aggregation time \textit{without} node subsampling. This table presents the time (in min) needed for \textit{one} evaluation of \aggregator{} over $\boldsymbol{n}$ ciphertexts. The bit precision $\boldsymbol{\delta}$ is also reported.
We present the total bandwidth (BW) in KBytes needed per node in one iteration. We guarantee a bit security $\boldsymbol{\lambda \geq 128}$.}
\label{table:he-times-no-subsampling}
\begin{tabular}{| c |c | c | c | c | c | c |}
    \hline
    Dataset & Model size $d$ & $\delta$ & $n$ & Cores & Time & BW \\
    \hline
    \hline
    \multirow{2}{*}{MNIST} & \multirow{2}{*}{$79510$} & \multirow{2}{*}{$2$} & \multirow{2}{*}{$15$} & $1$ & $17.34$ & \multirow{2}{*}{$5$} \\
    \cline{5-6}
    & & & & $\geq 16$ & $1.24$ & \\
    \hline
    \hline
    \multirow{4}{*}{Fashion} & \multirow{4}{*}{$431080$} & \multirow{4}{*}{$3$} & \multirow{4}{*}{$15$} & $1$ & $145.13$ & \multirow{4}{*}{$27$}\\
    \cline{5-6}
    & & & & $16$ & $9.68$ & \\
    \cline{5-6}
    & & & & $32$ & $5.81$ & \\
    \cline{5-6}
    & & & & 64 & 3.87 & \\
    \hline
    \hline
    \multirow{4}{*}{CIFAR} & \multirow{4}{*}{$712854$} & \multirow{4}{*}{$4$} & \multirow{4}{*}{$9$} & $1$ & $125.16$ & \multirow{4}{*}{$88$} \\
    \cline{5-6}
    & & & & $16$ & $8.07$ & \\
    \cline{5-6}
    & & & & $32$ & $4.04$ & \\
    \cline{5-6}
    & & & & $64$ & $2.02$ & \\
    \hline
\end{tabular}

\end{table}

%% file: table_subsampling_new.tex
\begin{table}[ht]
\centering
\caption{ \justifying Aggregation time \textit{with} node subsampling. This table presents the time (in min) needed for \textit{one} evaluation of \aggregator{} over $\boldsymbol{2f+1}$ ciphertexts.
The acceleration "Acc." with respect to Table~\ref{table:he-times-no-subsampling} is also reported. We guarantee a bit security $\boldsymbol{\lambda \geq 128}$.
}
\label{table:he-times-with-subsampling}
\begin{tabular}{| c |c | c | c | c | c |}
    \hline
    Dataset & Model size $d$ & $f$ & Cores & Time  &  Acc. ($\times$)\\
    \hline
    \hline
    \multirow{6}{*}{MNIST} & \multirow{6}{*}{$79510$} & \multirow{3}{*}{$3$} & $1$ & $2.61$ & 6.64\\
    \cline{4-6}
    & & & $16$ & $0.29$ & 4.28 \\
    \cline{4-6}
    & & & $\geq32$ & $0.14$ & 8.86 \\
    \cline{3-6}
    & & \multirow{3}{*}{$5$} & $1$ & $9.15$ & 1.90 \\
    \cline{4-6}
    & & & $16$ & $0.65$ & 1.91 \\
    \cline{4-6}
    & & & $\geq 32$ & $0.65$ & 1.91 \\
    \hline
    \hline
    \multirow{8}{*}{Fashion} & \multirow{8}{*}{$431080$} & \multirow{4}{*}{$3$} & $1$ & $29.48$ & 4.92 \\
    \cline{4-6}
    & & & $16$ & $1.97$ & 4.91 \\
    \cline{4-6}
    & & & $32$ & $1.18$ & 4.92 \\
    \cline{4-6}
    & & & $64$ & $0.79$ & 4.90  \\
    \cline{3-6}
    & & \multirow{4}{*}{$5$} & $1$ & $79.94$ & 1.82\\
    \cline{4-6}
    & & & $16$ & $5.33$ & 1.82 \\
    \cline{4-6}
    & & & $32$ & $3.20$ & 1.82 \\
    \cline{4-6}
    & & & $64$ & $2.13$ & 1.82 \\
    \hline
    \hline
    \multirow{8}{*}{CIFAR} & \multirow{8}{*}{$712854$} & \multirow{4}{*}{$2$} & $1$ & $37.84$ & 3.31\\
    \cline{4-6}
    & & & $16$ & $2.38$ & 3.39 \\
    \cline{4-6}
    & & & $32$ & $1.30$ & 3.11 \\
    \cline{4-6}
    & & & $64$ & $0.65$ & 3.11 \\
    \cline{3-6}
    & & \multirow{4}{*}{$3$} & $1$ & $83.57$ & 1.50 \\
    \cline{4-6}
    & & & $16$ & $5.39$ & 1.50 \\
    \cline{4-6}
    & & & $32$ & $2.70$ & 1.50 \\
    \cline{4-6}
    & & & $64$ & $1.35$ & 1.50 \\
    \hline
\end{tabular}

\end{table}

%% file: 6_related_work.tex
\section{Related Work and Conclusion}\label{sec_conclusion}
In the past, efforts to address data privacy and Byzantine robustness in distributed learning have mainly progressed independently.
Notably, several works~\cite{phong2018, zhang2020, hybrid_dp, ma2022, sok23} explored the use of HE to enhance privacy in distributed learning when all nodes are honest.

Secure multi-party computation (MPC) was used in \cite{Bonawitz17} to make federated averaging privacy-preserving in the absence of Byzantine nodes.
Yet, the proposed approach does not take into account node misbehavior that occurs often in practical systems.
On the other hand, several Byzantine resilient algorithms have been proposed for distributed learning with meaningful convergence guarantees~\cite{RESAM, allouah2023fixing, karimireddy2022byzantinerobust, farhadkhani23a}.
Despite their advancements in addressing Byzantine threats, these works neglect the major privacy risks that ensue from sending the gradients to the server~\cite{DLG, IDLG, InvertingGradients}, unlike \algoname{}.

There have recently been attempts to simultaneously tackle the notions of privacy and Byzantine robustness in distributed learning. For example, \cite{DP_PODC, zhu2022bridging, allouah2023trilemna} investigate the use of differential privacy in Byzantine resilient learning algorithms to protect the privacy of the nodes against the server. These techniques usually rely on injecting noise into the gradients of the nodes. While these schemes provide strong privacy guarantees, it is usually at the expense of accuracy as noise injection introduces a fundamental trade-off between accuracy and privacy, when defending against Byzantine nodes~\cite{allouah2023trilemna}.
In Appendix~\ref{app_exp_results_DP}, we showcase the superior performance of \algoname{} in terms of accuracy compared to differentially-private Byzantine robust algorithms.

Hardware-based solutions (e.g., Intel SGX) have been used to prevent the server from seeing the data~\cite{sear, FLATEE}. However, these solutions rely on trusted dedicated hardware and are therefore very different in spirit from the cryptography-based and thus software-only approaches such as the one investigated in this paper.

Other works~\cite{hao21, he2020secure, prio, velicheti2021secure, brea, mlguard, lsfl, FLGuard, chowdhury22EIFFeL} make use of secure MPC to protect the privacy of the nodes.
Some~\cite{hao21, he2020secure, prio, FLGuard, mlguard, lsfl} assume the presence of two non-colluding honest-but-curious servers in the system, which is notoriously known for being a much stronger assumption than the single server setting that we consider.
Furthermore, the approaches proposed in~\cite{brea, he2020secure} do not protect the pairwise distances between model updates of honest nodes, that are leaked to the servers, hence constituting a significant privacy breach.
Additionally, some works~\cite{hao21, velicheti2021secure, chowdhury22EIFFeL} rely on weaker notions of Byzantine robustness, significantly weakening the convergence guarantees of the proposed solutions.
For example, the algorithm of \cite{hao21} considers a Byzantine robustness model based on FLTrust~\cite{FLTrust}, necessitating the server to possess a clean trusted dataset.
The algorithm relies on the strong assumption that the server has access to additional knowledge to filter out Byzantine nodes, as opposed to our approach which solely relies on the nodes' gradients.
While the solution presented in~\cite{chowdhury22EIFFeL} works well in practice, it does not offer any convergence guarantees which are crucial in Byzantine ML~\cite{karimireddy2022byzantinerobust, RESAM, allouah2023fixing, Karimireddy2021History}, thus limiting its applicability and theoretical relevance.
Moreover, the Byzantine resilient approach proposed in~\cite{velicheti2021secure} tolerates a smaller number of Byzantine nodes in the system and makes non-standard ML assumptions (e.g., bounded node updates).

Finally, a handful of prior works~\cite{wang21, ma21, PBFL, rahulamathavan2023fhefl} investigate the use of HE to grant privacy to Byzantine robust ML algorithms.
The previous work~\cite{wang21} closest to \algoname{} attempts to implement the Multi-Krum~\cite{krum} aggregator using the additively homomorphic Paillier cryptosystem. 
Their approach however does not fully implement Multi-Krum in the encrypted domain in order to fit the constraint of that cryptosystem. In turn, this solution induces significant leakage towards the aggregation nodes (equivalent to the server in our setting), hence significantly diminishing the privacy of the solution.
Additionally, in the solution proposed in~\cite{rahulamathavan2023fhefl}, the server has access to the resulting model at every step of the learning procedure, also endangering the privacy of the nodes.

A couple of approaches combine HE with weaker notions of Byzantine robustness \cite{PBFL, ma21}. 
Some \cite{PBFL} use the FLTrust robustness model (mentioned above), while others \cite{ma21} homomorphically implement Robust Stochastic Aggregation (RSA)~\cite{rsa}, a robust variant of SGD.
Yet, RSA has weak theoretical guarantees: it is only proven for strongly convex functions (which is seldom true for modern neural networks), and it provides sub-optimal guarantees as it does not employ noise reduction techniques~\cite[Theorem 2]{Karimireddy2021History}.

This paper presents \algoname{}, a new algorithm extending Robust-DSGD with training data confidentiality guarantees.
\algoname{} defends against Byzantine nodes with state-of-the-art robustness guarantees and protects the privacy of the nodes from the (single) server using only HE as security primitive.
At the core of \algoname{} lies our main contribution, namely \aggregator{}, our novel homomorphic operator for the robust aggregator CWTM~\cite{yin2018byzantine}.
To develop an efficient \aggregator{}, we propose a plaintext encoding method that has a general applicability, and demonstrate its usefulness within our implementation on the task of homomorphic sorting.
Most importantly, \aggregator{} would have been more challenging to implement efficiently without this contribution.
An additional by-product of our work is the ability to implement \aggregatortwo{}, a homomorphic operator for CWMED~\cite{yin2018byzantine}, without incurring additional costs.

%% file: notation.tex
\section{Table of notations}\label{sec:notation}

\begin{table}[htbp]
\begin{center}
\caption{ \justifying Table of notations used in this paper.}
\label{table:notation}
\begin{tabular}{r c p{5cm} }
\toprule
$n$ & $\triangleq$ & Number of nodes\\
$f$ & $\triangleq$ & Number of Byzantine nodes\\
$d$ & $\triangleq$ & Size of model/gradients\\
$T$ & $\triangleq$ & Number of steps of \algoname{}\\
$g_{t}^{(i)}$ & $\triangleq$ & Gradient of node $i$ at step $t$\\
$m_{t}^{(i)}$ & $\triangleq$ & Momentum of node $i$ at step $t$\\
$\vec{c}_{t}^{(i)}$ & $\triangleq$ & Vector of ciphertexts of node $i$ at step $t$\\
$\theta_{t}$ & $\triangleq$ & Model parameters at step $t$\\
$\gamma$ & $\triangleq$ & Learning rate\\
$\beta$ & $\triangleq$ & Momentum coefficient\\
$\delta$ & $\triangleq$ & Bit precision\\
$C$ & $\triangleq$ & Clamp parameter\\
\hline
\hline
$m$ & $\triangleq$ & Index of cyclotomic polynomial\\
$\varphi(m)$ & $\triangleq$ & Degree of cyclotomic polynomial\\
$p$ & $\triangleq$ & Plaintext modulus\\
$q$ & $\triangleq$ & Ciphertext modulus\\
$\lambda$ & $\triangleq$ & Cryptographic bit security\\
$B$ & $\triangleq$ & Base of digit expansion\\
$\ordp$ & $\triangleq$ & Length of digit decomposition\\
$\nslots$ & $\triangleq$ & Number of slots\\
\bottomrule
\end{tabular}
\end{center}
\end{table}

In this section, we present a comprehensive table of notations employed throughout the paper for clarity and reference.

%% file: app_byzantine_friendliness.tex
\section{HE-compliance of Byzantine Aggregators (Full Version)}\label{app_fhe_frendliness}
In this section, we qualitatively review six prominent Byzantine aggregators with respect to their compliance with the mainstream FHE (and SHE) cryptosystems in terms of evaluation cost.
We classify these operators into two categories.
First, we have geometric approaches which select one or more vectors based on a global criterion, but which are subject to a selection bottleneck in high dimensions.
Second, coordinate-wise approaches, which manipulate each coordinate independently, appear to be the most promising, in particular those with a static selection criterion.
We base our analysis on the characteristics of the mainstream FHE schemes discussed in Section~\ref{sec_he_for_ml},
which we use as a yardstick to determine the feasibility of each of these methods.

\subsection{Geometric Aggregators}
Geometric aggregators generally require the calculation of criteria (e.g., norms, distances) over the nodes' vectors, which then have to be compared to select one or more vectors.
Furthermore, they usually require several levels of comparisons, min/max and argmin/max operators, etc. As such, geometric methods should be better handled by the TFHE cryptosystem. However, since TFHE does not support batching, effectively extracting the item of interest is problematic in the case of high-dimensional vectors. For example, once the index of the argmin/max is determined (e.g., giving an encryption of the one-hot encoding of that index), one has to perform a dot-product for each component of the vector to retrieve, component by component, the encrypted vector.
Scaling abilities are thus limited to very small models ($\sim$ 1000 parameters at most), which is definitely not representative of modern deep neural networks. This trade-off between the heavy dependence on the model size and the efficiency of implementing these methods over TFHE makes geometric approaches generally difficult to implement efficiently in the homomorphic domain.
We next have a more in-depth look at three prominent geometric aggregators from the literature. 

\textbf{(Multi-)Krum~\cite{krum}.}
To illustrate the above, Krum and Multi-Krum would be very difficult to implement fully in the encrypted domain. Indeed, both approaches require first selecting the $n-f$~\footnote{Recall that $n$ is the total number of nodes, and $f$ is the number of Byzantine nodes.} nearest neighbors of each vector, then computing a score per vector based on the distances to these neighbors, and finally selecting one or more vectors of minimum score. Although \textit{homomorphic nearest neighbors} has been shown to be practical for low-dimensional vectors~\cite{ZS21,CZ22} over TFHE, the fact that this method requires multiple nearest neighbor operations at once renders the overall scheme computationally heavy. Furthermore, an additional round of selection is then required to determine the vector(s) with minimum score. Due to their two-level selection process, both Krum and Multi-Krum would thus only be practically achievable over TFHE. Even then, because TFHE offers no batching capabilities, both approaches are subject to the selection bottleneck and cannot be expected to scale to large models.

\textbf{Minimum Diameter Averaging (MDA)~\cite{brute_bulyan}.}
This method requires solving a combinatorial optimization problem to determine the subset of vectors of size $n-f$ with minimum diameter.
The average of these vectors is then output by the method.
From an FHE perspective, this implies distance computations, followed by three levels of argmin/argmax computations (nearest neighbors, diameter, subset selection), and terminated by an averaging operation.
This renders the computations even more expensive than Krum.

\textbf{Geometric Median (GM)~\cite{chen2017distributed}.}
Computing the geometric median requires solving a continuous optimization problem, most likely via an iterative algorithm. As such, it involves a prohibitive number of homomorphic operations. Additionally, the most well-known approximation method for GM, namely the Smoothed Weiszfeld algorithm~\cite{pillutla2019robust}, uses a ciphertext-vs-ciphertext division that cannot be escaped.

\subsection{Coordinate-wise Aggregators}
Coordinate-wise aggregators are intrinsically batching-friendly since all vector coordinates 
can be processed independently. These approaches are therefore a priori promising candidates to be implemented over BGV or any-other batching-able FHE scheme.
In fact, their homomorphic evaluation would scale well to large models, thanks to both the SIMD parallelism available in each ciphertext and the fact that many such ciphertexts can be straightforwardly processed in parallel. However, all presently known batching-friendly cryptosystems have generally higher computational complexities than TFHE when implementing comparisons and min/max operators. So, the computational times are expected to exhibit a higher sensitivity to the number of nodes.

\textbf{Coordinate-wise median (CWMED)~\cite{yin2018byzantine}.}
Median calculations have recently been studied over batching-friendly cryptosystems~\cite{IZ21}. However, computing a median even over a small number of values requires large computational times. For example, \cite{IZ21} reports that a median over $16$ $8$-bit vectors of size $9352$ takes around $1$ hour. The \emph{amortized} time per component appears more reasonable ($\approx 0.38s$). These results are promising (by FHE standards) and hint that more practical computational times may be obtained using further optimizations, as well as by decreasing the precision of the vectors and/or the number of values involved in the calculation.


\textbf{Coordinate-wise trimmed mean (CWTM)~\cite{yin2018byzantine}.}
This method performs a coordinate-wise sorting and then averages the ($n-2f$) coordinates between positions $f$ and $n-f-1$. 
Specifically, given $n$ vectors $x_0, ..., x_{n-1}$ of dimension $d$, CWTM sorts the $n$ vectors per coordinate to obtain $\overline{x}_0, \dots, \overline{x}_{n-1}$ satisfying
\begin{equation}
    \overline{x}_{0}^{(j)} \leq ... \leq \overline{x}_{n-1}^{(j)} \qquad 0\le j\le d-1\,,
\end{equation}
and returns the vector $y$ such that
\begin{equation}
    y = \frac{1}{n-2f} \sum_{i=f}^{n-f-1} \overline{x}_{i}
\end{equation}
This operator should not be much more costly than CWMED, since the selection is replaced by an averaging over statically selected coordinates and summation is generally a low-cost homomorphic operation.
Note that since divisions are quite impractical in the homomorphic domain, the averaging operation in CWTM can be replaced by a homomorphic summation followed by a post-decryption division on the nodes' side.
Indeed, there is no added value in terms of security to perform the division prior to the decryption.

\textbf{Remark on CWTM vs CWMED.}
Be it for the trimmed mean or the median, the algorithm in the encrypted domain requires to sort the values and compute their ranks.
Since data-dependent branching is not possible in a homomorphic setting, this implies that all pairs of values must be compared, thus yielding quadratic complexity.
Also, in both algorithms, the rank of all elements must be homomorphically determined in order to decide which ones enter the final computation.
In other words, both algorithms exhibit the same computational complexity and multiplicative depth.

\textbf{Mean around Median (MeaMed)~\cite{meamed}.}
Although similar to the two previous operators, this algorithm is much more difficult to perform in the encrypted domain as it requires to compute a coordinate-wise average over the $n-f$ closest values to the median (per coordinate). That is, contrary to CWTM and CWMED, the selection criterion of the coordinates involved in the average is not static (equivalent to an $n-f$ nearest neighbors in one dimension), significantly increasing the volume and complexity of the required homomorphic computations compared to the previous two operators.







\subsection{Takeaway}
Following the above discussion, from an HE execution viewpoint, our qualitative analysis a priori favors the coordinate-wise methods CWTM and CWMED to be the most promising. They are expected to scale better in terms of model size over batching-friendly schemes compared to the geometric approaches.
Fortunately, from an ML viewpoint, these two popular methods have been shown to provide state-of-the-art Byzantine robustness~\cite{RESAM, Karimireddy2021History, karimireddy2022byzantinerobust}.
Furthermore, CWTM has been shown to be empirically and theoretically superior to CWMED in~\cite{allouah2023fixing}.
As such, we choose in this work to achieve Byzantine resilience via a homomorphic implementation of CWTM over the BGV cryptosystem.

%% file: app_exp_setup.tex
\section{Additional Information on Experimental Setup}
We present additional information on the experimental setup that could not be placed in Section~\ref{sec_exp} due to space constraints.

\subsection{Dataset Distribution and Preprocessing}\label{app_exp_setup_dist}
\begin{figure*}[ht!]
    \centering
    \includegraphics[width=0.45\textwidth]{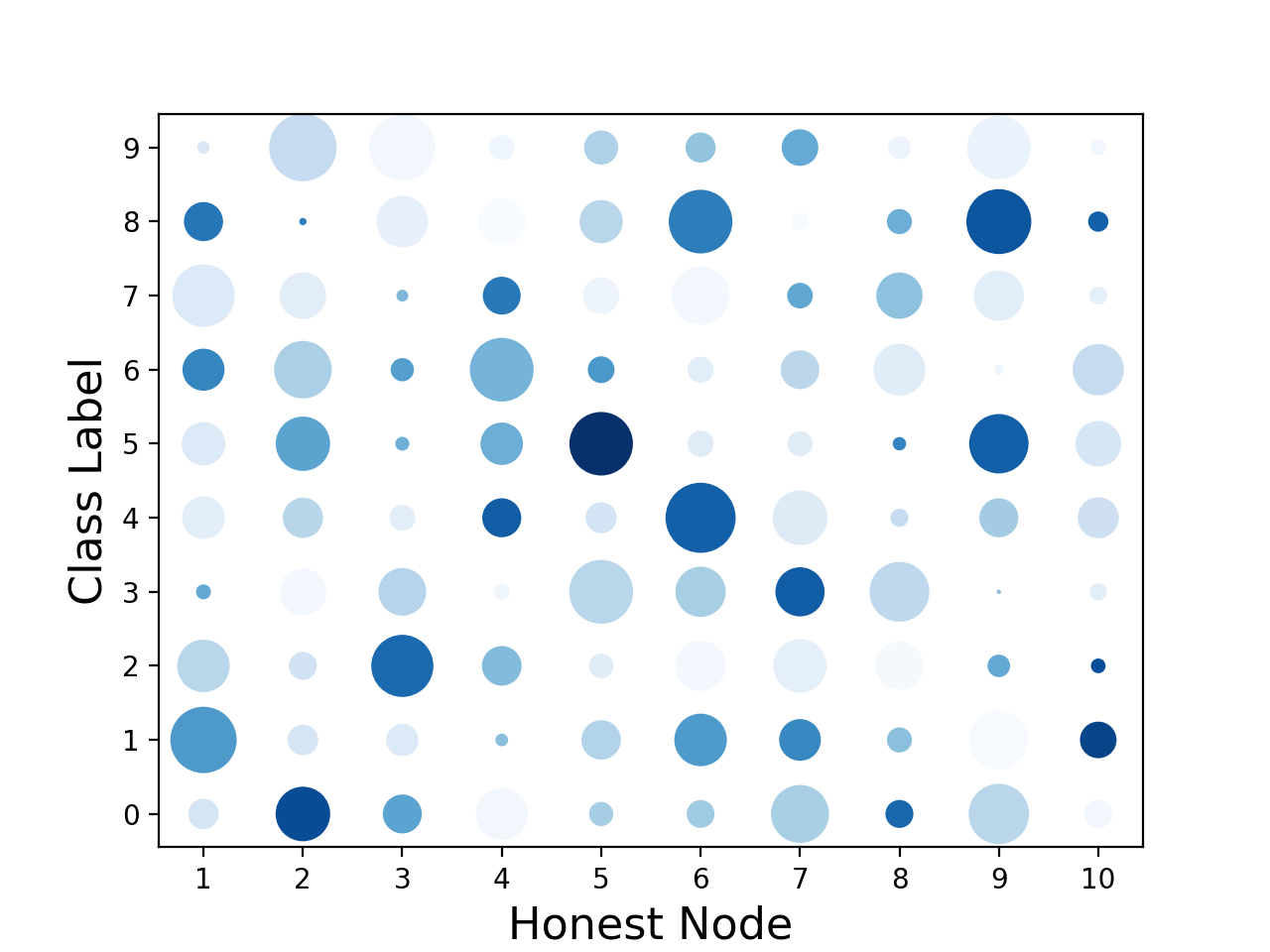}%
    \includegraphics[width=0.45\textwidth]{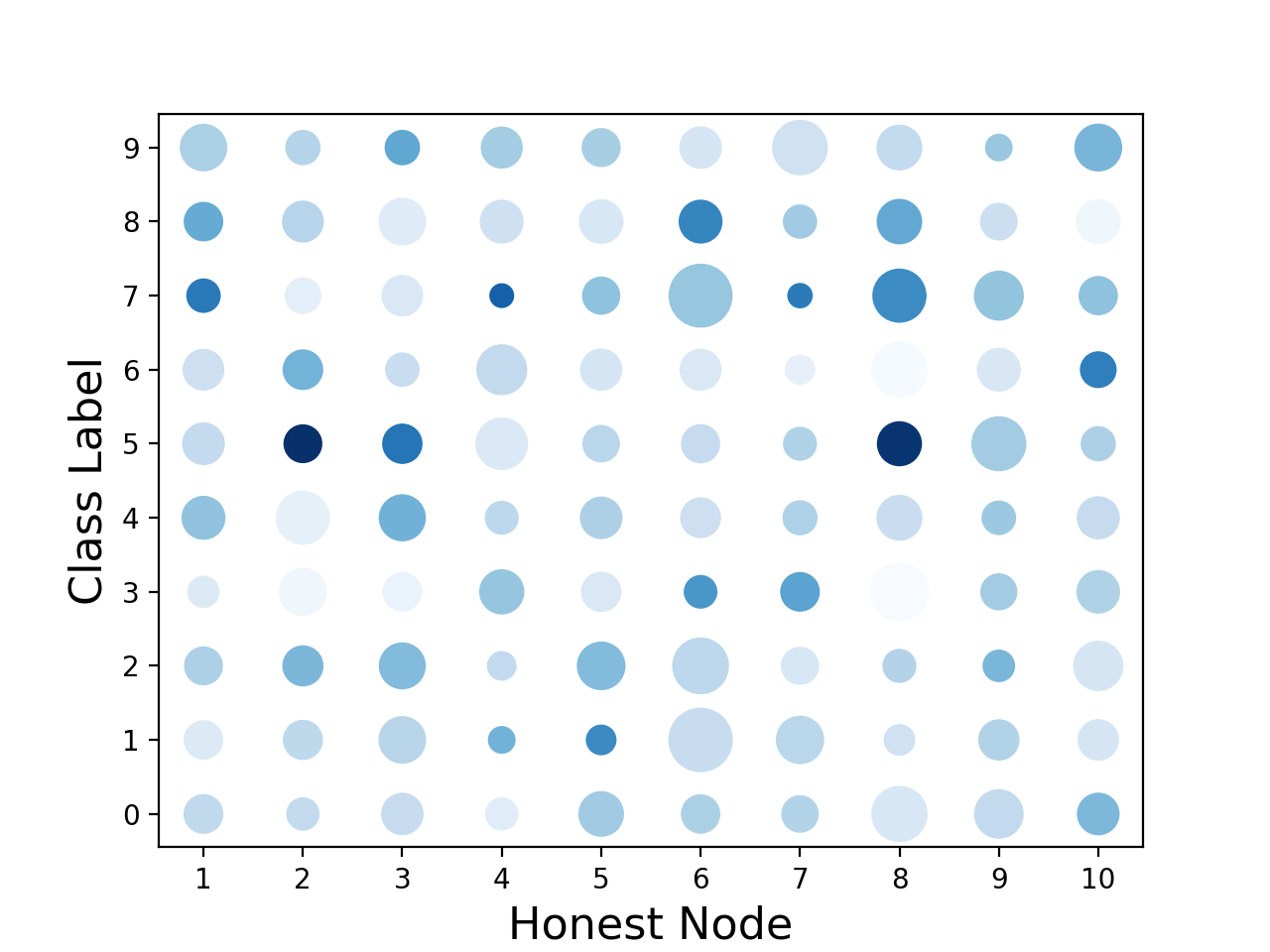}%
    \caption{ \justifying Distribution of class labels across honest nodes on MNIST (left) and Fashion-MNIST (right), in the setting presented in Section~\ref{sec_exp_results} (i.e., $f = 5$).
    The honest nodes sample from a Dirichlet distribution of parameter $\alpha = 1$ (left) and $\alpha = 5$ (right). The larger the circle, the more data samples of the corresponding class label are present (the shading of the blue color is not indicative).}
\label{fig:label_distribution}
\end{figure*}

First, we present pictorial representations of the data heterogeneity induced in our experiments on MNIST and Fashion-MNIST.
Indeed, we present in Figure~\ref{fig:label_distribution} the distribution of class labels across honest nodes when $\alpha = 1$ (left) and 5 (right).
Recall that $\alpha$ is the parameter of the Dirichlet distribution (see Section~\ref{sec_exp_setup}).
Second, we employ the following classical input transformations on the three datasets~\cite{RESAM, allouah2023fixing}.
The images of MNIST are normalized with mean $0.1307$ and standard deviation $0.3081$, while the images of Fashion-MNIST are horizontally flipped.
CIFAR-10 is expanded with horizontally flipped images, along with per channel normalization of means 0.4914, 0.4822, 0.4465, and standard deviations 0.2023, 0.1994, 0.2010.

\subsection{Model Architectures}\label{app_sec_model_arch}
To present the architecture of the models used, we employ the following compact notation used in~\cite{RESAM, allouah2023fixing}.
\newline
\newline
\noindent \fcolorbox{black}{blue!30}{
\parbox{0.45\textwidth}{
L(\#inputs, \#outputs) represents a \textbf{fully-connected linear layer}, R stands for \textbf{ReLU activation}, S stands for \textbf{log-softmax}, C(\#in\_channels, \#out\_channels) represents a \textbf{fully-connected 2D-convolutional layer} (kernel size 5, padding 0, stride 1), and M stands for \textbf{2D-maxpool} (kernel size 2).}}
\newline
\newline
\newline
The architectures of the models used on MNIST, Fashion-MNIST, and CIFAR-10 are thus the following:
\begin{itemize}
    \item MNIST: L(784, 100) - R - L(100, 10) - S
    \item Fashion-MNIST: C(1, 20) - R - M - C(20, 50) - R - M - L(800, 500) - R - L(500, 10) - S
    \item CIFAR-10: C(3, 20) - R - M - C(20, 200) - R - M - L(5000, 120) - R - L(120, 84) - R - L(84, 10)
\end{itemize}

%% file: app_exp_results.tex
\section{Additional Experimental Results}\label{app_exp_results}
\begin{figure}[!ht]
    \centering
    \includegraphics[width=0.45\textwidth]{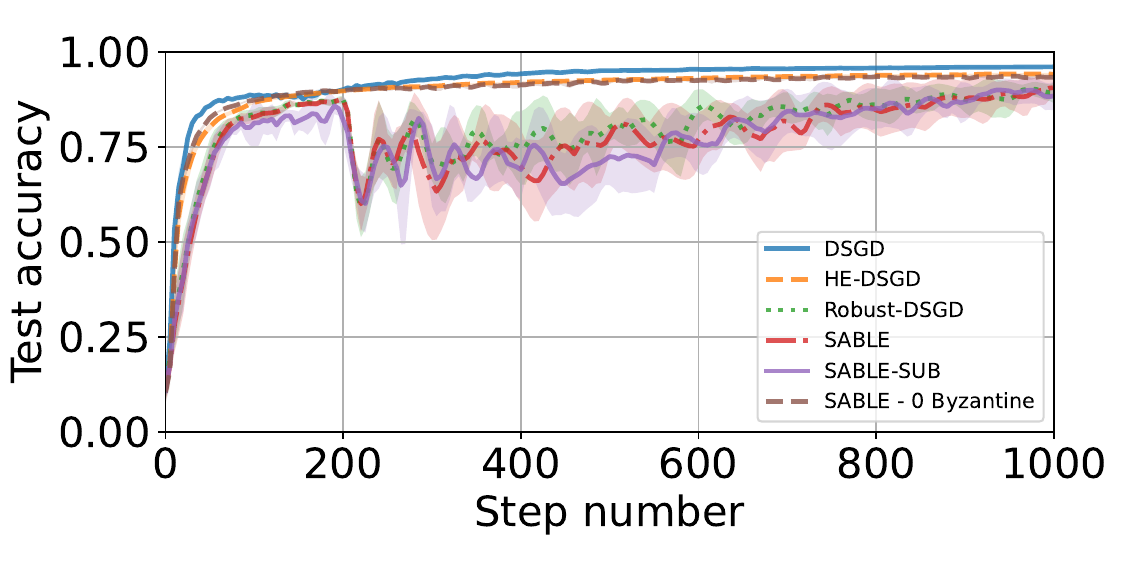}%
    \caption{\justifying Experiments on heterogeneous MNIST with $\alpha=1$ and $f=5$ Byzantine nodes among $n = 15$. The Byzantine nodes execute the mimic attack. This figure complements Figure~\ref{fig:plots_main} in the main paper.} 
\label{fig:plots_mnist_f=5_hetero}
\end{figure}

\begin{figure*}[!ht]
    \centering
    \includegraphics[width=0.45\textwidth]{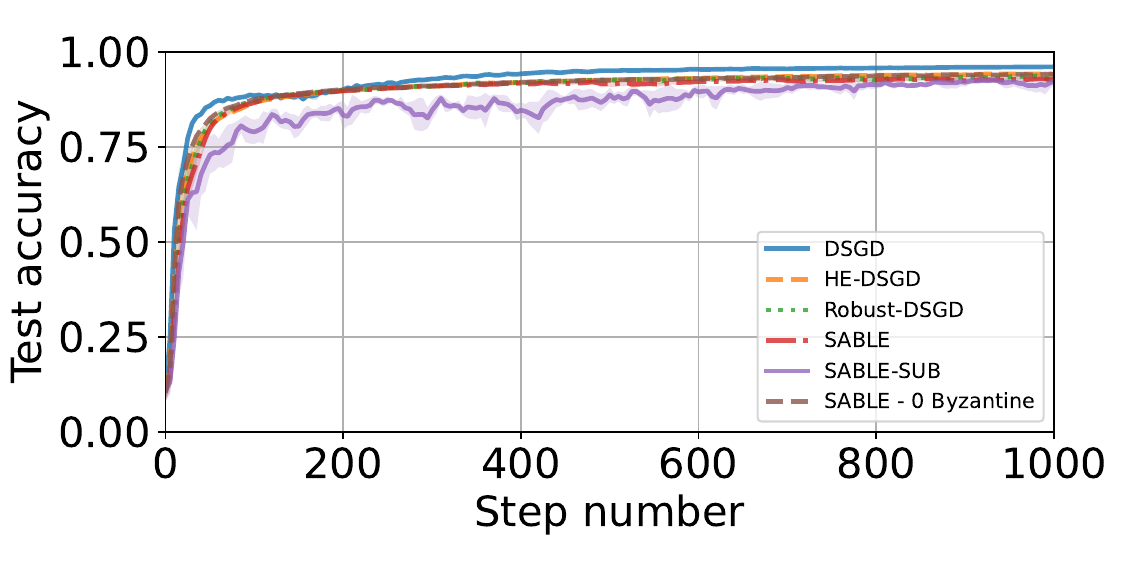}
    \includegraphics[width=0.45\textwidth]{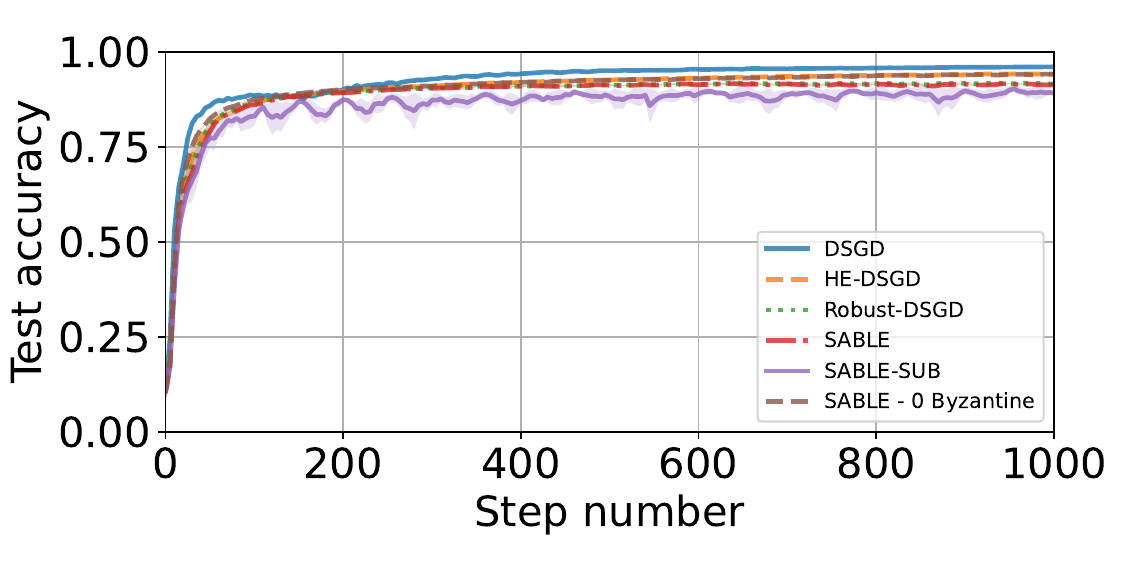}\\
    \vspace{-2mm}
    \includegraphics[width=0.45\textwidth]{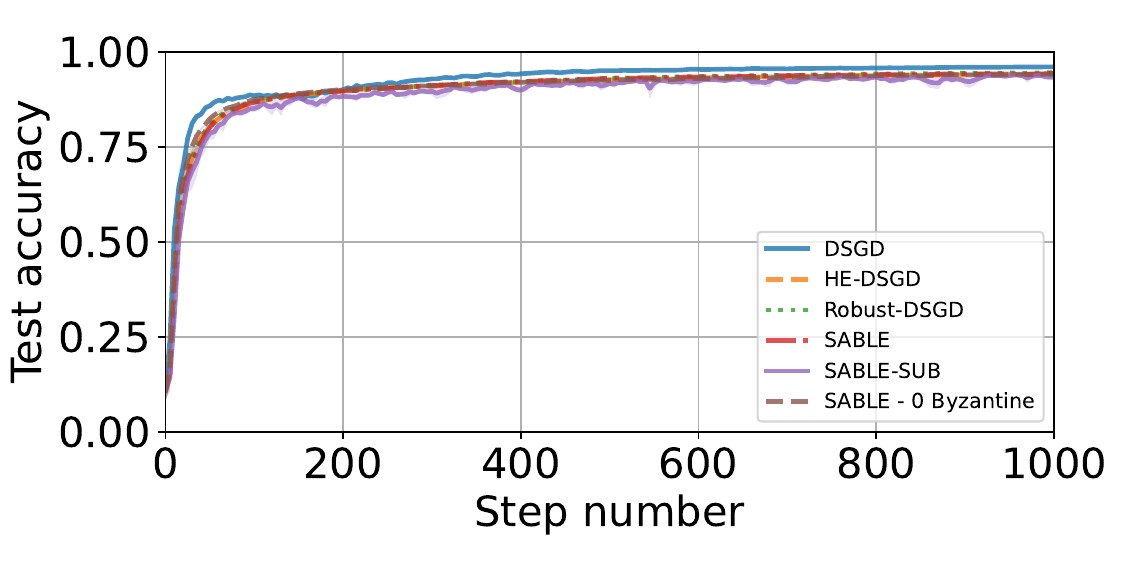}%
    \includegraphics[width=0.45\textwidth]{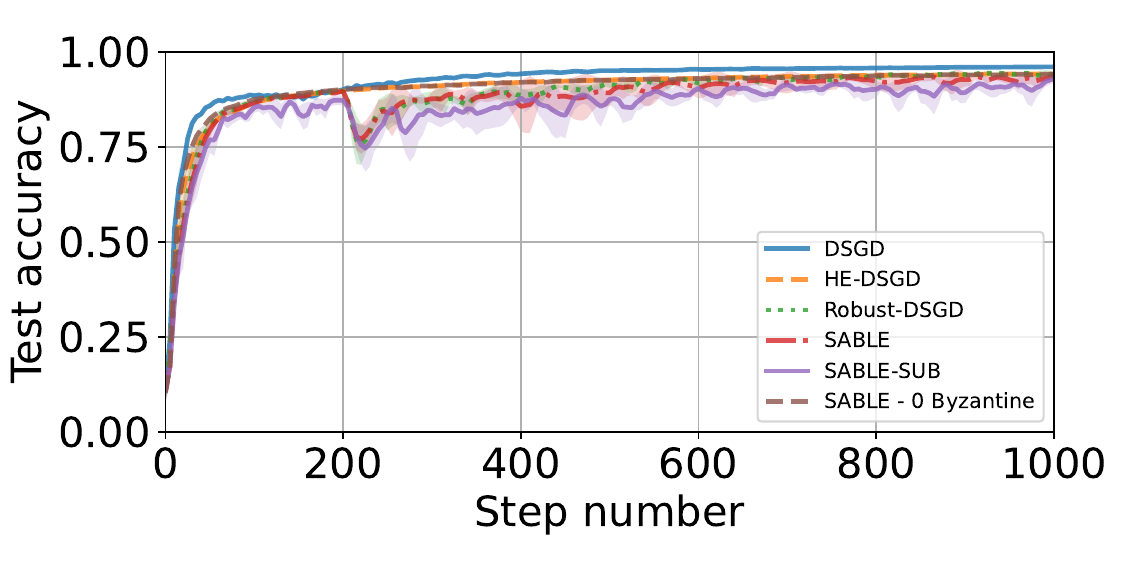}%
    \vspace{-4mm}
    \caption{\justifying MNIST with $f=3$ Byzantine nodes among $n = 15$. Byzantine attacks: FOE (\textit{row1, left}), ALIE (\textit{row 1, right}), LF(\textit{row 2, left}), and mimic (\textit{row 2, right}).}
\label{fig:plots_mnist_f=3_hetero}
\end{figure*}

\begin{figure*}[!ht]
    \centering
    \includegraphics[width=0.45\textwidth]{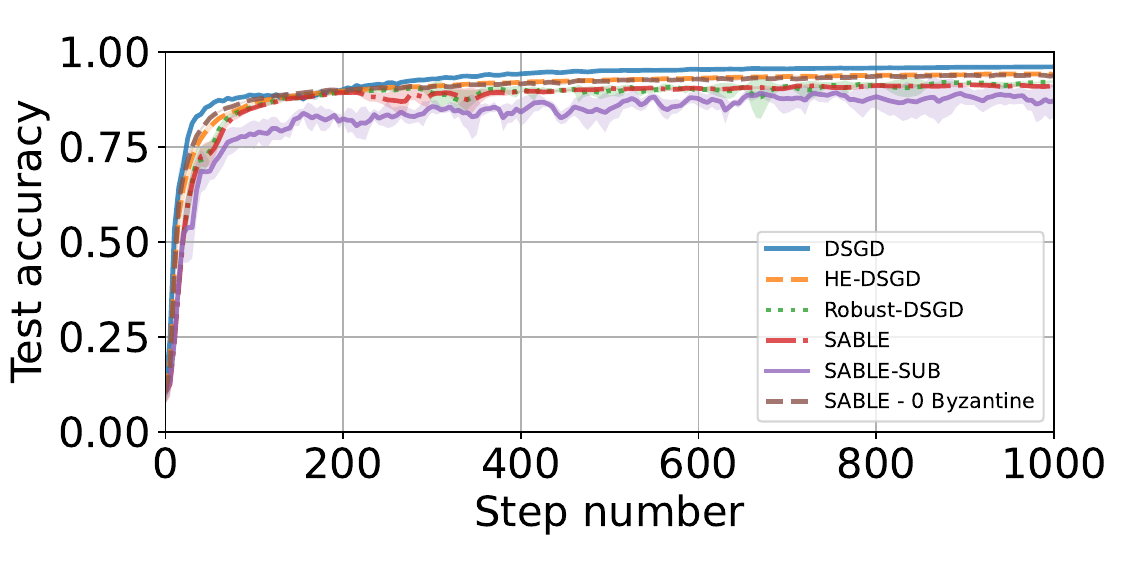}
    \includegraphics[width=0.45\textwidth]{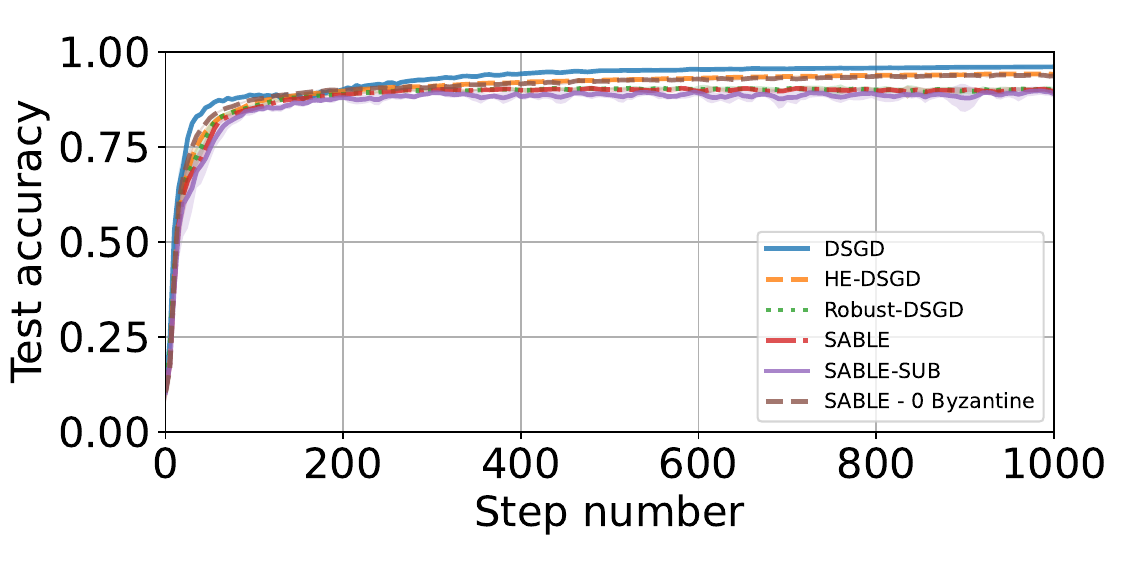}\\
    \vspace{-2mm}
    \includegraphics[width=0.45\textwidth]{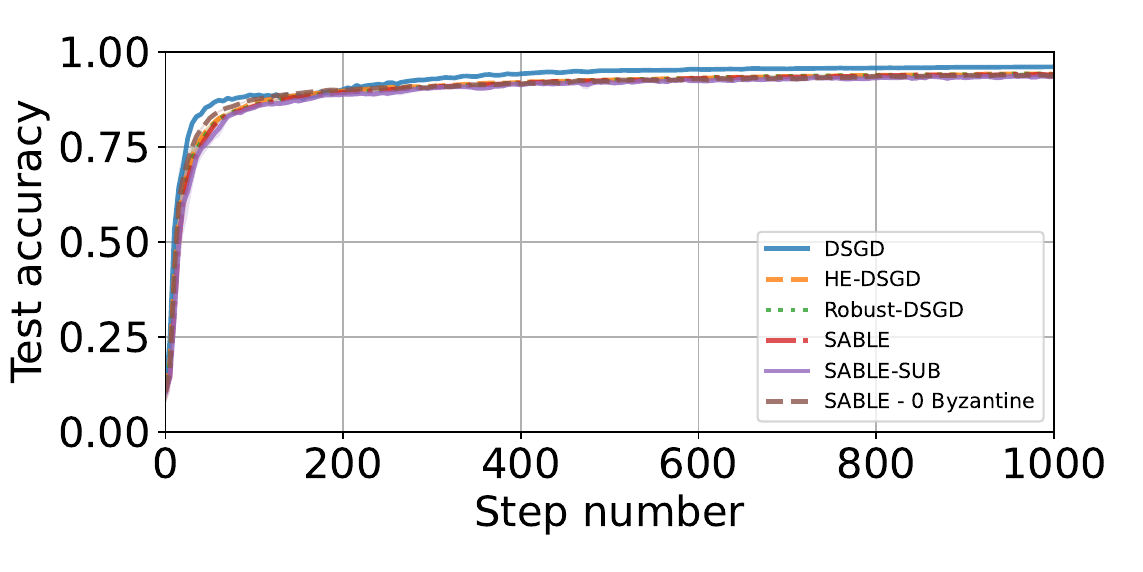}%
    \includegraphics[width=0.45\textwidth]{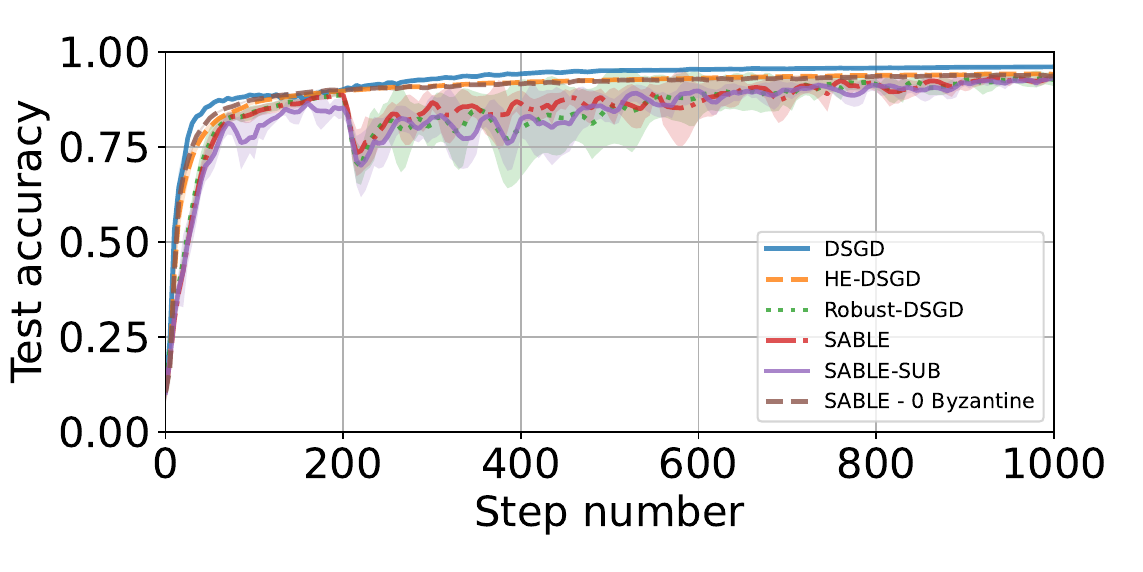}%
    \vspace{-4mm}
    \caption{\justifying MNIST with $f=4$ Byzantine nodes among $n = 15$. Byzantine attacks: FOE (\textit{row1, left}), ALIE (\textit{row 1, right}), LF(\textit{row 2, left}), and mimic (\textit{row 2, right}).}
\label{fig:plots_mnist_f=4_hetero}
\end{figure*}

\begin{figure*}[!ht]
    \centering
    \includegraphics[width=0.45\textwidth]{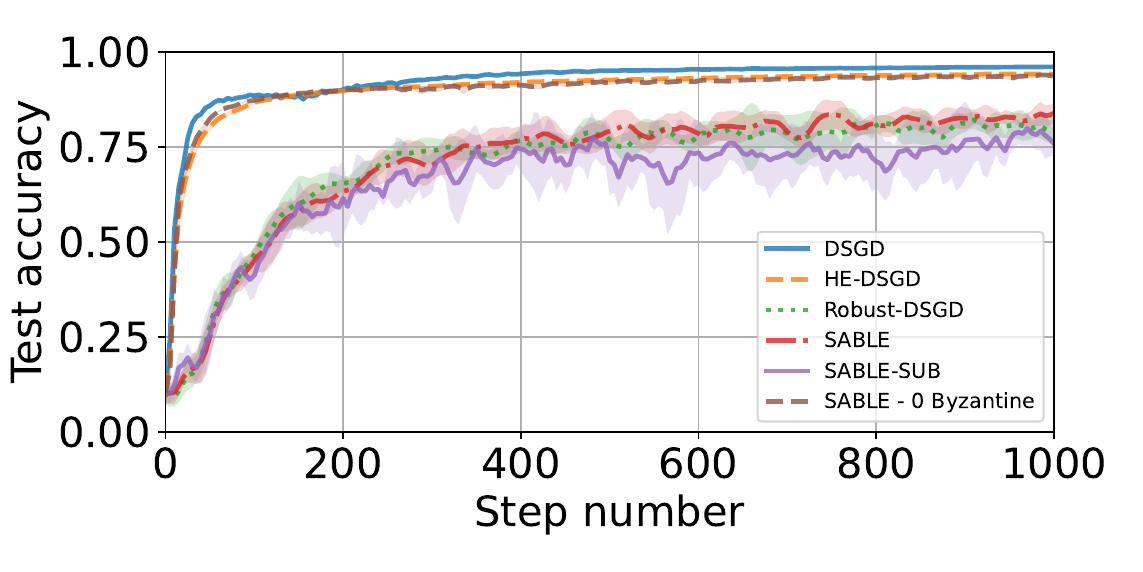}
    \includegraphics[width=0.45\textwidth]{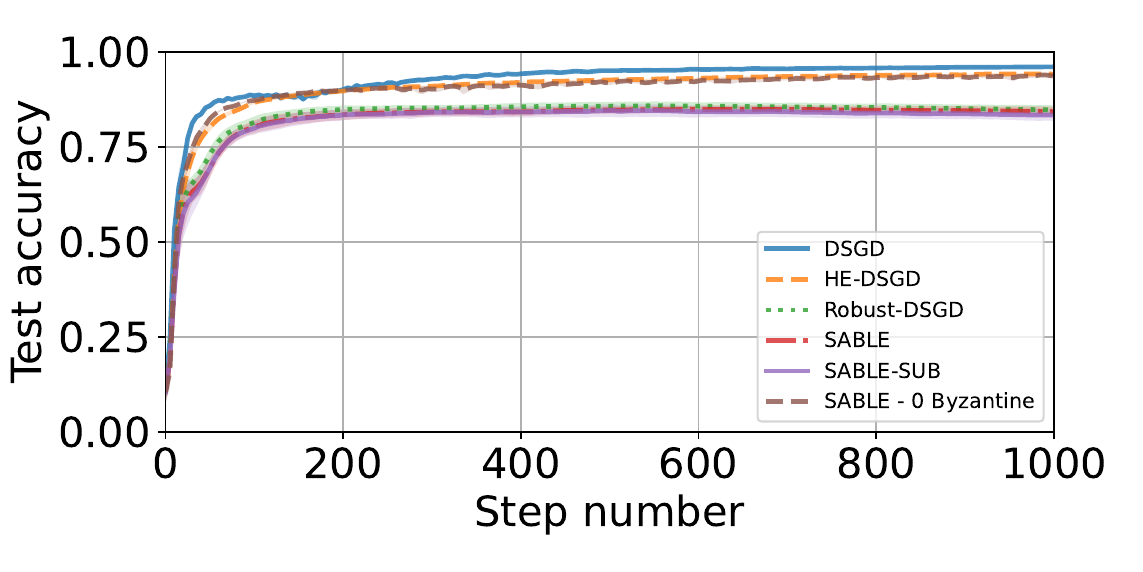}\\
    \vspace{-2mm}
    \includegraphics[width=0.45\textwidth]{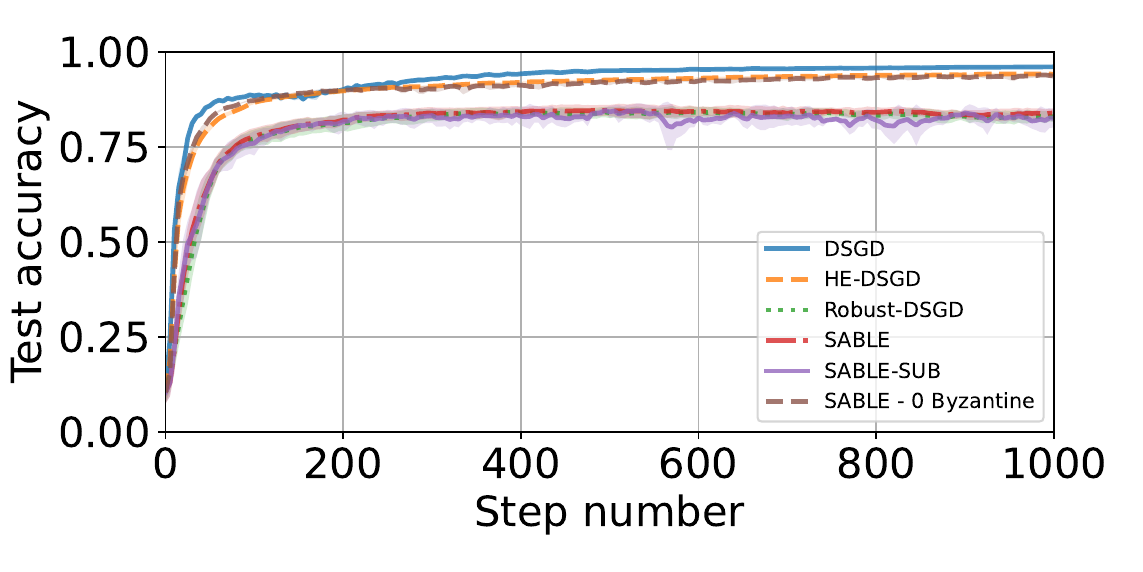}%
    \includegraphics[width=0.45\textwidth]{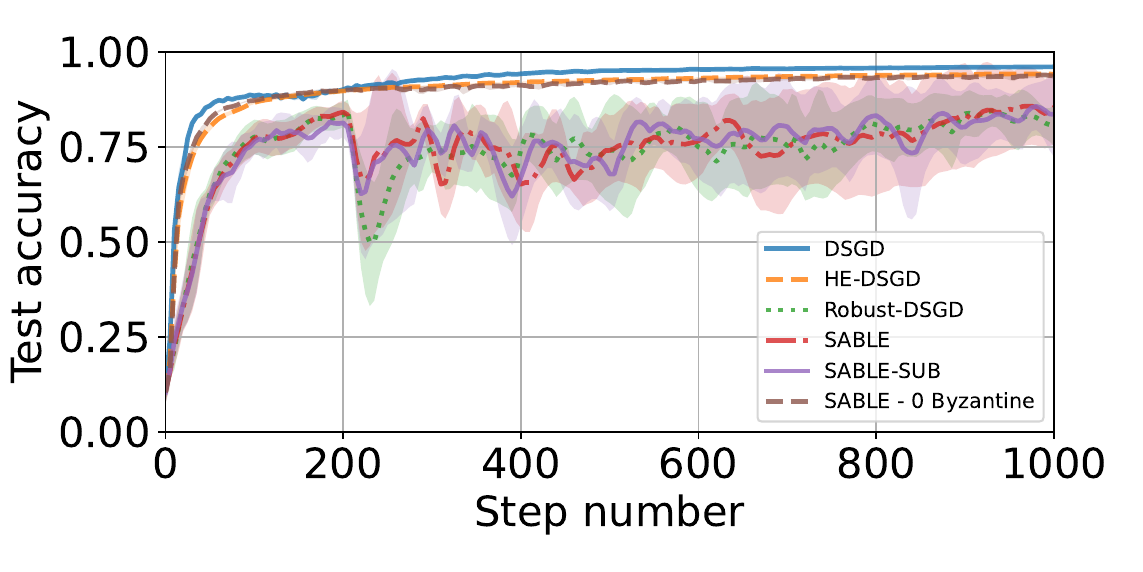}%
    \vspace{-4mm}
    \caption{\justifying MNIST with $f=6$ Byzantine nodes among $n = 15$. Byzantine attacks: FOE (\textit{row1, left}), ALIE (\textit{row 1, right}), LF(\textit{row 2, left}), and mimic (\textit{row 2, right}).}
\label{fig:plots_mnist_f=6_hetero}
\end{figure*}

\begin{figure*}[!ht]
    \centering
    \includegraphics[width=0.45\textwidth]{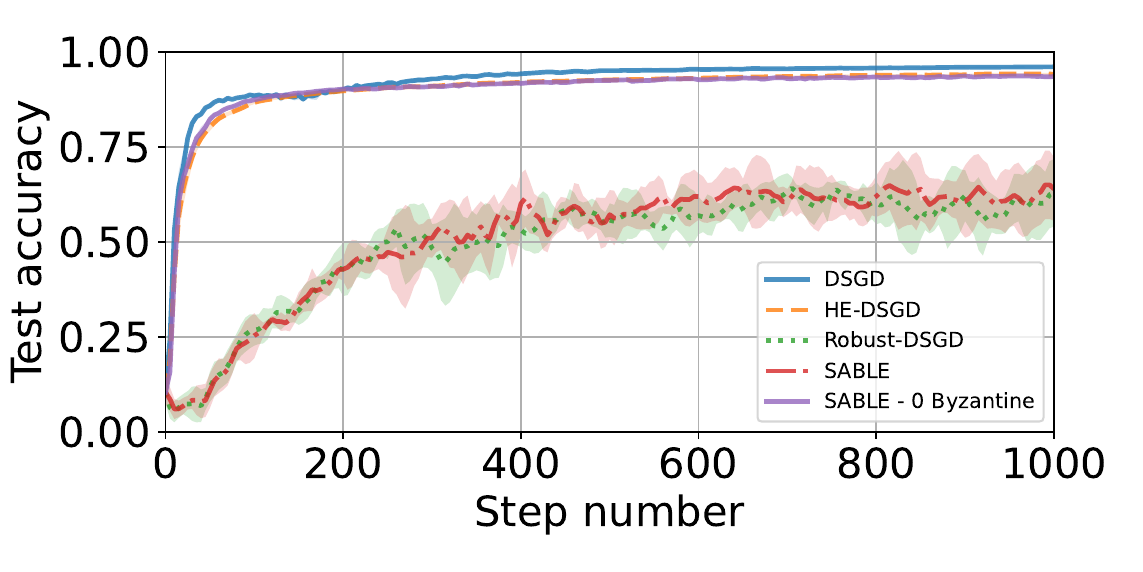}
    \includegraphics[width=0.45\textwidth]{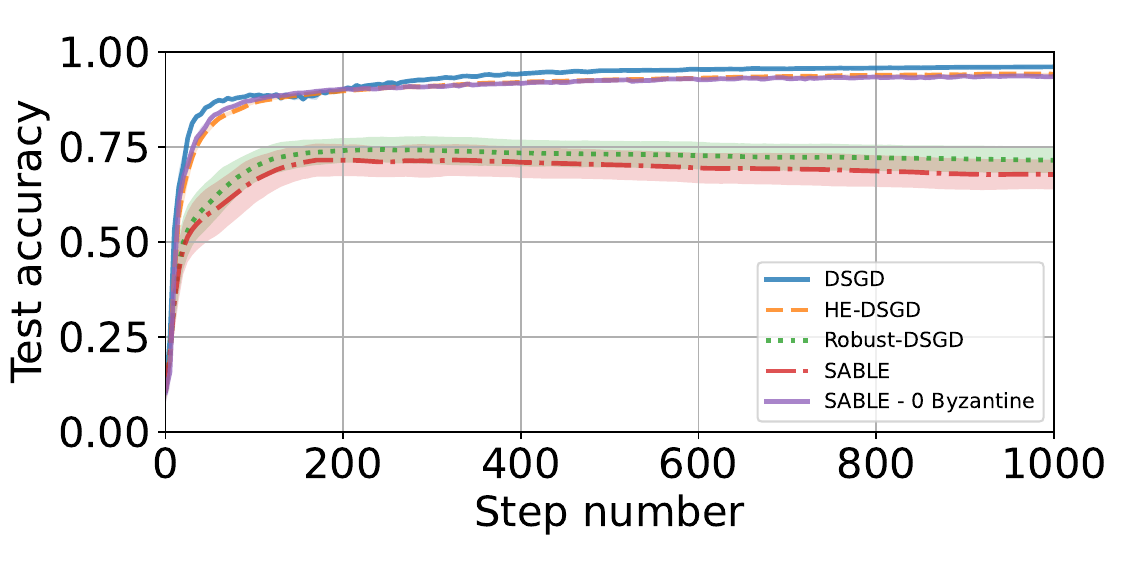}\\
    \vspace{-2mm}
    \includegraphics[width=0.45\textwidth]{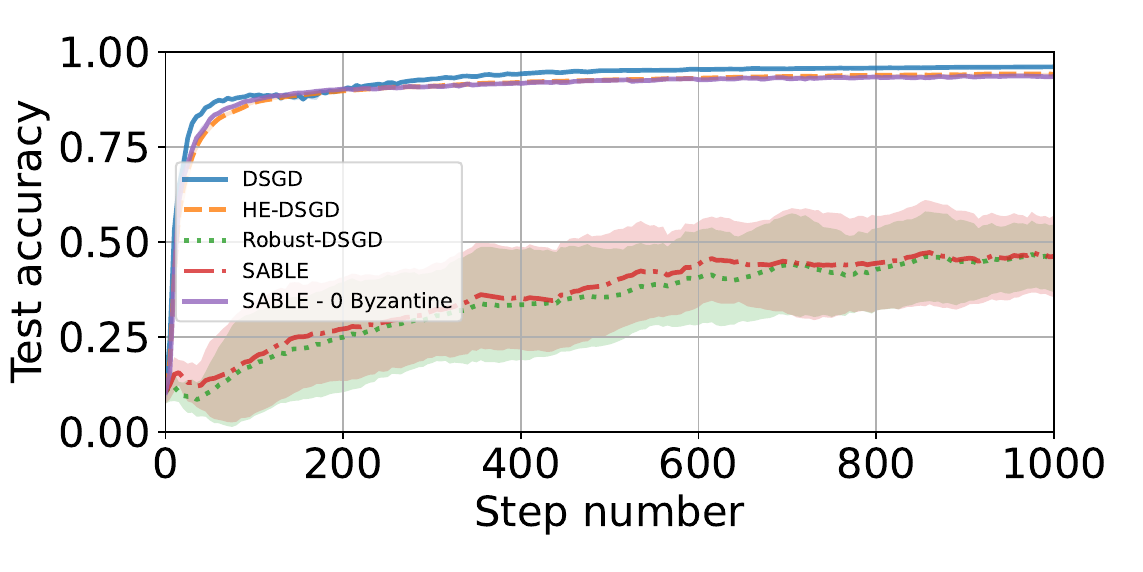}%
    \includegraphics[width=0.45\textwidth]{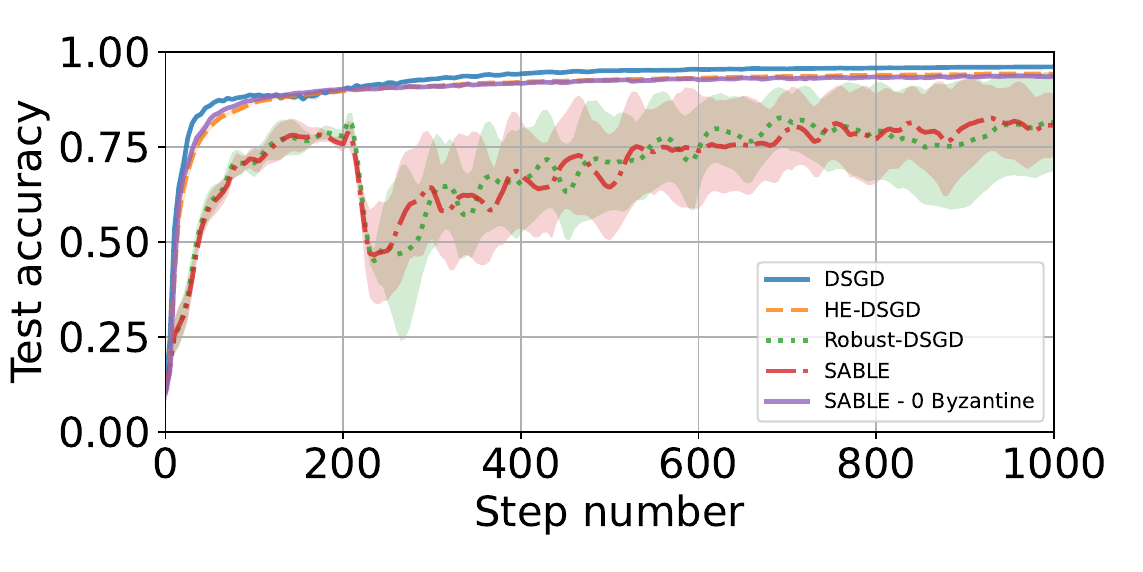}%
    \vspace{-4mm}
    \caption{\justifying MNIST with $f=7$ Byzantine nodes among $n = 15$. Byzantine attacks: FOE (\textit{row1, left}), ALIE (\textit{row 1, right}), LF(\textit{row 2, left}), and mimic (\textit{row 2, right}).}
\label{fig:plots_mnist_f=7_hetero}
\end{figure*}

\begin{figure*}[!ht]
    \centering
    \includegraphics[width=0.45\textwidth]{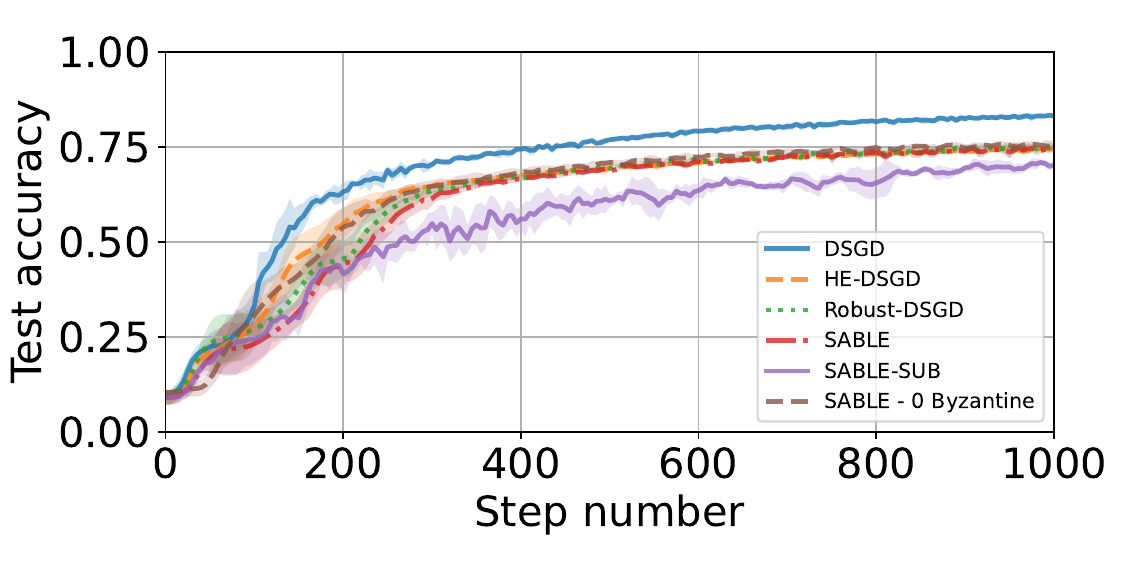}
    \includegraphics[width=0.45\textwidth]{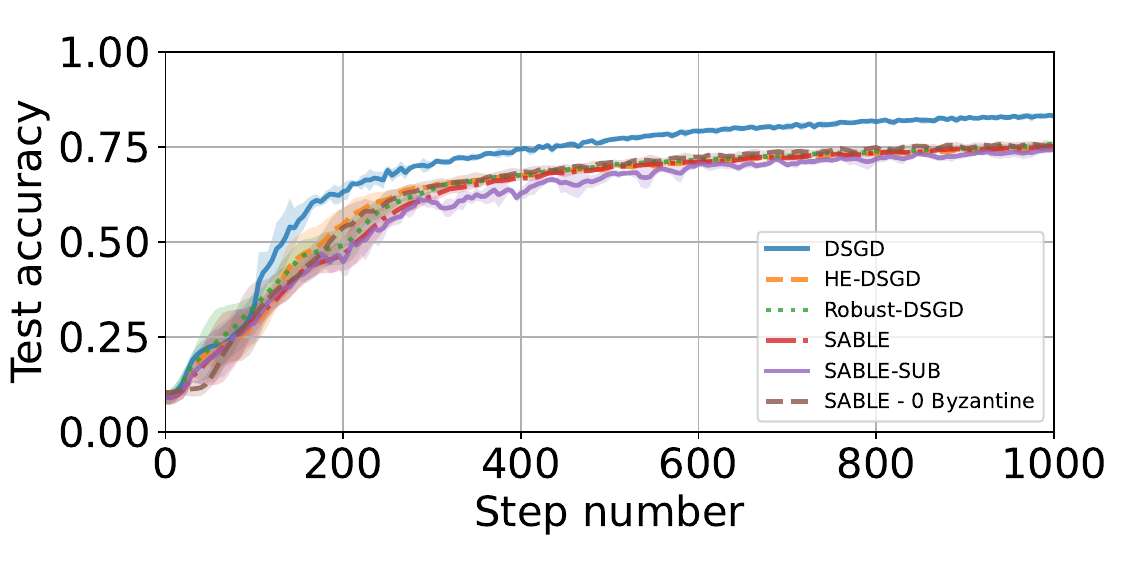}\\
    \vspace{-2mm}
    \includegraphics[width=0.45\textwidth]{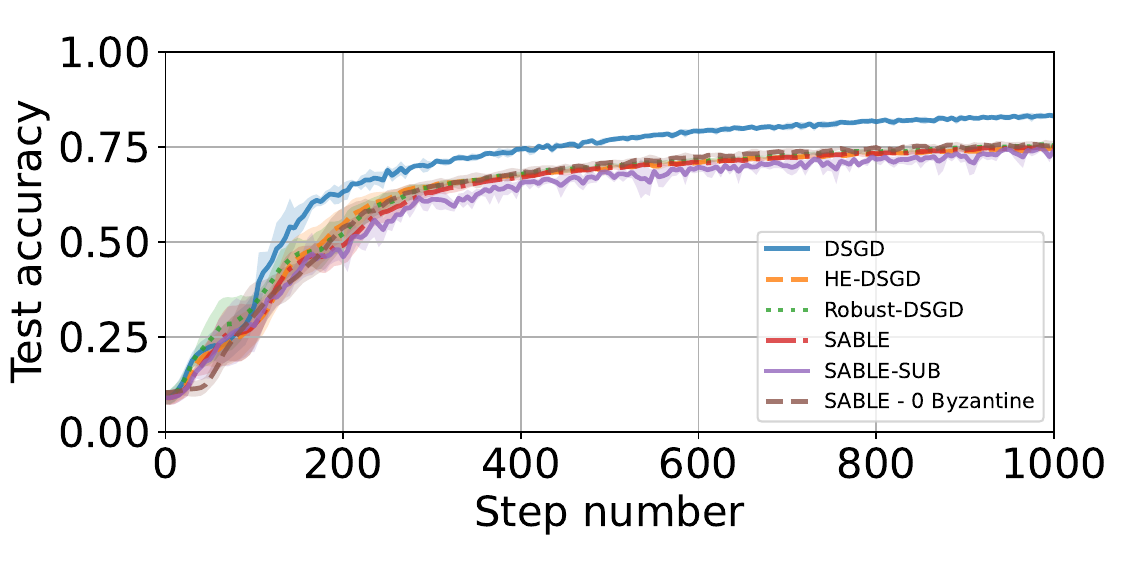}%
    \includegraphics[width=0.45\textwidth]{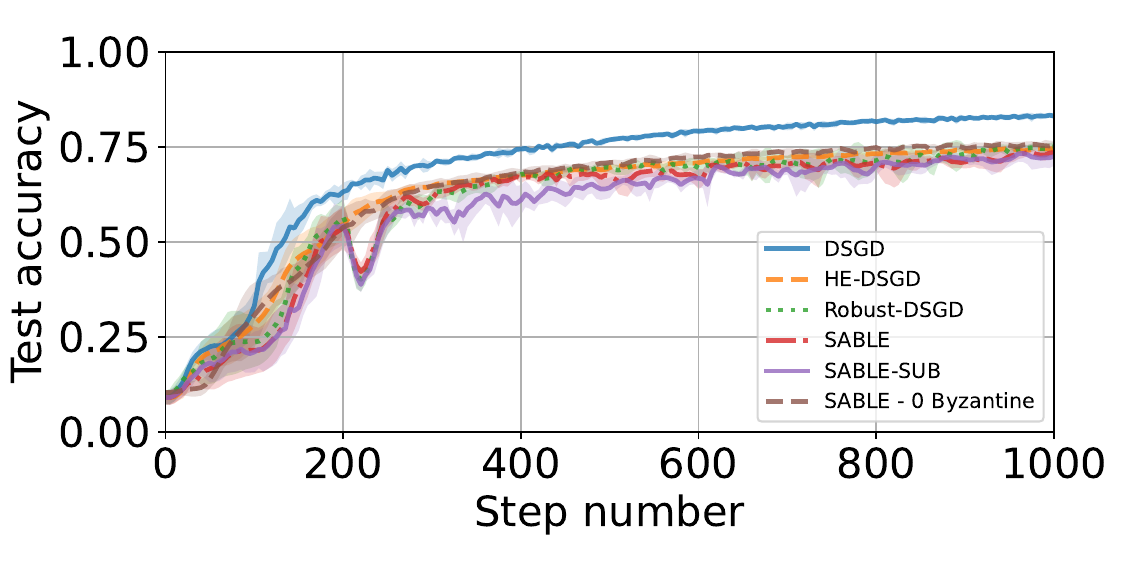}%
    \vspace{-4mm}
    \caption{\justifying Fashion-MNIST with $f=3$ among $n = 15$. Byzantine attacks: FOE (\textit{row1, left}), ALIE (\textit{row 1, right}), LF(\textit{row 2, left}), and mimic (\textit{row 2, right}).}
\label{fig:plots_fashionmnist_f=3_hetero}
\end{figure*}

\begin{figure*}[!ht]
    \centering
    \includegraphics[width=0.45\textwidth]{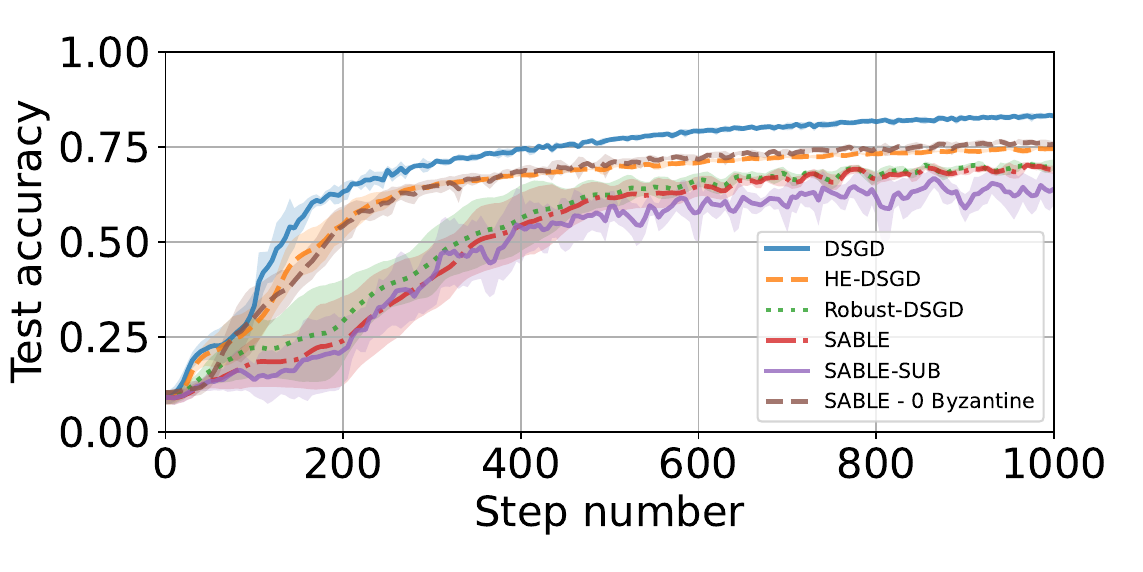}
    \includegraphics[width=0.45\textwidth]{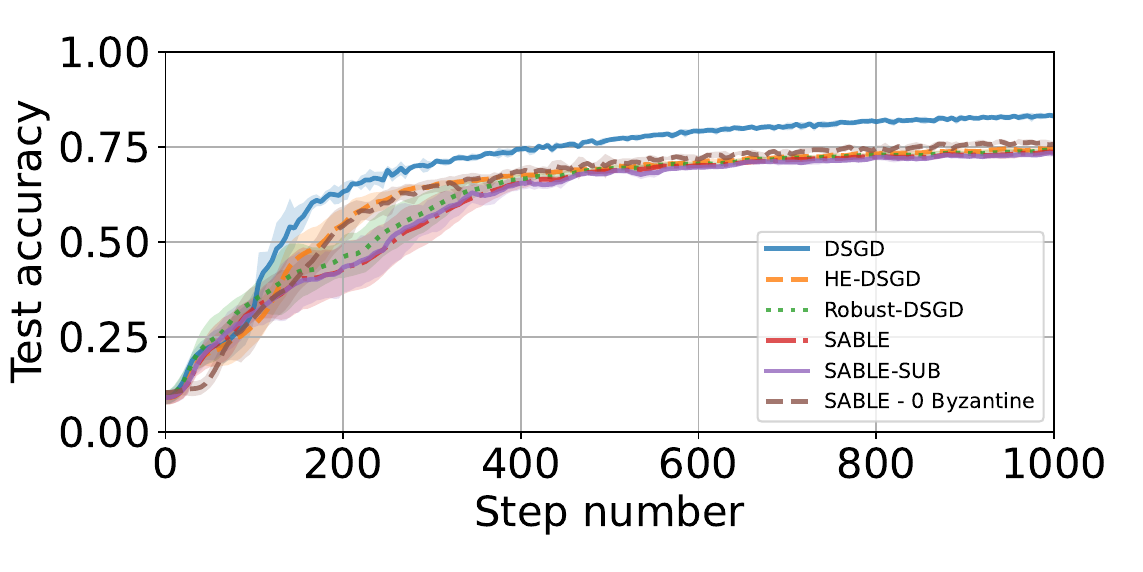}\\
    \vspace{-2mm}
    \includegraphics[width=0.45\textwidth]{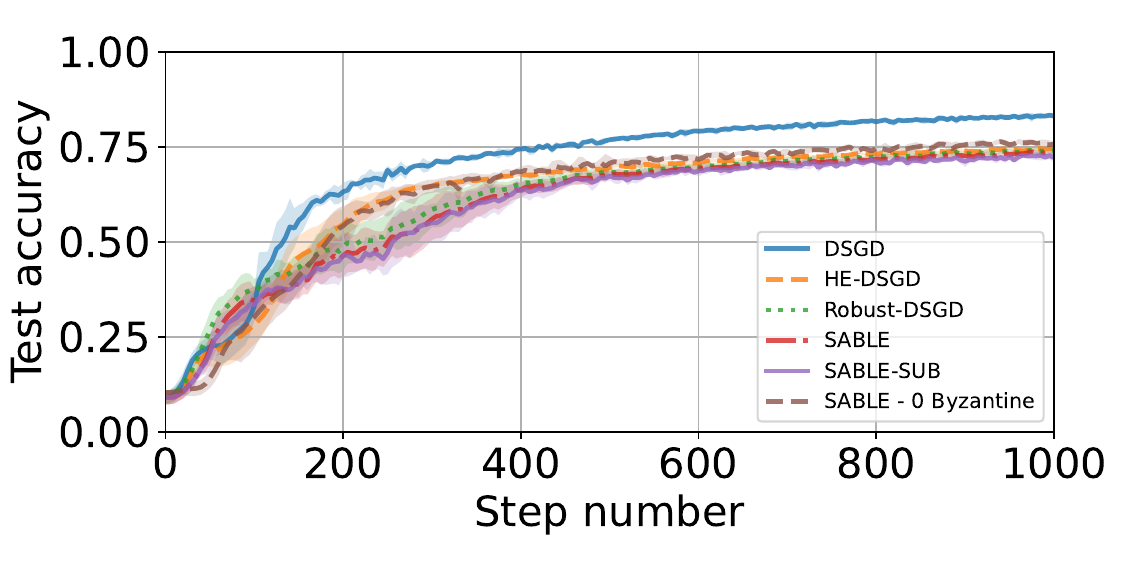}%
    \includegraphics[width=0.45\textwidth]{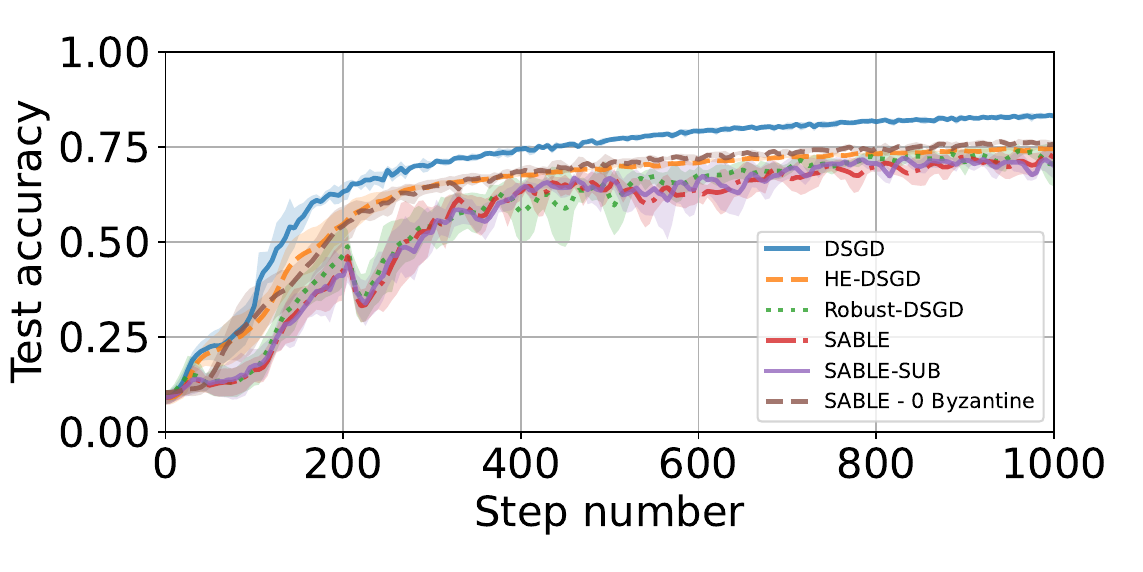}%
    \vspace{-4mm}
    \caption{\justifying Fashion-MNIST with $f=5$ among $n = 15$. Byzantine attacks: FOE (\textit{row1, left}), ALIE (\textit{row 1, right}), LF(\textit{row 2, left}), and mimic (\textit{row 2, right}).}
\label{fig:plots_fashionmnist_f=5_hetero}
\end{figure*}

\begin{figure*}[!ht]
    \centering
    \includegraphics[width=0.45\textwidth]{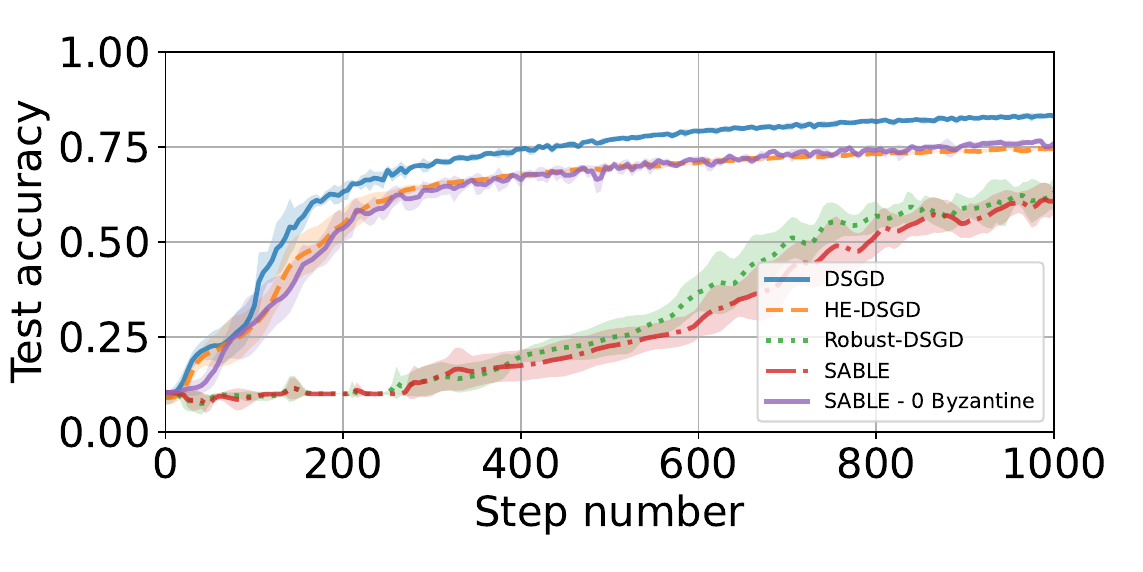}
    \includegraphics[width=0.45\textwidth]{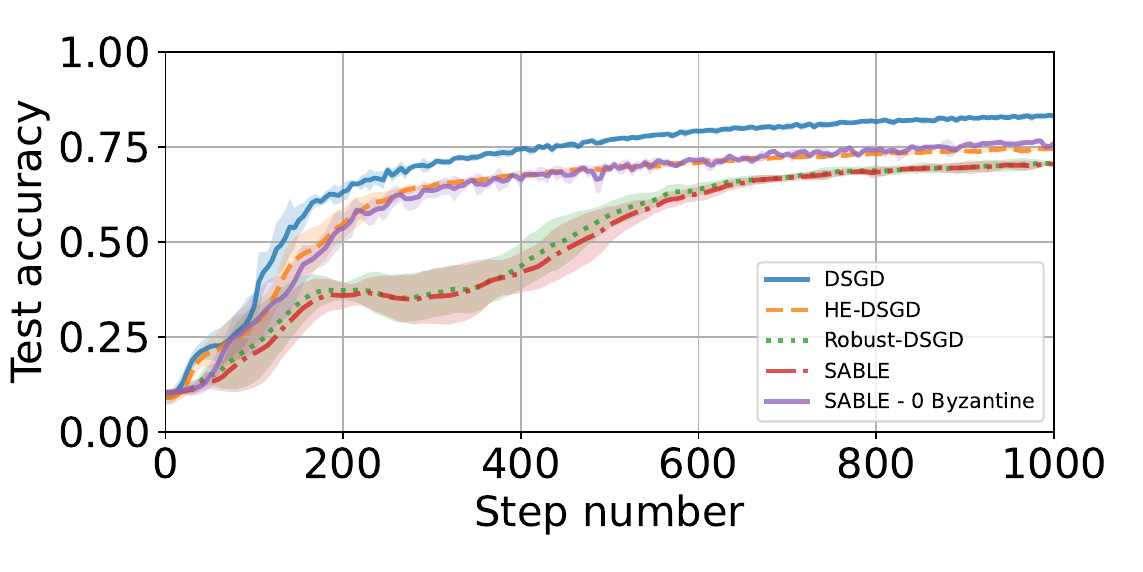}\\
    \vspace{-2mm}
    \includegraphics[width=0.45\textwidth]{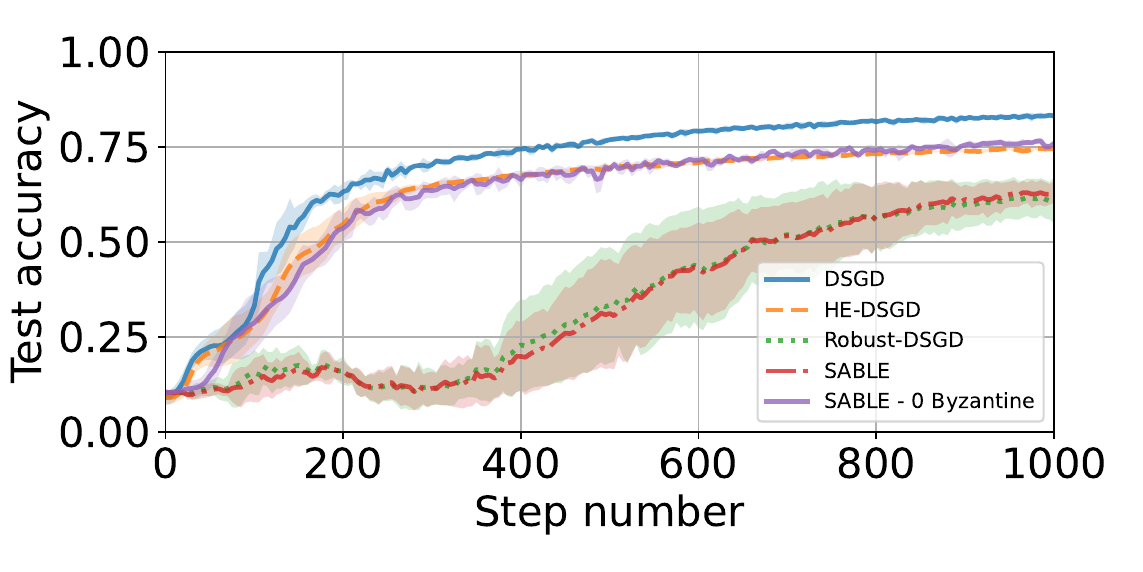}%
    \includegraphics[width=0.45\textwidth]{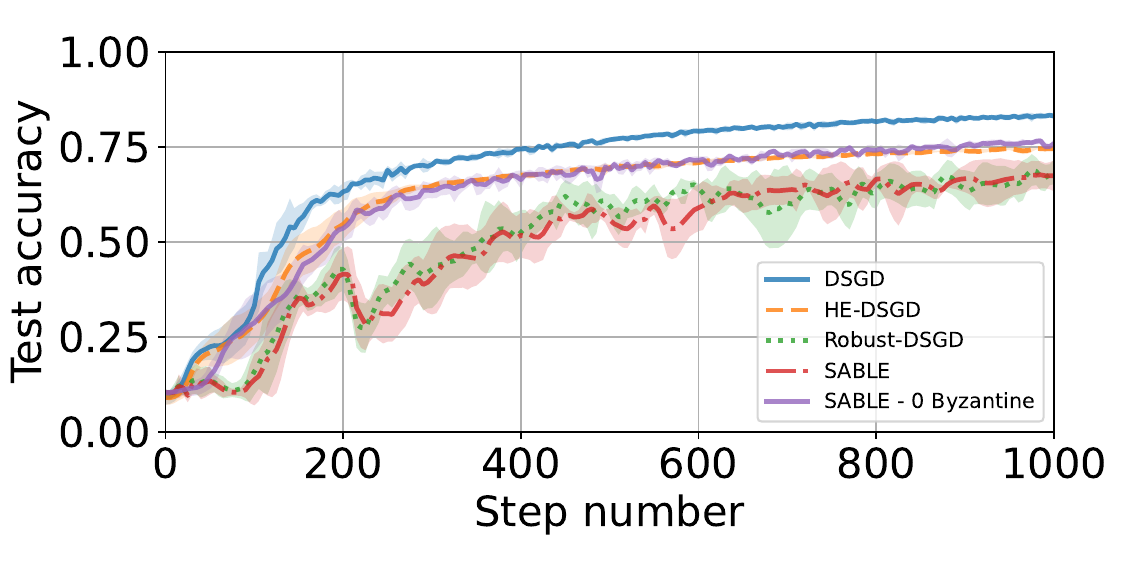}%
    \vspace{-4mm}
    \caption{\justifying Fashion-MNIST with $f=7$ among $n = 15$. Byzantine attacks: FOE (\textit{row1, left}), ALIE (\textit{row 1, right}), LF(\textit{row 2, left}), and mimic (\textit{row 2, right}).}
\label{fig:plots_fashionmnist_f=7_hetero}
\end{figure*}

\begin{figure*}[!ht]
    \centering
    \includegraphics[width=0.45\textwidth]{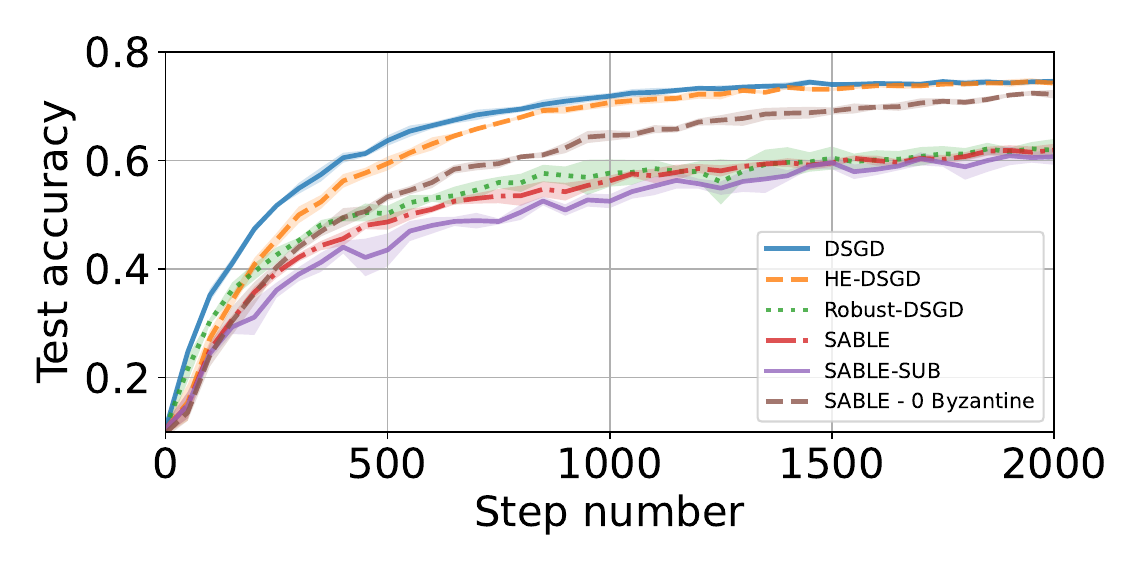}
    \includegraphics[width=0.45\textwidth]{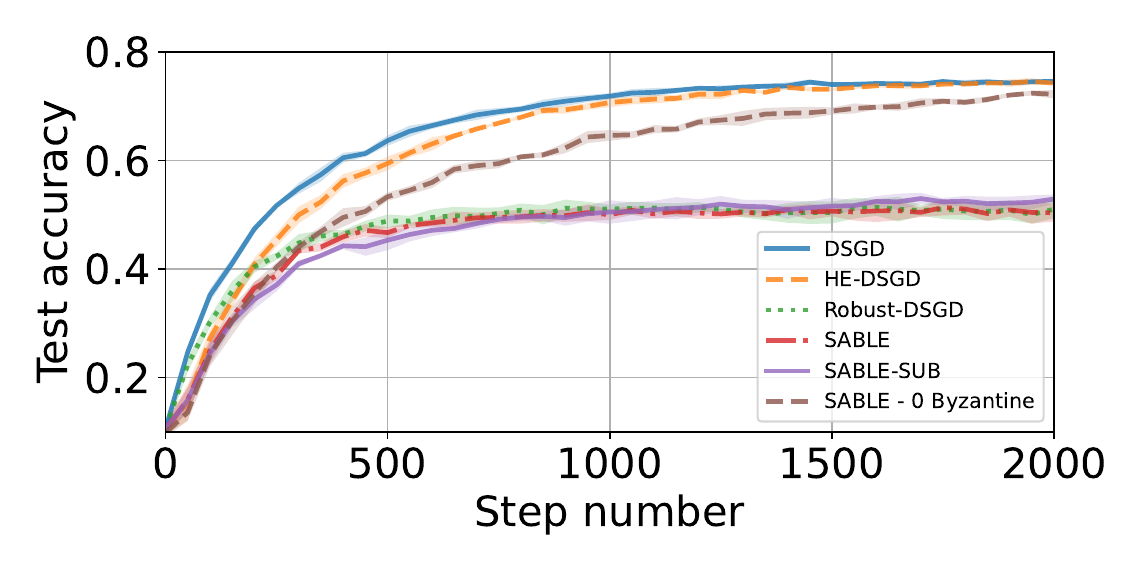}\\
    \vspace{-2mm}
    \includegraphics[width=0.45\textwidth]{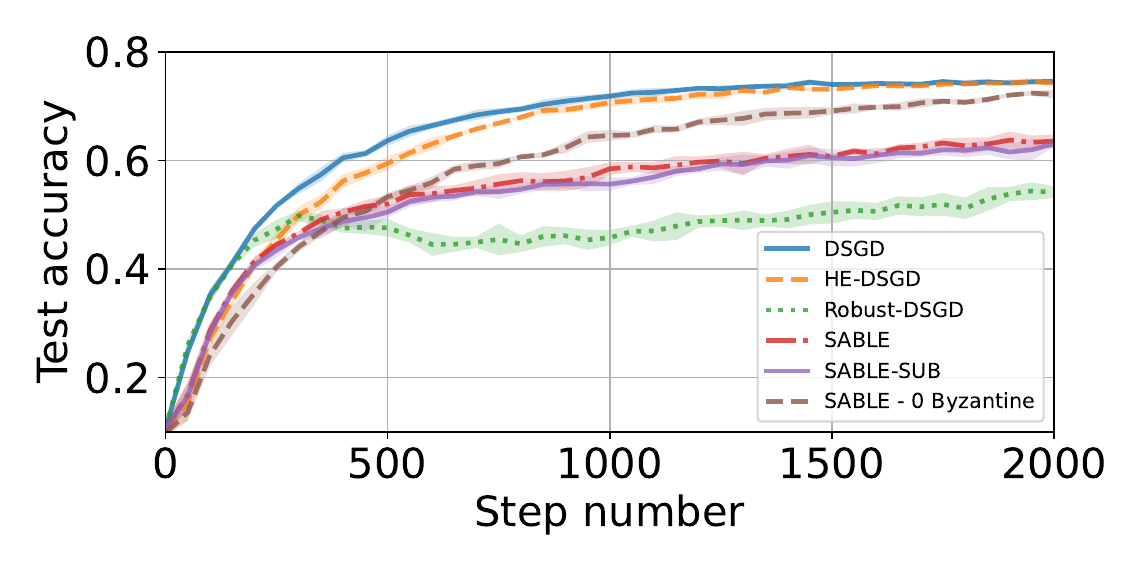}%
    \vspace{-4mm}
    \caption{\justifying CIFAR-10 with and $f=3$ Byzantine nodes among $n = 9$. The Byzantine nodes execute the FOE (\textit{left}), ALIE (\textit{right}), and LF (\textit{down}) attacks.}
\label{fig:plots_cifar10_f=3_homo}
\end{figure*}

\begin{figure*}[!ht]
    \centering
    \includegraphics[width=0.45\textwidth]{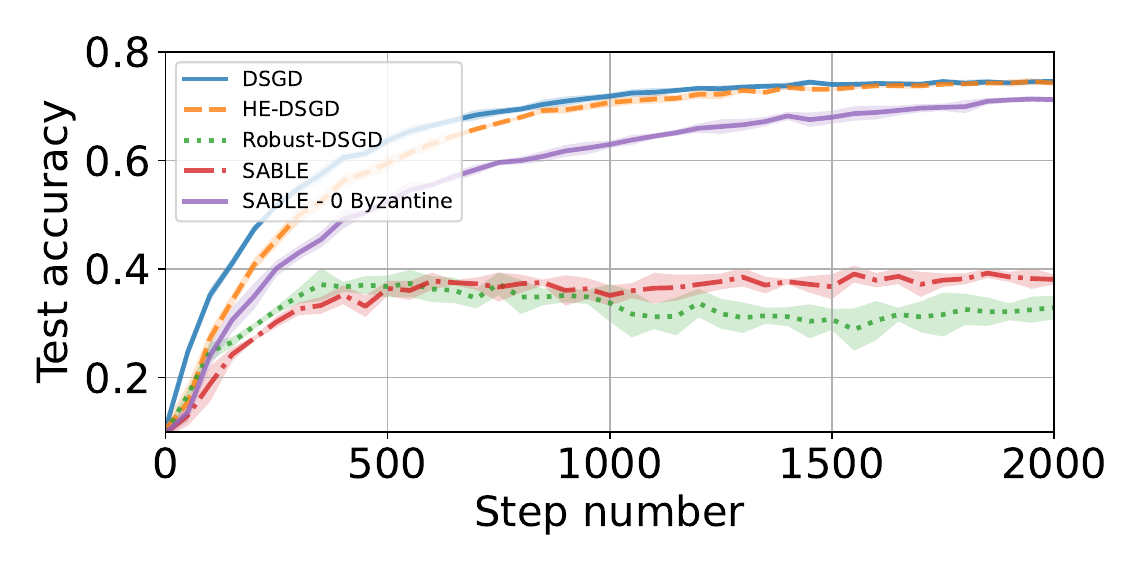}
    \includegraphics[width=0.45\textwidth]{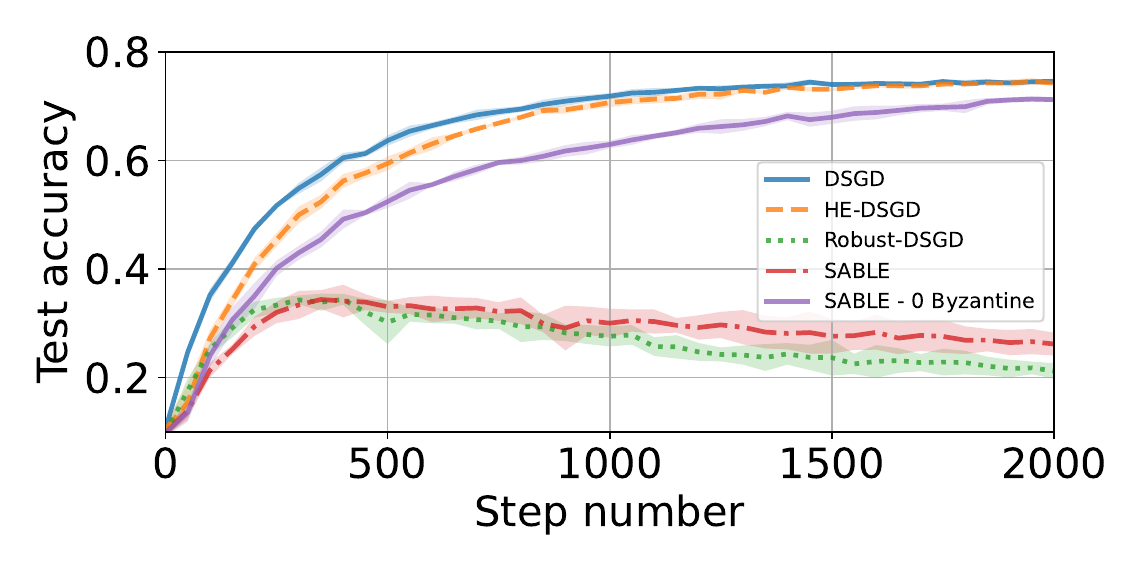}\\
    \vspace{-2mm}
    \includegraphics[width=0.45\textwidth]{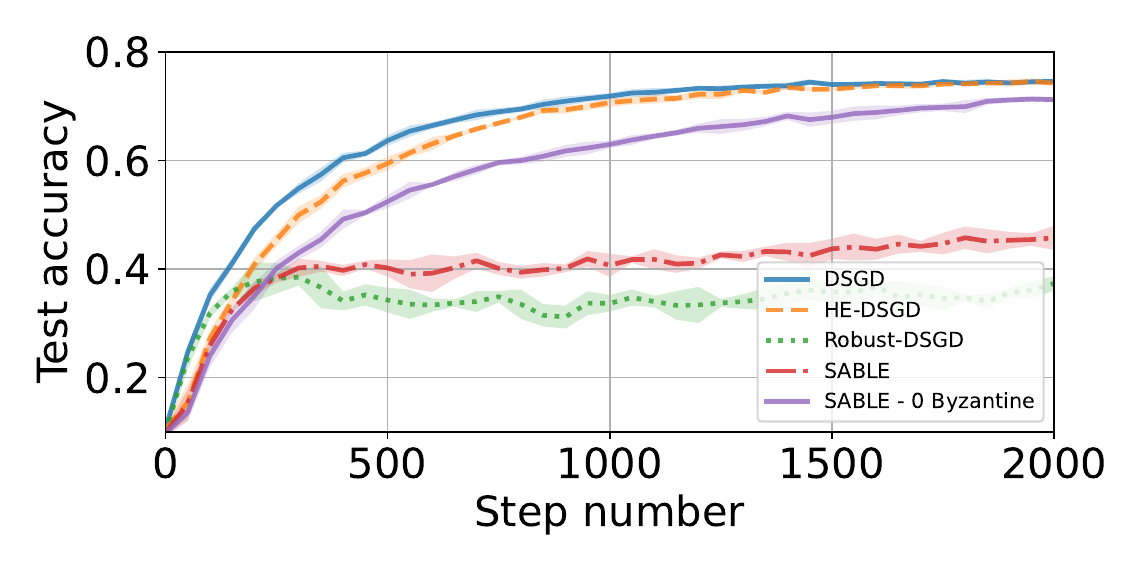}%
    \vspace{-4mm}
    \caption{\justifying CIFAR-10 with and $f=4$ Byzantine nodes among $n = 9$. The Byzantine nodes execute the FOE (\textit{left}), ALIE (\textit{right}), and LF (\textit{down}) attacks.}
\label{fig:plots_cifar10_f=4_homo}
\end{figure*}

In this section, we complete the experimental results that could not be placed in the main paper.
Figures~\ref{fig:plots_mnist_f=5_hetero} to~\ref{fig:plots_mnist_f=7_hetero} show the performance of \algoname{} on MNIST for $f \in \{3, 4, 5, 6, 7\}$.
Figures~\ref{fig:plots_fashionmnist_f=3_hetero},~\ref{fig:plots_fashionmnist_f=5_hetero}, and~\ref{fig:plots_fashionmnist_f=7_hetero} evaluate our algorithm on Fashion-MNIST when $f = 3, 5,$ and 7, respectively.
Finally, Figures~\ref{fig:plots_cifar10_f=3_homo} and~\ref{fig:plots_cifar10_f=4_homo} present our results on CIFAR-10 when $f = 3$ and 4, respectively.
These additional plots convey the same observations made in Section~\ref{sec_exp_results}.

\input{comparison_with_DP}

%% file: comparison_with_DP.tex
\section{Comparison with Differential Privacy (DP) + Robust-DSGD}\label{app_exp_results_DP}
\begin{figure*}[!ht]
    \centering
    \includegraphics[width=0.45\textwidth]{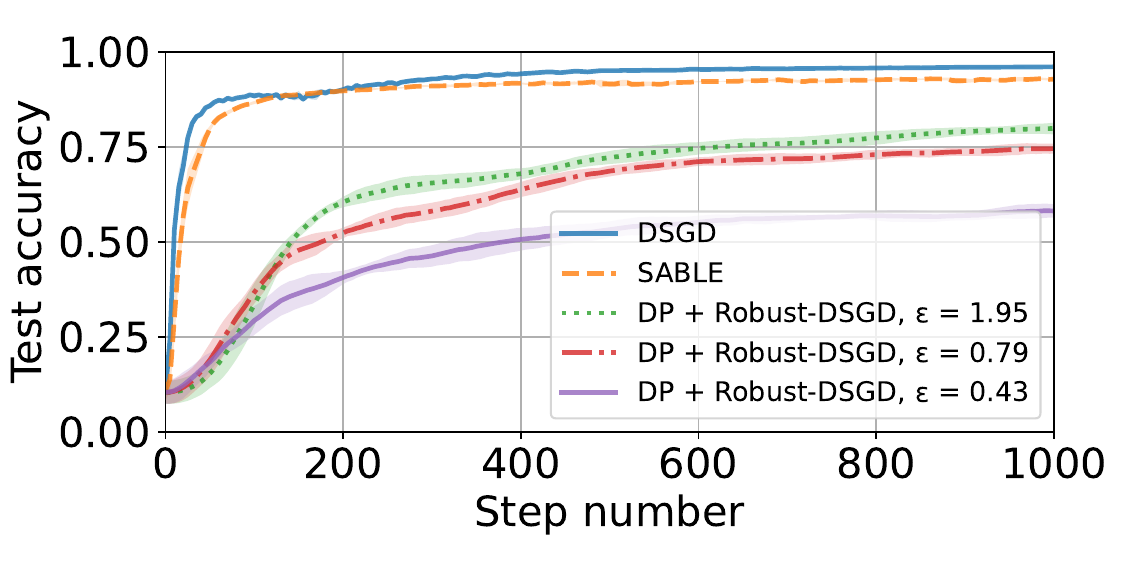}
    \includegraphics[width=0.45\textwidth]{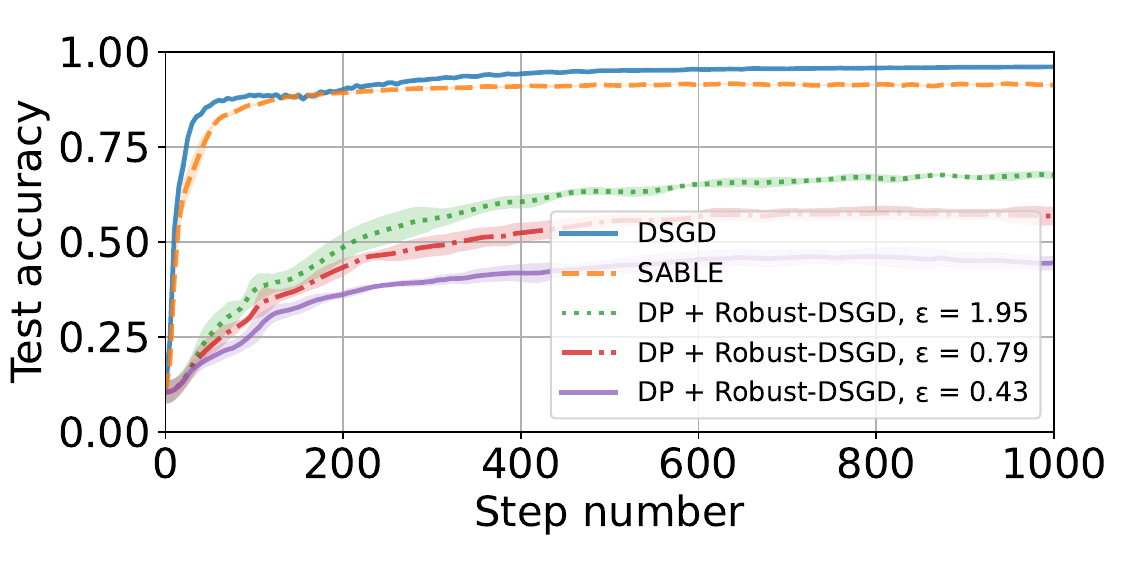}\\
    \vspace{-2mm}
    \includegraphics[width=0.45\textwidth]{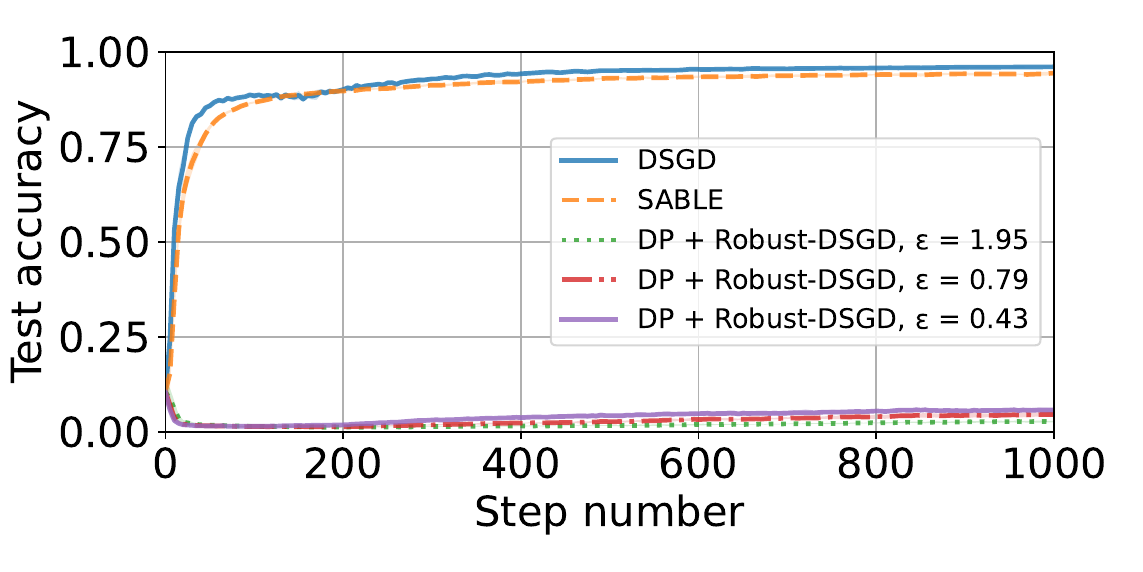}%
    \includegraphics[width=0.45\textwidth]{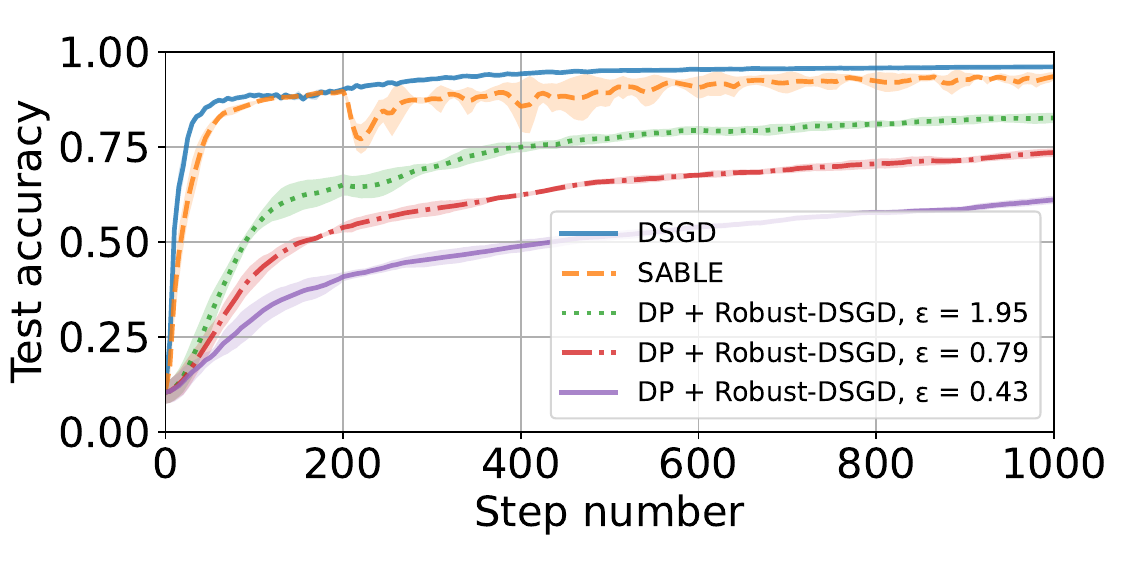}%
    \vspace{-4mm}
    \caption{\justifying Experiments comparing the performances of \algoname{} and DP + Robust-DSGD on MNIST with $f=3$ Byzantine nodes among $n = 15$. The Byzantine nodes execute the FOE (\textit{row1, left}), ALIE (\textit{row 1, right}), LF(\textit{row 2, left}), and mimic (\textit{row 2, right}) attacks.}
\label{fig:plots_mnist_f=3_DP}
\end{figure*}

In this section, we compare the performance of \algoname{} with a DP-based Byzantine robust learning algorithm, namely DP + Robust-DSGD.
This algorithm uses DP (instead of HE) to protect the privacy of the workers, and CWTM~\cite{yin2018byzantine} to mitigate the impact of Byzantine nodes.
More specifically, in every step of the learning, DP + Robust-DSGD consists in injecting DP noise to the gradients of the workers, and aggregating the workers' gradients using CWTM on the server.
The workers inject a privacy noise $\sigma_{\mathrm{DP}} = \frac{2\rho}{b} \times \sigma_{\mathrm{NM}}$ to their gradients, where $\sigma_{\mathrm{NM}}$ is referred to as the noise multiplier, $\rho = 0.3$ is the clipping threshold, and $b = 25$ is the batch size.
In our experiments, we consider three privacy regimes; namely \textit{low} privacy where $\sigma_{\mathrm{NM}} = 1.5$, \textit{moderate} privacy where $\sigma_{\mathrm{NM}} = 3$, and \textit{high} privacy where $\sigma_{\mathrm{NM}} = 5$.
We use Opacus~\cite{opacus}, a DP library for deep learning in PyTorch~\cite{pytorch}, to estimate the privacy budgets achieved at the end of the learning.
The aggregate privacy budgets after $T = 1000$ steps of learning are $(\epsilon, \delta) = (1.95, 10^{-4})$ in the \textit{low} privacy regime, $(\epsilon, \delta) = (0.79, 10^{-4})$ in the \textit{moderate} privacy regime, and $(\epsilon, \delta) = (0.43, 10^{-4})$ in the \textit{high} privacy regime.
Figure~\ref{fig:plots_mnist_f=3_DP} compares the performances of \algoname{} and DP + Robust-DSGD on MNIST with $\alpha = 1$, in a distributed system of $n = 15$ workers among which $f = 3$ are Byzantine.
While \algoname{} suffers from a slight degradation in utility (i.e., loss in ML accuracy) due to the quantization of gradients prior to encryption, this loss is very small compared to the decrease in performance of DP + Robust-DSGD.
Indeed, even under the low privacy regime, there exists a difference of at least 10\% between the final accuracies of \algoname{} and DP + Robust-DSGD across all Byzantine attacks (even more in the high privacy regime).
Furthermore, the LF attack completely deteriorates the robustness of the algorithm, under all privacy regimes.
These empirical results validate the fundamental utility cost that arises when simultaneously enforcing DP and Byzantine robustness, as shown previously~\cite{allouah2023trilemna}.

